\documentclass[twoside,11pt]{article}

\usepackage[preprint]{jmlr2e}

\usepackage{amsmath,amsfonts,bm}

\def\eqref#1{equation~\ref{#1}}

\def\1{\bm{1}}

\DeclareMathAlphabet{\mathsfit}{\encodingdefault}{\sfdefault}{m}{sl}
\SetMathAlphabet{\mathsfit}{bold}{\encodingdefault}{\sfdefault}{bx}{n}

\newcommand{\E}{\mathbb{E}}

\newcommand{\R}{\mathbb{R}}

\newcommand{\KL}{D_{\mathrm{KL}}}
\newcommand{\Var}{\mathrm{Var}}

\DeclareMathOperator*{\argmax}{arg\,max}
\DeclareMathOperator*{\argmin}{arg\,min}

\usepackage{booktabs}
\usepackage{enumitem}
\usepackage{mathtools}
\usepackage{amssymb}
\usepackage{bbm}
\usepackage{bm}
\usepackage{array}
\usepackage{multirow}
\usepackage{multicol}
\usepackage{makecell}
\usepackage{csquotes}

\usepackage{amssymb}
\usepackage{pifont}
\newcommand{\cmark}{\checkmark}

\usepackage[dvipsnames]{xcolor}

\newcommand{\inv}{^{-1}}

\usepackage{lastpage}
\jmlrheading{22}{2021}{1-\pageref{LastPage}}{11/20; Revised 6/21}{9/21}{20-1316}{Ian Covert, Scott Lundberg and Su-In Lee}
\ShortHeadings{Explaining by Removing}{Covert, Lundberg \& Lee}
\firstpageno{1}

\begin{document}

\title{Explaining by Removing: \\A Unified Framework for Model Explanation}

\author{\name Ian C. Covert
       \email icovert@cs.washington.edu \\
       \addr Paul G. Allen School of Computer Science \& Engineering \\
       University of Washington\\
       Seattle, WA 98195, USA
       \AND
       \name Scott Lundberg
       \email scott.lundberg@microsoft.com \\
       \addr Microsoft Research \\
       Microsoft Corporation\\
       Redmond, WA 98052, USA
       \AND
       \name Su-In Lee
       \email suinlee@cs.washington.edu \\
       \addr Paul G. Allen School of Computer Science \& Engineering \\
       University of Washington\\
       Seattle, WA 98195, USA}

\editor{Samuel Kaski}

\maketitle

\begin{abstract}%
Researchers have proposed a wide variety of model explanation approaches, but it remains unclear how most methods are related or when one method is preferable to another. We describe a new unified class of methods, \textit{removal-based explanations}, that are based on the principle of simulating feature removal to quantify each feature's influence. These methods vary in several respects, so we develop a framework that characterizes each method along three dimensions: 1)~how the method removes features, 2)~what model behavior the method explains, and 3)~how the method summarizes each feature's influence. Our framework unifies 26 existing methods, including several of the most widely used approaches: SHAP, LIME, Meaningful Perturbations, and permutation tests. This newly understood class of explanation methods has rich connections that we examine using tools that have been largely overlooked by the explainability literature. To anchor removal-based explanations in cognitive psychology, we show that feature removal is a simple application of subtractive counterfactual reasoning. Ideas from cooperative game theory shed light on the relationships and trade-offs among different methods, and we derive conditions under which all removal-based explanations have information-theoretic interpretations. Through this analysis, we develop a unified framework that helps practitioners better understand model explanation tools, and that offers a strong theoretical foundation upon which future explainability research can build.
\end{abstract}

\vskip 0.1in


\newpage

\section{Introduction}

The proliferation of black-box models has made machine learning (ML) explainability increasingly important, and researchers have now proposed a variety of model explanation approaches \citep{zeiler2014visualizing, ribeiro2016should, lundberg2017unified}. Despite progress in the field, the relationships and trade-offs among these methods have not been rigorously investigated, and researchers have not always formalized their fundamental ideas about how to interpret models \citep{lipton2018mythos}. This makes the interpretability literature difficult to navigate and raises questions about whether existing methods relate to human processes for explaining complex decisions \citep{miller2017explainable, miller2019explanation}.

\begin{figure*}[t]
\begin{center}
\includegraphics[width=\columnwidth]{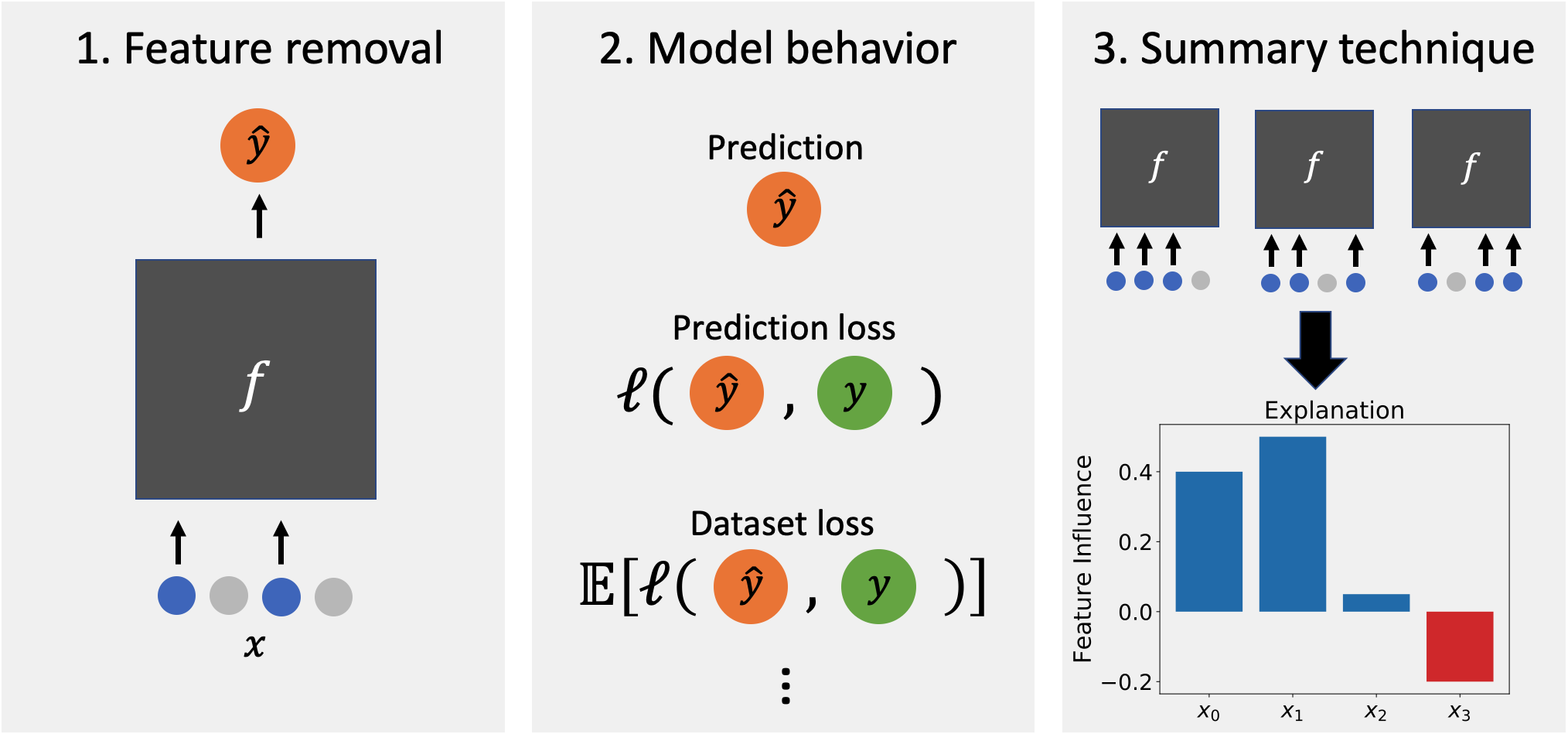}
\caption{A unified framework for \textit{removal-based explanations}. Each method is determined by three choices: how it removes features, what model behavior it analyzes, and how it summarizes feature influence.}
\label{fig:concept}
\end{center}
\vskip -0.1in
\end{figure*}

Here, we present a comprehensive new framework that unifies a substantial portion of the model explanation literature. Our framework is based on the observation that many methods can be understood as \textit{simulating feature removal} to quantify each feature's influence on a model. The intuition behind these methods is similar (depicted in Figure~\ref{fig:concept}), but each one takes a slightly different approach to the removal operation: some replace features with neutral values \citep{zeiler2014visualizing, petsiuk2018rise}, others marginalize over a distribution of values \citep{strobl2008conditional, lundberg2017unified}, and still others train separate models for each subset of features \citep{lipovetsky2001analysis, vstrumbelj2009explaining}. These methods also vary in other respects, as we describe below.

We refer to this class of approaches as \textit{removal-based explanations} and identify 26\footnote{This total count does not include minor variations on the approaches we identified.} existing methods that rely on the feature removal principle, including several of the most widely used methods (SHAP, LIME, Meaningful Perturbations, permutation tests). We then develop a framework that shows how each method arises from various combinations of three choices: 1)~how the method removes features from the model, 2)~what model behavior the method analyzes, and 3)~how the method summarizes each feature's influence on the model. By characterizing each method in terms of three precise mathematical choices, we are able to systematize their shared elements and show that they all rely on the same fundamental approach---feature removal.

The model explanation field has grown significantly in the past decade, and we take a broader view of the literature than existing unification theories. Our framework's flexibility lets us establish links between disparate classes of methods (e.g., computer vision-focused methods, global methods, game-theoretic methods, feature selection methods) and show that the literature is more interconnected than previously recognized. Exposing these relationships makes the literature more coherent, and it simplifies the process of reasoning about the benefits of each method by showing that their differences often amount to minor, interchangeable design choices.

To better understand the unique advantages of each method, we thoroughly analyze our framework's theoretical foundation by examining its connections with related fields. In particular, we find that cognitive psychology, cooperative game theory and information theory are intimately connected to removal-based explanations and help shed light on the trade-offs between different approaches. The extent of these links is perhaps surprising, because few methods explicitly reference these related fields.

Our approach yields many new results and provides a strong theoretical foundation for understanding existing methods and guiding future work. Our contributions include:

\begin{enumerate}
    \item We present a \textbf{unified framework} that characterizes 26 existing explanation methods and formalizes a new class of \textbf{removal-based explanations}. The framework integrates classes of methods that may have previously appeared disjoint, including local and global approaches as well as feature attribution and feature selection methods. In our experiments, we develop and compare 60$+$ new explanation approaches by mixing and matching the choices that methods make in each dimension of our framework.
    
    \item We develop \textbf{new mathematical tools} to represent different approaches for removing features from ML models. Then, by incorporating an underlying data distribution, we argue that marginalizing out features using their conditional distribution is the only approach that is consistent with standard probability axioms. Finally, we prove that several alternative choices approximate this approach and that this approach gives all removal-based explanations an \textbf{information-theoretic interpretation}.
    
    \item We demonstrate that \textbf{all removal-based explanations are implicitly tied to cooperative game theory}, and we leverage decades of game theory research to highlight advantages of the Shapley value over alternative allocation strategies. Building on these findings, we also show that several feature attribution techniques are generalized by LIME \citep{ribeiro2016should}, and that several feature selection techniques are generalized by the
    Masking Model approach \citep{dabkowski2017real}.
    
    \item We consult social science research to understand the intuition behind feature removal as an approach to model explanation. We find that feature removal is a simple application of \textbf{subtractive counterfactual reasoning}, or, equivalently, of Mill's \textbf{method of difference} from the philosophy of scientific induction \citep{mill1884system}.
\end{enumerate}

The paper is organized as follows. We begin with background on the model explanation problem and a review of prior work (Section~\ref{sec:background}), and we then give an overview of our framework (Section~\ref{sec:removal_based_explanations}). We then present our framework in detail, describing how methods remove features (Section~\ref{sec:removal}), formalizing the model behaviors analyzed by each method (Section~\ref{sec:behaviors}), and examining each method's approach to summarizing feature influence (Section~\ref{sec:explanations}). We next explore connections to related fields. First, we describe our framework's relationship with cooperative game theory (Section~\ref{sec:cooperative_game_theory}). Next, we prove that under certain conditions, removal-based explanations analyze the information communicated by each feature (Section~\ref{sec:information}). Then, we refer to the psychology literature to establish a cognitive basis for removal-based explanations (Section~\ref{sec:psychology}). In our experiments, we provide empirical comparisons between existing methods and new combinations of existing approaches, as well as a discussion of our framework's implications for common explainability metrics (Section~\ref{sec:experiments}). Finally, we recap and conclude our discussion (Section~\ref{sec:discussion}).

\section{Background} \label{sec:background}

Here, we introduce the model explanation problem and briefly review existing approaches and related unification theories.

\subsection{Preliminaries}

Consider a supervised ML model $f$ that is used to predict a response variable $Y \in \mathcal{Y}$ using the input $X = (X_1, X_2, \ldots, X_d)$, where each $X_i$ represents an individual feature, such as a patient's age. We use uppercase symbols (e.g., $X$) to denote random variables and lowercase ones (e.g., $x$) to denote their values. We also use $\mathcal{X}$ to denote the domain of the full feature vector $X$ and $\mathcal{X}_i$ to denote the domain of each feature $X_i$. Finally, $x_S \equiv \{x_i : i \in S\}$ denotes a subset of features for $S \subseteq D \equiv \{1, 2, \ldots d\}$, and $\bar S \equiv D \setminus S$ represents a set's complement.

ML interpretability broadly aims to provide insight into how models make predictions. This is particularly important when $f$ is a complex model, such as a neural network or a decision forest. The most active area of recent research is \textit{local interpretability}, which explains individual predictions, such as an individual patient diagnosis (e.g., \citealp{ribeiro2016should, lundberg2017unified, sundararajan2017axiomatic}); in contrast, \textit{global interpretability} explains the model's behavior across the entire dataset (e.g., \citealp{breiman2001random, owen2014sobol, covert2020understanding}). Both problems are usually addressed using \textit{feature attribution}, where scores are assigned to quantify each feature's influence. However, recent work has also proposed the strategy of \textit{local feature selection} \citep{chen2018learning}, and other papers have introduced methods to isolate sets of relevant features \citep{zhou2014object, fong2017interpretable, dabkowski2017real}.

Whether the aim is local or global interpretability, explaining the inner workings of complex models is fundamentally difficult, so it is no surprise that researchers continue to devise new approaches. Commonly cited categories of approaches include perturbation-based methods (e.g., \citealp{zeiler2014visualizing, lundberg2017unified}), gradient-based methods (e.g., \citealp{simonyan2013deep, montavon2017explaining, sundararajan2017axiomatic, selvaraju2017grad}), more general propagation-based methods (e.g., \citealp{springenberg2014striving, bach2015pixel, kindermans2017learning, zhang2018top}), and inherently interpretable models (e.g., \citealp{zhou2016learning, rudin2019stop}). However, these categories refer to loose collections of approaches that seldom share a precise mechanism.

Among the various approaches in the literature, many methods generate explanations by considering some class of perturbation to the input and the corresponding impact on the model's predictions.
Certain methods consider infinitesimal perturbations by calculating gradients\footnote{A view also mentioned by \cite{bhatt2020explainable}, for example.} \citep{simonyan2013deep, sundararajan2017axiomatic, smilkov2017smoothgrad, erion2019learning, xu2020attribution}, but there are many possible input perturbations \citep{zeiler2014visualizing, ribeiro2016should, fong2017interpretable, lundberg2017unified}. Our work is based on the observation that many perturbation strategies can be understood as simulating feature removal.

\subsection{Related work}

Prior work has made solid progress in exposing connections among disparate explanation methods. \cite{lundberg2017unified} proposed the unifying framework of \textit{additive feature attribution methods} and showed that LIME, DeepLIFT, LRP and QII are all related to SHAP \citep{bach2015pixel, ribeiro2016should, shrikumar2016not, datta2016algorithmic}. Similarly, \cite{ancona2017towards} showed that Grad * Input, DeepLIFT, LRP and Integrated Gradients can all be understood
as modified gradient backpropagations. Most recently, \cite{covert2020understanding} showed that several global explanation methods can be viewed as \textit{additive importance measures}, including permutation tests, Shapley Net Effects, feature ablation and SAGE \citep{breiman2001random, lipovetsky2001analysis, lei2018distribution}.

Relative to prior work, the unification we propose is considerably broader but nonetheless precise. By focusing on the common mechanism of removing features from a model, we encompass far more methods, including both local and global ones. We also provide a considerably richer theoretical analysis by exploring underlying connections with cooperative game theory, information theory and cognitive psychology.

As we describe below, our framework characterizes methods along three dimensions. The choice of how to remove features has been considered by many works \citep{lundberg2017unified, chang2018explaining, janzing2019feature, sundararajan2019many, merrick2019explanation, aas2019explaining, hooker2019please, agarwal2019explaining, frye2020shapley}. However, the choice of what model behavior to analyze has been considered explicitly by only a few works \citep{lundberg2020local, covert2020understanding}, as has the choice of how to summarize each feature's influence based on a cooperative game \citep{vstrumbelj2009explaining, datta2016algorithmic, lundberg2017unified, frye2019asymmetric, covert2020understanding}. To our knowledge, ours is the first work to consider all three dimensions simultaneously and discuss them under a single unified framework.

Besides the methods that we focus on, there are also many methods that do not rely on the feature removal principle. We direct readers to survey articles for a broader overview of the literature \citep{adadi2018peeking, guidotti2018survey}.

\section{Removal-Based Explanations} \label{sec:removal_based_explanations}

We now introduce our framework and briefly describe the methods it unifies.

\begin{table}[t]
\caption{Choices made by existing removal-based explanations.}
\label{tab:methods}
\begin{center}
\begin{scriptsize}
\begin{tabular}{lccc}
\toprule
\textsc{Method} & \textsc{Removal} & \textsc{Behavior} & \textsc{Summary} \\
\midrule
IME (2009) & Separate models & Prediction & Shapley value \\
IME (2010) & Marginalize (uniform) & Prediction & Shapley value \\
QII & Marginalize (marginals product) & Prediction & Shapley value \\
SHAP & Marginalize (conditional/marginal) & Prediction & Shapley value \\
KernelSHAP & Marginalize (marginal) & Prediction & Shapley value \\
TreeSHAP & Tree distribution & Prediction & Shapley value \\
LossSHAP & Marginalize (conditional) & Prediction loss & Shapley value \\
SAGE & Marginalize (conditional) & Dataset loss (label) & Shapley value \\
Shapley Net Effects & Separate models (linear) & Dataset loss (label) & Shapley value \\
SPVIM & Separate models & Dataset loss (label) & Shapley value \\
Shapley Effects & Marginalize (conditional) & Dataset loss (output) & Shapley value \\
Permutation Test & Marginalize (marginal) & Dataset loss (label) & Remove individual \\
Conditional Perm. Test & Marginalize (conditional) & Dataset loss (label) & Remove individual \\
Feature Ablation (LOCO) & Separate models & Dataset loss (label) & Remove individual \\
Univariate Predictors & Separate models & Dataset loss (label) & Include individual \\
L2X & Surrogate & Prediction loss (output) & High-value subset \\
REAL-X & Surrogate & Prediction loss (output) & High-value subset \\
INVASE & Missingness during training & Prediction mean loss & High-value subset \\
LIME (Images) & Default values & Prediction & Additive model \\
LIME (Tabular) & Marginalize (replacement dist.) & Prediction & Additive model \\
PredDiff & Marginalize (conditional) & Prediction & Remove individual \\
Occlusion & Zeros & Prediction & Remove individual \\
CXPlain & Zeros & Prediction loss & Remove individual \\
RISE & Zeros & Prediction & Mean when included \\
MM & Default values & Prediction & Partitioned subsets \\
MIR & Extend pixel values & Prediction & High-value subset \\
MP & Blurring & Prediction & Low-value subset \\
EP & Blurring & Prediction & High-value subset \\
FIDO-CA & Generative model & Prediction & High-value subset \\
\bottomrule
\end{tabular}
\end{scriptsize}
\end{center}
\vskip -0.2in
\end{table}

\subsection{A unified framework}

We develop a unified model explanation framework by connecting methods that define each feature's influence through the impact of removing it from a model. This principle describes a substantial portion of the explainability literature: we find that 26 existing methods rely on this mechanism, including many of the most widely used approaches \citep{breiman2001random, ribeiro2016should, fong2017interpretable, lundberg2017unified}. Several works have described their methods as either \textit{removing}, \textit{ignoring} or \textit{deleting} information from a model, but our work is the first to precisely characterize this approach and document its use throughout the model explanation field.

The methods that we identify all remove groups of features from the model, but, beyond that, they take a diverse set of approaches. For example, LIME fits a linear model to an interpretable representation of the input \citep{ribeiro2016should}, L2X selects the most informative features for a single example \citep{chen2018learning}, and Shapley Effects examines how much of the model's variance is explained by each feature \citep{owen2014sobol}. Perhaps surprisingly, the differences between these methods are easy to systematize because they are all based on removing discrete sets of features.

As our main contribution, we introduce a framework that shows how these methods can be specified using only three choices.

\begin{definition}
    {\normalfont \textbf{Removal-based explanations}} are model explanations that quantify the impact of removing groups of features from the model. These methods are determined by three choices:
    
    \begin{enumerate}
        \item (Feature removal) How the method removes features from the model (e.g. by setting them to default values, or by marginalizing over a distribution of values)
        \item (Model behavior) What model behavior the method analyzes (e.g., the probability of the true class, or the model loss)
        \item (Summary technique) How the method summarizes each feature's impact on the model (e.g., by removing a feature individually, or by calculating Shapley values)
    \end{enumerate}
\end{definition}

These three dimensions are independent of one another (i.e., any combination of choices is possible), but all three are necessary to fully specify a removal-based explanation. The first two choices, feature removal and model behavior, allow us to probe how a model's predictions change when given access to arbitrary subsets of features---including the behavior with all features, or with one or more features removed. Then, because there are an exponential number of feature combinations to consider, a summary technique is required to condense this information into a human-interpretable explanation, typically using either attribution scores or a subset of highly influential features.

As we show in the following sections, each dimension of the framework is represented by a specific mathematical choice. This precise yet flexible framework allows us to unify disparate classes of explanation methods, and, by unraveling each method's choices, offers a step towards a better understanding of the literature.

\subsection{Overview of existing approaches}

We now outline our findings, which we present in more detail in the remainder of the paper. In particular, we preview how existing methods fit into our framework and highlight groups of methods that appear closely related in light of our feature removal perspective.

Table~\ref{tab:methods} lists the methods unified by our framework, with acronyms and the original works introduced in Section~\ref{sec:removal}. These methods represent diverse parts of the interpretability literature, including global interpretability methods \citep{breiman2001random, owen2014sobol, covert2020understanding}, computer vision-focused methods \citep{zeiler2014visualizing, zhou2014object, fong2017interpretable, petsiuk2018rise}, game-theoretic methods \citep{vstrumbelj2010efficient, datta2016algorithmic, lundberg2017unified} and feature selection methods \citep{chen2018learning, yoon2018invase, fong2019understanding}. They are all unified by their reliance on feature removal, and they can be described concisely via their three choices within our framework.

Disentangling the details of each method shows that many approaches share one or more of the same choices. For example, most methods choose to explain individual predictions (the model behavior), and the Shapley value \citep{shapley1953value} is the most popular summary technique. These common choices reveal that methods sometimes differ along only one or two dimensions, making it easier to reason about the trade-offs among approaches that might otherwise be viewed as monolithic algorithms.

\begin{figure*}[t]
\vskip -0.2in
\begin{center}
\includegraphics[trim=1.4cm 0cm 6.7cm 0cm, clip=true, width=\textwidth]{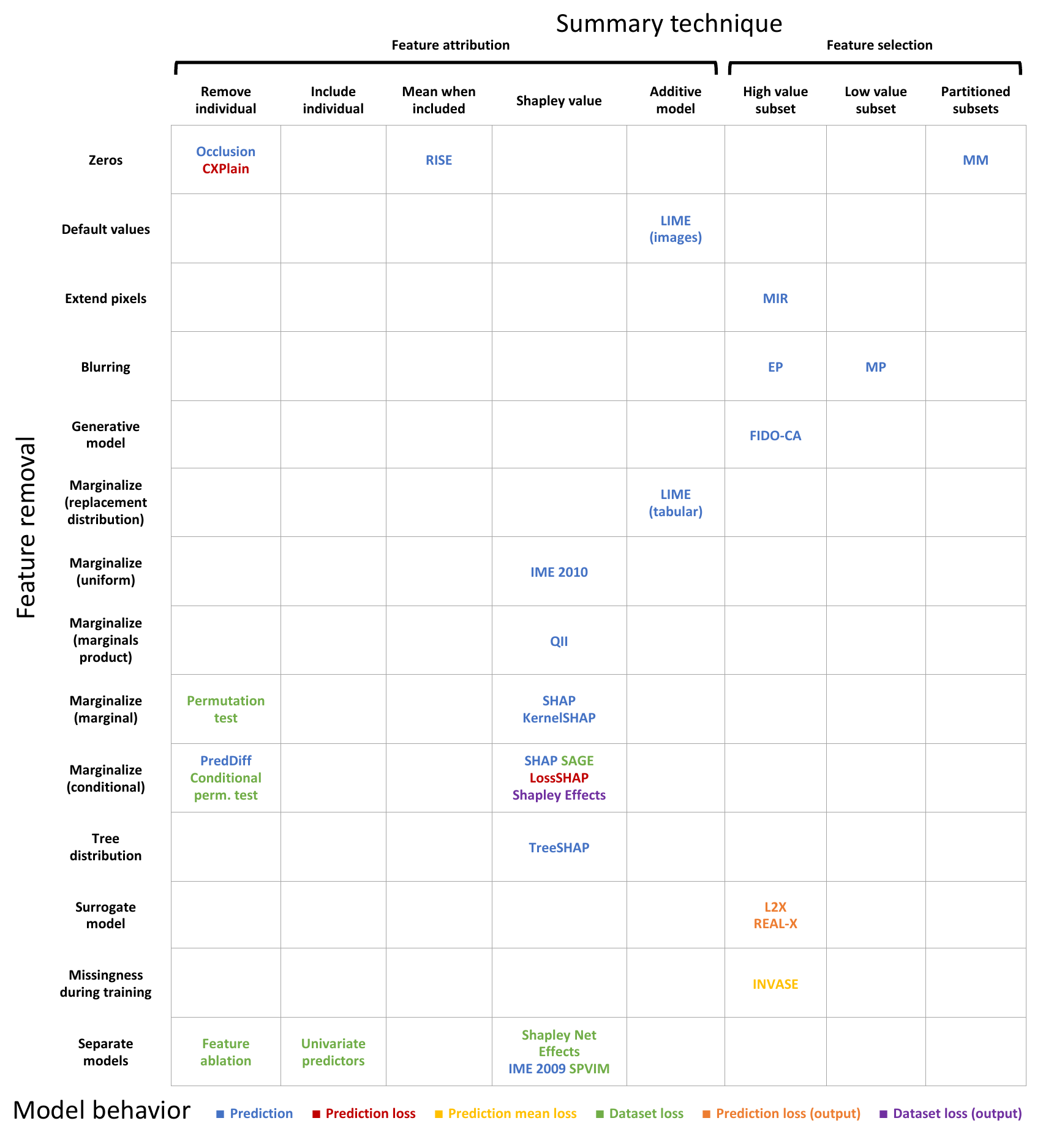}
\caption{Visual depiction of the space of removal-based explanations.}
\label{fig:methods_grid}
\end{center}
\vskip -0.2in
\end{figure*}

To further highlight similarities among these methods, we visually depict the space of removal-based explanations in Figure~\ref{fig:methods_grid}. Visualizing our framework reveals several regions in the space of methods that are crowded (e.g., methods that marginalize out removed features with their conditional distribution and that calculate Shapley values), while certain methods are relatively unique and spatially isolated (e.g., RISE; LIME for tabular data; INVASE). Empty positions in the grid reveal opportunities to develop new methods; in our experiments, we explore these possibilities by partially filling out the space of removal-based explanations (Section~\ref{sec:experiments}).

\begin{table}[t]
\caption{Common combinations of choices in existing methods. Check marks (\cmark) indicate choices that are identical between methods.}
\label{tab:common_combinations}
\begin{center}
\begin{small}
\begin{tabular}{cccc}
\toprule
\textsc{Removal} & \textsc{Behavior} & \textsc{Summary} & \textsc{Methods} \\
\midrule
& \cmark & \cmark & \makecell{IME, QII, SHAP, KernelSHAP, TreeSHAP} \\
\midrule
\cmark & & \cmark & \makecell{SHAP, LossSHAP, SAGE, Shapley Effects} \\
\midrule
\cmark & \cmark & & \makecell{Occlusion, LIME (images), MM, RISE} \\ 
\midrule
& \cmark & \cmark & \makecell{Feature ablation (LOCO), permutation tests,\\conditional permutation tests} \\
\midrule
\cmark & \cmark & & \makecell{Univariate predictors, feature ablation (LOCO),\\Shapley Net Effects, SPVIM} \\
\midrule
& \cmark & \cmark & SAGE, Shapley Net Effects, SPVIM \\
\midrule
\cmark & \cmark & & \makecell{SAGE, conditional permutation tests} \\
\midrule
\cmark & & \cmark & \makecell{Shapley Net Effects, SPVIM, IME (2009)} \\
\midrule
\cmark & & \cmark & Occlusion, CXPlain \\
\midrule
& \cmark & \cmark & Occlusion, PredDiff \\ 
\midrule
\cmark & & \cmark & \makecell{Conditional permutation tests, PredDiff} \\
\midrule
\cmark & \cmark & & SHAP, PredDiff \\
\midrule
\cmark & \cmark & & MP, EP \\
\midrule
& \cmark & \cmark & EP, FIDO-CA \\
\midrule
\cmark & \cmark & \cmark & L2X, REAL-X\footnotemark[3] \\
\midrule
\cmark & \cmark & \cmark & Shapley Net Effects, SPVIM\footnotemark[4] \\
\bottomrule
\end{tabular}
\end{small}
\end{center}
\vskip -0.2in
\end{table}

\footnotetext[3]{Although they share all three choices, L2X and REAL-X generate different explanations due to REAL-X's modified surrogate model training approach (Appendix~\ref{app:methods}).}
\addtocounter{footnote}{1}
\footnotetext[4]{SPVIM generalizes Shapley Net Effects to black-box models by using an efficient Shapley value estimation technique (Appendix~\ref{app:methods}).}
\addtocounter{footnote}{1}

Finally, Table~\ref{tab:common_combinations} shows groups of methods that differ in only one dimension of the framework. These methods are neighbors in the space of explanation methods (Figure~\ref{fig:methods_grid}), and it is noteworthy how many instances of neighboring methods exist in the literature. Certain methods even have neighbors along every dimension of the framework (e.g., SHAP, SAGE, Occlusion, PredDiff, conditional permutation tests), reflecting how intimately connected the field has become. In the remainder of the paper, after analyzing the choices made by each method, we consider perspectives from related fields that help reason about the conceptual and computational advantages that arise from the sometimes subtle differences between methods.

\section{Feature Removal} \label{sec:removal}

Here, we begin presenting our framework for removal-based explanations in detail. We first define the mathematical tools necessary to remove features from ML models, and we then examine how existing explanation methods remove features.

\subsection{Functions on feature subsets}

The principle behind removal-based explanations is to remove groups of features to understand their impact on a model, but since most models require all the features to make predictions, removing a feature is more complicated than simply not giving the model access to it. Most ML models make predictions given a specific set of features $X = (X_1, \ldots, X_d)$, and mathematically, these models are functions are of the form $f: \mathcal{X} \mapsto \mathcal{Y}$, where we use $\mathcal{F}$ to denote the set of all such possible mappings.

To remove features from a model, or to make predictions given a subset of features, we require a different mathematical object than $f \in \mathcal{F}$. Instead of functions with domain $\mathcal{X}$, we consider functions with domain $\mathcal{X} \times \mathcal{P}(D)$, where $\mathcal{P}(D)$ denotes the power set of $D \equiv \{1, \ldots, d\}$. To ensure invariance to the held out features, these functions must depend only on features specified by the subset $S \in \mathcal{P}(D)$, so we formalize \textit{subset functions} as follows.

\begin{definition} \label{def:subset_function}
    {\normalfont A \textbf{subset function}} is a mapping of the form
    
    \begin{equation*}
        F: \mathcal{X} \times \mathcal{P}(D) \mapsto \mathcal{Y}
    \end{equation*}
    
    \noindent that is invariant to the dimensions that are not in the specified subset. That is, we have $F(x, S) = F(x', S)$ for all $(x, x', S)$ such that $x_S = x'_S$. We define $F(x_S) \equiv F(x, S)$ for convenience because the held out values $x_{\bar S}$ are not used by $F$.
\end{definition}

A subset function's invariance property is crucial to ensure that only the specified feature values determine the function's output, while guaranteeing that the other feature values do not matter. Another way of viewing subset functions is that they provide an approach to accommodate missing data. While we use $\mathcal{F}$ to represent standard prediction functions, we use $\mathfrak{F}$ to denote the set of all possible subset functions.

We introduce subset functions here because they help conceptualize how different methods remove features from ML models. Removal-based explanations typically begin with an existing model $f \in \mathcal{F}$, and in order to quantify each feature's influence, they must establish a convention for removing it from the model. A natural approach is to define a subset function $F \in \mathfrak{F}$ based on the original model $f$. To formalize this idea, we define a model's \textit{subset extension} as follows.

\begin{definition} \label{def:extension}
    An {\normalfont \textbf{subset extension}} of a model $f \in \mathcal{F}$ is a subset function $F \in \mathfrak{F}$ that agrees with $f$ in the presence of all features. That is, the model $f$ and its subset extension $F$ must satisfy
    
    \begin{equation*}
        F(x) = f(x) \quad \forall x \in \mathcal{X}.
    \end{equation*}
\end{definition}

As we show next, specifying a subset function $F \in \mathfrak{F}$, often as a subset extension of an existing model $f \in \mathcal{F}$, is the first step towards defining a removal-based explanation.

\subsection{Removing features from machine learning models} \label{sec:missingness_strategies}

Existing methods have devised numerous ways to evaluate models while withholding groups of features. Although certain methods use different terminology to describe their approaches (e.g., deleting information, ignoring features, using neutral values, etc.), the goal of all these methods is to measure a feature's influence through the impact of removing it from the model. Most proposed techniques can be understood as subset extensions $F \in \mathfrak{F}$ of an existing model $f \in \mathcal{F}$ (Definition~\ref{def:extension}).

The various approaches used in existing work (see Appendix~\ref{app:methods} for more details) include:

\begin{itemize}
    \item (Zeros) Occlusion \citep{zeiler2014visualizing}, RISE \citep{petsiuk2018rise} and CXPlain \citep{schwab2019cxplain} remove features simply by setting them to zero:
    
    \begin{equation}
        F(x_S) = f(x_S, 0). \label{eq:zeros}
    \end{equation}
    
    \item (Default values) LIME for image data \citep{ribeiro2016should} and the Masking Model method (MM, \citealp{dabkowski2017real}) remove features by setting them to user-defined default values (e.g., gray pixels for images). Given default values $r \in \mathcal{X}$, these methods calculate

    \begin{equation}
        F(x_S) = f(x_S, r_{\bar S}). \label{eq:default}
    \end{equation}
    
    Sometimes referred to as the \textit{baseline} method \citep{sundararajan2019many}, this is a generalization of the previous approach, and in some cases features may be given different default values (e.g., their mean).
    
    
    \item (Extend pixel values) Minimal image representation (MIR, \citealp{zhou2014object}) removes features from images by extending the values of neighboring pixels. This effect is achieved through a gradient-space manipulation.

    
    \item (Blurring) Meaningful Perturbations (MP, \citealp{fong2017interpretable}) and Extremal Perturbations (EP, \citealp{fong2019understanding}) remove features from images by blurring them with a Gaussian kernel. This approach is \textit{not} a subset extension of $f$ because the blurred image retains dependence on the removed features. Blurring fails to remove large, low frequency objects (e.g., mountains), but it provides an approximate way to remove information from images.
    
    \item (Generative model) FIDO-CA \citep{chang2018explaining} removes features by replacing them with samples from a conditional generative model (e.g. \citealp{yu2018generative}). The held out features are drawn from a generative model represented by $p_G(X_{\bar S} \mid X_S)$, or $\tilde x_{\bar S} \sim p_G(X_{\bar S} \mid X_S)$, and predictions are made as follows:
    
    \begin{equation}
        F(x_S) = f(x_S, \tilde x_{\bar S}). \label{eq:generative}
    \end{equation}

    \item (Marginalize with conditional) SHAP \citep{lundberg2017unified}, LossSHAP \citep{lundberg2020local} and SAGE \citep{covert2020understanding} present a strategy for removing features by marginalizing them out using their conditional distribution $p(X_{\bar S} \mid X_S = x_S)$:
    
    \begin{equation}
        F(x_S) = \E\big[f(X) \mid X_S = x_S\big]. \label{eq:conditional_expectation}
    \end{equation}
    
    This approach is computationally challenging in practice, but recent work tries to achieve close approximations \citep{aas2019explaining, aas2021explaining, frye2020shapley}. Shapley Effects \citep{owen2014sobol} implicitly uses this convention to analyze function sensitivity, while conditional permutation tests \citep{strobl2008conditional} and Prediction Difference Analysis (PredDiff, \citealp{zintgraf2017visualizing}) propose simple approximations, with the latter conditioning only on groups of bordering pixels.

    \item (Marginalize with marginal) KernelSHAP (a practical implementation of SHAP) removes features by marginalizing them out using their joint marginal distribution $p(X_{\bar S})$:
    
    \begin{equation}
        F(x_S) = \E\big[f(x_S, X_{\bar S})\big]. \label{eq:marginal}
    \end{equation}
    
    This is the default behavior in SHAP's implementation,\footnote{\url{https://github.com/slundberg/shap}} and recent work discusses its potential benefits over conditional marginalization \citep{janzing2019feature}. Permutation tests \citep{breiman2001random} use this approach to remove individual features.

    \item (Marginalize with product of marginals) Quantitative Input Influence (QII, \citealp{datta2016algorithmic}) removes held out features by marginalizing them out using the product of the marginal distributions $p(X_i)$:
    
    \begin{equation}
        F(x_S) = \E_{\prod_{i \in D} p(X_i)}\big[f(x_S, X_{\bar S})\big]. \label{eq:qii}
    \end{equation}
    
    \item (Marginalize with uniform) The updated version of the Interactions Method for Explanation (IME, \citealp{vstrumbelj2010efficient}) removes features by marginalizing them out with a uniform distribution over the feature space. If we let $u_i(X_i)$ denote a uniform distribution over $\mathcal{X}_i$ (with extremal values defining the boundaries for continuous features), then features are removed as follows:
    
    \begin{equation}
        F(x_S) = \E_{\prod_{i \in D} u_i(X_i)}\big[f(x_S, X_{\bar S})\big]. \label{eq:ime_uniform}
    \end{equation}
    
    \item (Marginalize with replacement distributions) LIME for tabular data replaces features with independent draws from \textit{replacement distributions} (our term), each of which depends on the original feature values. When a feature $X_i$ with value $x_i$ is removed, discrete features are drawn from the distribution $p(X_i \mid X_i \neq x_i)$; when quantization is used for continuous features (LIME's default behavior\footnote{\url{https://github.com/marcotcr/lime}}), continuous features are simulated by first generating a different quantile and then simulating from a truncated normal distribution within that bin. If we denote each feature's replacement distribution given the original value $x_i$ as $q_{x_i}(X_i)$, then LIME for tabular data removes features as follows:
    
    \begin{equation}
        F(x, S) = \E_{\prod_{i \in D} q_{x_i}(X_i)}\big[f(x_S, X_{\bar S})\big].
    \end{equation}
    
    Although this function $F$ agrees with $f$ given all features, it is \textit{not} a subset extension because it does not satisfy the invariance property necessary for subset functions.
    
    \item (Tree distribution) Dependent TreeSHAP \citep{lundberg2020local} removes features using the distribution induced by the underlying tree model, which roughly approximates the conditional distribution. When splits for removed features are encountered in the model's trees, TreeSHAP averages predictions from the multiple paths in proportion to how often the dataset follows each path.
    
    \item (Surrogate model) Learning to Explain (L2X, \citealp{chen2018learning}) and REAL-X \citep{jethani2021have} train separate surrogate models $F$ to match the original model's predictions when groups of features are held out. The surrogate model accommodates missing features, allowing us to represent it as a subset function $F \in \mathfrak{F}$, and it aims to provide the following approximation:
    
    \begin{equation}
        F(x_S) \approx \E\big[f(X) \mid X_S = x_S\big]. \label{eq:surrogate}
    \end{equation}
    
    The surrogate model approach was also proposed separately in the context of Shapley values \citep{frye2020shapley}.
    
    \item (Missingness during training) Instance-wise Variable Selection (INVASE, \citealp{yoon2018invase}) uses a model that has missingness introduced at training time. Removed features are replaced with zeros, so that the model makes the following approximation:
    
    \begin{equation}
        F(x_S) = f(x_S, 0) \approx p(Y \mid X_S = x_S). \label{eq:missingness}
    \end{equation}
    
    This approximation occurs for models trained cross entropy loss, but other loss functions may lead to different results (e.g., the conditional expectation for MSE loss). Introducing missingness during training differs from the default values approach because the model is trained to recognize zeros (or other replacement values) as missing values rather than zero-valued features.
    
    \item (Separate models) The original version of IME \citep{vstrumbelj2009explaining} is not based on a single model $f$, but rather on separate models trained for each feature subset, or $\{f_S : S \subseteq D\}$. The prediction for a subset of features is given by that subset's model:
    
    \begin{equation}
        F(x_S) = f_S(x_S).
    \end{equation}
    
    Shapley Net Effects \citep{lipovetsky2001analysis} uses an identical approach in the context of linear models, with SPVIM generalizing the approach to black-box models \citep{williamson2020efficient}. Similarly, feature ablation, also known as leave-one-covariate-out (LOCO, \citealp{lei2018distribution}), trains models to remove individual features, and the univariate predictors approach (used mainly for feature selection) uses models trained with individual features \citep{guyon2003introduction}.
    Although the separate models approach is technically a subset extension of the model $f_D$ trained with all features, its predictions given subsets of features are not based on $f_D$.
\end{itemize}

Most of these approaches can be viewed as subset extensions of an existing model $f$, so our formalisms provide useful tools for understanding how removal-based explanations remove features from models. However, there are two exceptions: the blurring technique (MP and EP) and LIME's approach with tabular data. Both provide functions of the form $F: \mathcal{X} \times \mathcal{P}(D) \mapsto \mathcal{Y}$ that agree with $f$ given all features, but that still exhibit dependence on removed features. Based on our invariance property for held out features (Definition~\ref{def:subset_function}), we argue that these approaches do not fully remove features from the model.

We conclude that the first dimension of our framework amounts to choosing a subset function $F \in \mathfrak{F}$, often via a subset extension to an existing model $f \in \mathcal{F}$. We defer consideration of the trade-offs between these approaches until Section~\ref{sec:information}, where we show that one approach to removing features yields to connections with information theory.

\section{Explaining Different Model Behaviors} \label{sec:behaviors}

Removal-based explanations all aim to demonstrate how a model functions, but they can do so by analyzing different model behaviors. We now consider the various choices of target quantities to observe as different features are withheld from the model.

The feature removal principle is flexible enough to explain virtually any function. For example, methods can explain a model's prediction, a model's loss function, a hidden layer in a neural network, or any node in a computation graph. In fact, removal-based explanations need not be restricted to the ML context: any function that accommodates missing inputs can be explained via feature removal by observing either its output or some function of its output as groups of inputs are removed. This perspective shows the broad potential applications for removal-based explanations.

Because our focus is the ML context, we proceed by examining how existing methods work. Each explanation method's target quantity can be understood as a function of the model output, which for simplicity is represented by a subset function $F(x_S)$. Many methods explain the model output or a simple function of the output, such as the logits or log-odds ratio. Other methods take into account a measure of the model's loss, for either an individual input or the entire dataset. Ultimately, as we show below, each method generates explanations based on a set function of the form

\begin{equation}
    u: \mathcal{P}(D) \mapsto \R,
\end{equation}

\noindent which represents a value associated with each subset of features $S \subseteq D$. This set function corresponds to the model behavior that a method is designed to explain.

We now examine the specific choices made by existing methods (see Appendix~\ref{app:methods} for further details). The various model behaviors that methods analyze, and their corresponding set functions, include:

\begin{itemize}
    \item (Prediction) Occlusion, RISE, PredDiff, MP, EP, MM, FIDO-CA, MIR, LIME, SHAP (including KernelSHAP and TreeSHAP), IME and QII all analyze a model's prediction for an individual input $x \in \mathcal{X}$:
    
    \begin{equation}
        u_x(S) = F(x_S). \label{eq:output_game}
    \end{equation}
    
    These methods examine how holding out different features makes an individual prediction either higher or lower. For multi-class classification models, methods often use a single output that corresponds to the class of interest, and they can optionally apply a simple function to the model's output (for example, using the log-odds ratio rather than the classification probability).
    
    \item (Prediction loss) LossSHAP and CXPlain take into account the true label $y$ for an input $x$ and calculate the prediction loss using a loss function $\ell$:
    
    \begin{equation}
        v_{xy}(S) = - \ell\big(F(x_S), y\big). \label{eq:shap_loss_game}
    \end{equation}
    
    By incorporating the label, these methods quantify whether certain features make the prediction more or less correct. Note that the minus sign is necessary to give the set function a higher value when more informative features are included.
    
    \item (Prediction mean loss) INVASE considers the expected loss for a given input $x$ according to the label's conditional distribution $p(Y \mid X = x)$:
    
    \begin{equation}
        v_x(S) = - \E_{p(Y \mid X = x)}\Big[\ell\big(F(x_S), Y\big)\Big]. \label{eq:pml_game}
    \end{equation}
    
    By averaging the loss across the label's distribution, INVASE highlights features that correctly predict what \textit{could} have occurred, on average.
    
    \item (Dataset loss) Shapley Net Effects, SAGE, SPVIM, feature ablation, permutation tests and univariate predictors consider the expected loss across the entire dataset:
    
    \begin{equation}
        v(S) = - \E_{XY}\Big[\ell\big(F(X_S), Y\big)\Big]. \label{eq:expected_loss_game}
    \end{equation}
    
    These methods quantify how much the model's performance degrades when different features are removed. This set function can also be viewed as the predictive power derived from sets of features \citep{covert2020understanding}, and recent work has proposed a SHAP value aggregation that is a special case of this approach \citep{frye2020shapley}.
    
    \item (Prediction loss w.r.t. output) L2X and REAL-X consider the loss between the full model output and the prediction given a subset of features:
    
    \begin{equation}
        w_x(S) = - \ell\big(F(x_S), F(x)\big). \label{eq:plo_game}
    \end{equation}
    
    These methods highlight features that on their own lead to similar predictions as the full feature set.
    
    \item (Dataset loss w.r.t. output) Shapley Effects considers the expected loss with respect to the full model output:
    
    \begin{equation}
        w(S) = - \E_X\Big[\ell\big(F(X_S), F(X)\big)\Big]. \label{eq:shapley_effects_game}
    \end{equation}
    
    Though related to the dataset loss approach \citep{covert2020understanding}, this approach focuses on each feature's influence on the model output rather than on the model performance.
\end{itemize}

Each set function serves a distinct purpose in exposing a model's dependence on different features. Several of the approaches listed above analyze the model's behavior for individual predictions (local explanations), while some take into account the model's behavior across the entire dataset (global explanations). Although their aims differ, these set functions are all in fact related. Each builds upon the previous ones by accounting for either the loss or data distribution, and their relationships can be summarized as follows:

\begin{align}
    v_{xy}(S) &= - \ell\big(u_x(S), y\big) \\
    w_x(S) &= - \ell\big( u_x(S), u_x(D) \big) \\
    v_x(S) &= \E_{p(Y \mid X = x)}\big[v_{xY}(S)\big] \\
    v(S) &= \E_{XY}\big[v_{XY}(S)\big] \label{eq:loss_shap_sage} \\
    w(S) &= \E_X\big[w_X(S)\big]
\end{align}

These relationships show that explanations based on one set function are in some cases related to explanations based on another. For example, \cite{covert2020understanding} showed that SAGE explanations are the expectation of explanations provided by LossSHAP---a relationship reflected in Eq.~\ref{eq:loss_shap_sage}.

Understanding these connections is made easier by the fact that our framework disentangles each method's choices rather than viewing each method as a monolithic algorithm. We conclude by reiterating that removal-based explanations can explain virtually any function, and that choosing what model behavior to explain amounts to selecting a set function $u: \mathcal{P}(D) \mapsto \R$ to represent the model's dependence on different sets of features.

\section{Summarizing Feature Influence} \label{sec:explanations}

The third choice for removal-based explanations is how to summarize each feature's influence on the model. We examine the various summarization techniques and then discuss their computational complexity and approximation approaches.

\subsection{Explaining set functions} \label{sec:summarization_strategies}

The set functions that represent a model's dependence on different features (Section~\ref{sec:behaviors}) are complicated mathematical objects that are difficult to communicate to users: the model's behavior can be observed for any subset of features, but there are an exponential number of feature subsets to consider. Removal-based explanations handle this challenge by providing users with a concise summary of each feature's influence.

We distinguish between two main types of summarization approaches: \textit{feature attribution} and \textit{feature selection}. Many methods provide explanations in the form of feature attributions, which are numerical scores $a_i\in \R$ given to each feature $i = 1, \ldots, d$. If we use $\mathcal{U}$ to denote the set of all functions $u: \mathcal{P}(D) \mapsto \R$, then we can represent feature attributions as mappings of the form $E: \mathcal{U} \mapsto \R^d$, which we refer to as \textit{explanation mappings}. Other methods take the alternative approach of summarizing set functions with a set $S^* \subseteq D$ of the most influential features. We represent these feature selection summaries as explanation mappings of the form $E: \mathcal{U} \mapsto \mathcal{P}(D)$. Both approaches provide users with simple summaries of a feature's contribution to the set function.

We now consider the specific choices made by each method (see Appendix~\ref{app:methods} for further details). For simplicity, we let $u$ denote the set function each method analyzes. Surveying the various removal-based explanation methods, the techniques for summarizing each feature's influence include:

\begin{itemize}
    \item (Remove individual) Occlusion, PredDiff, CXPlain, permutation tests and feature ablation (LOCO) calculate the impact of removing a single feature from the model, resulting in the following attribution values:
    
    \begin{equation}
        a_i= u(D) - u(D \setminus \{i\}). \label{eq:remove_individual}
    \end{equation}
    
    Occlusion, PredDiff and CXPlain can also be applied with groups of features, or superpixels, in image contexts.
    
    
    \item (Include individual) The univariate predictors approach calculates the impact of including individual features, resulting in the following attribution values:
    
    \begin{equation}
        a_i= u(\{i\}) - u(\{\}). \label{eq:include_individual}
    \end{equation}
    
    This is essentially the reverse of the previous approach: rather than removing individual features from the complete set, this approach adds individual features to the empty set.
    
    \item (Additive model) LIME fits a regularized additive model to a dataset of perturbed examples. In the limit of an infinite number of samples, this process approximates the following attribution values:
    
    \begin{equation}
        a_1, \ldots, a_d = \argmin_{b_0, \ldots, b_d} \; \sum_{S \subseteq D}\pi(S)\Big(b_0 + \sum_{i \in S} b_i - u(S)\Big)^2 + \Omega(b_1, \ldots, b_d). \label{eq:lime}
    \end{equation}
    
    In this problem, $\pi$ represents a weighting kernel and $\Omega$ is a regularization function that is often set to the $\ell_1$ penalty to encourage sparse attributions \citep{tibshirani1996regression}. Since this summary is based on an additive model, the learned coefficients $(a_1, \ldots, a_d)$ represent the incremental value associated with including each feature.
    
    \item (Mean when included) RISE determines feature attributions by sampling many subsets $S \subseteq D$ and then calculating the mean value when a feature is included. Denoting the distribution of subsets as $p(S)$ and the conditional distribution as $p(S \mid i \in S)$, the attribution values are defined as
    
    \begin{equation}
        a_i= \E_{p(S \mid i \in S)}\big[u(S)\big].
    \end{equation}
    
    In practice, RISE samples the subsets $S \subseteq D$ by removing each feature $i$ independently with probability $p$, using $p = 0.5$ in their experiments \citep{petsiuk2018rise}.
    
    \item (Shapley value) Shapley Net Effects, IME, Shapley Effects, QII, SHAP (including KernelSHAP, TreeSHAP and LossSHAP), SPVIM and SAGE all calculate feature attributions using the Shapley value, which we denote as $a_i = \phi_i(u)$. Described in more detail in Section~\ref{sec:cooperative_game_theory}, Shapley values are the only attributions that satisfy several desirable properties.
    
    \item (Low-value subset) MP selects a small set of features $S^*$ that can be removed to give the set function a low value. It does so by solving the following optimization problem:
    
    \begin{equation}
        S^* = \argmin_{S} \; u(D \setminus S) + \lambda |S|. \label{eq:mp_opt}
    \end{equation}
    
    In practice, MP incorporates additional regularizers and solves a relaxed version of this problem (see Section~\ref{sec:complexity}).
    
    \item (High-value subset) MIR solves an optimization problem to select a small set of features $S^*$ that alone can give the set function a high value. For a user-defined minimum value $t$, the problem is given by:

    \begin{equation}
        S^* = \argmin_S \; |S| \quad {\text s.t.} \;\; u(S) \geq t. \label{eq:mir_opt}
    \end{equation}
    
    L2X and EP solve a similar problem but switch the terms in the constraint and optimization objective. For a user-defined subset size $k$, the optimization problem is given by:
    
    \begin{equation}
        S^* = \argmax_S \; u(S) \quad \mathrm{s.t.} \;\; |S| = k. \label{eq:l2x_opt}
    \end{equation}
    
    Finally, INVASE, REAL-X and FIDO-CA solve a regularized version of the problem with a parameter $\lambda > 0$ controlling the trade-off between the subset value and subset size:
    
    \begin{equation}
        S^* = \argmax_S \; u(S) - \lambda |S|. \label{eq:invase_opt}
    \end{equation}
    
    \item (Partitioned subsets) MM solves an optimization problem to partition the features into $S^*$ and $D \setminus S^*$ while maximizing the difference in the set function's values. This approach is based on the idea that removing features to find a low-value subset (as in MP) and retaining features to get a high-value subset (as in MIR, L2X, EP, INVASE, REAL-X and FIDO-CA) are both reasonable approaches for identifying influential features. The problem is given by:
    
    \begin{equation}
        S^* = \argmax_S \; u(S) - \gamma u(D \setminus S) - \lambda |S|. \label{eq:mm_opt}
    \end{equation}
    
    In practice, MM also incorporates regularizers and monotonic link functions to enable a more flexible trade-off between $u(S)$ and $u(D \setminus S)$ (see Appendix~\ref{app:methods}).
\end{itemize}

\begin{figure*}[t]
\vskip -0.1in
\begin{center}
\includegraphics[width=\columnwidth]{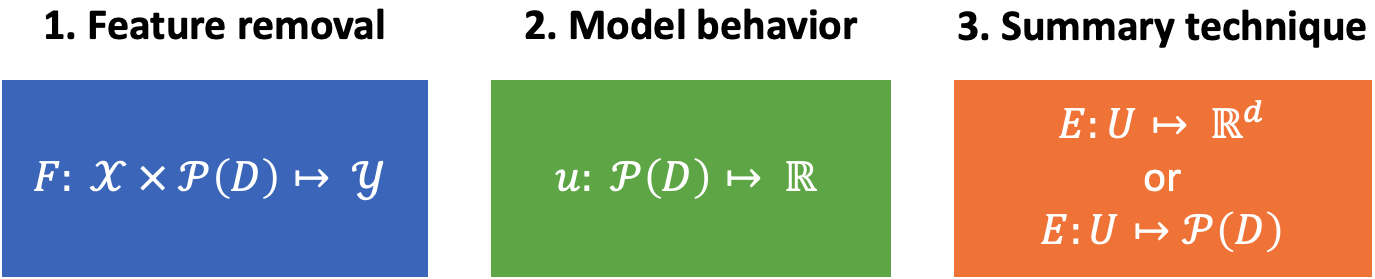}
\caption{Removal-based explanations are specified by three precise mathematical choices: a subset function $F \in \mathfrak{F}$, a set function $u \in \mathcal{U}$, and an explanation mapping $E$ (for feature attribution or selection).}
\label{fig:framework}
\end{center}
\vskip -0.2in
\end{figure*}

As this discussion shows, every removal-based explanation generates summaries of each feature's influence on the underlying set function. In general, a model's dependencies are too complex to communicate fully, so explanations must provide users with a concise summary instead. Feature attributions provide a granular view of each feature's influence on the model, while feature selection summaries can be understood as coarse attributions that assign binary rather than real-valued importance.

Interestingly, if the high-value subset optimization problems solved by MIR, L2X, EP, INVASE, REAL-X and FIDO-CA were applied to the set function that represents the dataset loss (Eq.~\ref{eq:loss_shap_sage}), they would resemble conventional global feature selection problems \citep{guyon2003introduction}. The problem in Eq.~\ref{eq:l2x_opt} determines the set of $k$ features with maximum predictive power, the problem in Eq.~\ref{eq:mir_opt} determines the smallest set of features that achieve the performance given by $t$, and the problem in Eq.~\ref{eq:invase_opt} uses a parameter $\lambda$ to control the trade-off. Though not generally viewed as a model explanation approach, global feature selection serves an identical purpose of identifying highly predictive features.

We conclude by reiterating that the third dimension of our framework amounts to a choice of explanation mapping, which takes the form $E : \mathcal{U} \mapsto \R^d$ for feature attribution or $E: \mathcal{U} \mapsto \mathcal{P}(D)$ for feature selection. Our discussion so far has shown that removal-based explanations can be specified using three precise mathematical choices, as depicted in Figure~\ref{fig:framework}. These methods, which are often presented in ways that make their connections difficult to discern, are constructed in a remarkably similar fashion. The remainder of this work addresses the relationships and trade-offs among these different choices, beginning by examining the computational complexity of each summarization approach.

\subsection{Complexity and approximations} \label{sec:complexity}

Showing how certain explanation methods fit into our framework requires distinguishing between their implicit objectives and the approximations that make them practical. Our presentation of these methods deviates from the original papers, which often focus on details of a method's implementation. We now bridge the gap by describing these methods' computational complexity and the approximations that are sometimes used out of necessity.

The challenge with most of the summarization techniques described above is that they require calculating the underlying set function's value $u(S)$ for many subsets of features. In fact, without making any simplifying assumptions about the model or data distribution, several techniques must examine all $2^d$ subsets of features. This includes the Shapley value, RISE's summarization technique and LIME's additive model. Finding exact solutions to several of the optimization problems (MP, MIR, MM, INVASE, REAL-X, FIDO-CA) also requires examining all subsets of features, and solving the constrained optimization problem (EP, L2X) for $k$ features requires examining $\binom{d}{k}$ subsets, or $\mathcal{O}(2^d d^{-\frac{1}{2}})$ subsets in the worst case.\footnote{This can be seen by applying Stirling's approximation to $\binom{d}{d/2}$ as $d$ becomes large.}

The only approaches with lower computational complexity are those that remove individual features (Occlusion, PredDiff, CXPlain, permutation tests, feature ablation) or include individual features (univariate predictors). These require only one subset per feature, or $d$ total feature subsets.

Many summarization techniques have superpolynomial complexity in $d$, making them intractable for large numbers of features, at least with na\"ive implementations. These methods work in practice due to fast approximation approaches, and in some cases techniques have even been devised to generate real-time explanations. Several strategies that yield fast approximations include:

\begin{itemize}[leftmargin=2pc]
    \item Attribution values that are the expectation of a random variable can be estimated by Monte Carlo approximation. IME \citep{vstrumbelj2010efficient}, Shapley Effects \citep{song2016shapley} and SAGE \citep{covert2020understanding} use sampling strategies to approximate Shapley values, and RISE also estimates its attributions via sampling \citep{petsiuk2018rise}.
    
    \item KernelSHAP, LIME and SPVIM are based on linear regression models fitted to datasets containing an exponential number of datapoints. In practice, these techniques fit models to smaller sampled datasets, which means optimizing an approximate version of their objective function \citep{lundberg2017unified, covert2021improving}.
    
    \item TreeSHAP calculates Shapley values in polynomial time using a dynamic programming algorithm that exploits the structure of tree-based models. Similarly, L-Shapley and C-Shapley exploit the properties of models for structured data to provide fast Shapley value approximations \citep{chen2018shapley}.
    
    \item Several of the feature selection methods (MP, L2X, REAL-X, EP, MM, FIDO-CA) solve continuous relaxations of their discrete optimization problems. While these optimization problems can be solved by representing the set of features $S \subseteq D$ as a mask $m \in \{0, 1\}^d$, these methods instead use a continuous mask variable of the form $m \in [0, 1]^d$. When these methods incorporate a penalty on the subset size $|S|$, they also sometimes use the convex relaxation $||m||_1$.
    
    \item One feature selection method (MIR) uses a greedy optimization algorithm. MIR determines a set of influential features $S \subseteq D$ by iteratively removing groups of features that do not reduce the predicted probability for the correct class.
    
    \item One feature attribution method (CXPlain) and several feature selection methods (L2X, INVASE, REAL-X, MM) generate real-time explanations by learning separate explainer models. CXPlain learns an explainer model using a dataset consisting of manually calculated explanations, which removes the need to iterate over each feature when generating new explanations. L2X learns a model that outputs a set of features (represented by a $k$-hot vector) and INVASE/REAL-X learn similar selector models that can output arbitrary numbers of features. Similarly, MM learns a model that outputs masks of the form $m \in [0, 1]^d$ for images. These techniques can be viewed as \textit{amortized} approaches because they learn models that perform the summarization step in a single forward pass.
\end{itemize}

In conclusion, many methods have developed approximations that enable efficient model explanation, despite sometimes using summarization techniques that are inherently intractable (e.g., Shapley values). Certain techniques are considerably faster than others (i.e., the amortized approaches), and some can trade off computational cost for approximation accuracy \citep{vstrumbelj2010efficient, covert2021improving}, but they are all sufficiently fast to be used in practice.

We speculate, however, that more approaches will be made to run in real-time by learning separate explainer models, as in the MM, L2X, INVASE, CXPlain and REAL-X approaches \citep{dabkowski2017real, chen2018learning, yoon2018invase, schwab2019cxplain, jethani2021have}. Besides these methods, others have been proposed that learn the explanation process either as a component of the original model \citep{fan2017adversarial, taghanaki2019infomask} or as a separate model after training \citep{schulz2020restricting}. Such approaches may be necessary to bypass the need for multiple model evaluations and make removal-based explanations as fast as gradient-based and propagation-based methods.\footnote{While this work was under review, \cite{jethani2021fastshap} introduced an amortized approach for real-time Shapley value estimation.}

\section{Game-Theoretic Explanations} \label{sec:cooperative_game_theory}

The set functions analyzed by removal-based explanations (Section~\ref{sec:behaviors}) can be viewed as \textit{cooperative games}---mathematical objects studied by cooperative game theory. Only a few explanation methods explicitly consider game-theoretic connections, but we show that every method described thus far can be understood through the lens of cooperative game theory.

\subsection{Cooperative game theory background}

Cooperative games are functions of the form $u: \mathcal{P}(D) \mapsto \R$ (i.e., set functions) that describe the value achieved when sets of players $S \subseteq D$ participate in a game. Intuitively, a game might represent the profit made when a particular group of employees chooses to work together. Cooperative game theory research focuses on understanding the properties of different payoffs that can be offered to incentivize participation in the game, as well as predicting which groups of players will ultimately agree to participate.

For this discussion, we use $u$ to denote a cooperative game. To introduce terminology from cooperative game theory, the features $i = 1, 2, \ldots d$ are referred to as \textit{players}, sets of players $S \subseteq D$ are referred to as \textit{coalitions}, and the output $u(S)$ is referred to as the \textit{value} of $S$. Player $i$'s \textit{marginal contribution} to the coalition $S \in D \setminus \{i\}$ is defined as the difference in value $u(S \cup \{i\}) - u(S)$. \textit{Allocations} are vectors $z \in \R^d$ that represent payoffs proposed to each player in return for participating in the game.

Several fundamental concepts in cooperative game theory are related to the properties of allocations: the \textit{core} of a game, a game's \textit{nucleolus}, and its \textit{bargaining sets} are all based on whether players view certain allocations as favorable \citep{narahari2014game}. Perhaps surprisingly, every summarization technique used by removal-based explanations (Section~\ref{sec:explanations}) can be viewed in terms of allocations to players in the underlying game, enabling us to connect these explanation methods to ideas from
the game theory literature.

\subsection{Allocation strategies} \label{sec:solution_concepts}

Several summarization techniques used by removal-based explanation methods are related to \textit{solution concepts}, which in the cooperative game theory context are allocation strategies designed to be fair to the players. If we let $\mathcal{U}$ represent the set of all cooperative games with $d$ players, then solution concepts are represented by mappings of the form $E: \mathcal{U} \mapsto \R^d$, similar to explanation mappings that represent feature attributions (Section~\ref{sec:summarization_strategies}).

We first discuss the Shapley value, which assumes that the \textit{grand coalition} (the coalition containing all players) is participating and distributes the total value in proportion to each player's contributions \citep{shapley1953value}. The Shapley value can be derived axiomatically, and we list several of its properties below; to highlight its many desirable properties, we provide more axioms than are necessary to derive the Shapley value uniquely.
The Shapley values $\phi_i(u) \in \R$ for a game $u$ are the unique allocations that satisfy the following properties:

\begin{itemize}
    \item (Efficiency) The allocations $\phi_1(u), \ldots, \phi_d(u)$ add up to the difference in value between the grand coalition and the empty coalition:
    
    \begin{equation*}
        \sum_{i \in D} \phi_i(u) = u(D) - u(\{\}).
    \end{equation*}
    
    \item (Symmetry) Two players $i, j$ that make equal marginal contributions to all coalitions receive the same allocation:
    
    \begin{equation*}
        u(S \cup \{i\}) = u(S \cup \{j\}) \;\; \forall \;\; S \implies \phi_i(u) = \phi_j(u).
    \end{equation*}
    
    \item (Dummy) A player $i$ that makes zero marginal contribution receives zero allocation:
    
    \begin{equation*}
        u(S \cup \{i\}) = u(S) \;\; \forall \;\; S \implies \phi_i(u) = 0.
    \end{equation*}
    
    \item (Additivity) If we consider two games $u, u'$ and their respective allocations $\phi_i(u)$ and $\phi_i(u')$, then the cooperative game defined as their sum $u + u'$ has allocations defined as the sum of each game's allocations:
    
    \begin{equation*}
        \phi_i(u + u') = \phi_i(u) + \phi_i(u') \;\; \forall \;\; i.
    \end{equation*}
    
    \item (Marginalism) For two games $u, u'$ where all players have identical marginal contributions, the players receive equal allocations:
    
    \begin{equation*}
        u(S \cup \{i\}) - u(S) = u'(S \cup \{i\}) - u'(S) \;\; \forall \;\; (i, S) \implies \phi_i(u) = \phi_i(u') \;\; \forall \;\; i.
    \end{equation*}
\end{itemize}

The Shapley values are the unique allocations that satisfy these properties \citep{shapley1953value, monderer2002variations}, and the expression for each Shapley value $\phi_i(u)$ is

\begin{equation}
    \phi_i(u) = \frac{1}{d} \sum_{S \subseteq D \setminus \{i\}} \binom{d - 1}{|S|}\inv \Big( u(S \cup \{i\}) - u(S) \Big). \label{eq:shapley_expression}
\end{equation}

The Shapley value has found widespread use because of its axiomatic derivation, both within game theory and in other disciplines \citep{aumann1994economic, shorrocks1999decomposition, petrosjan2003time, tarashev2016risk}. In the context of model explanation, Shapley values define each feature's contribution while accounting for complex feature interactions, such as correlations, redundancy, and complementary behavior \citep{lipovetsky2001analysis, vstrumbelj2009explaining, owen2014sobol, datta2016algorithmic, lundberg2017unified, lundberg2020local, covert2020understanding, williamson2020efficient}.

Like the Shapley value, the Banzhaf value attempts to define fair allocations for each player in a cooperative game. It generalizes the Banzhaf power index, which is a technique for measuring the impact of players in the context of voting games \citep{banzhaf1964weighted}. Links between Shapley and Banzhaf values are described by \cite{dubey1979mathematical}, who show that the Shapley value can be understood as an enumeration over all \textit{permutations} of players, while the Banzhaf value can be understood as an enumeration over all \textit{subsets} of players. The expression for each Banzhaf value $\psi_i(u)$ is:

\begin{equation}
    \psi_i(u) = \frac{1}{2^{d - 1}} \sum_{S \subseteq D \setminus \{i\}} \Big(u(S \cup \{i\}) - u(S) \Big). \label{eq:banzhaf}
\end{equation}

The Banzhaf value fails to satisfy the Shapley value's efficiency property, but it can be derived axiomatically by introducing a variation on the efficiency axiom \citep{nowak1997axiomatization}:

\begin{itemize}
    \item (2-Efficiency) If two players $i, j$ merge into a new player $\{i, j\}$ and we re-define the game $u$ as a game $u'$ on the smaller set of players $\big(D \setminus \{i, j\}\big) \cup \big\{\{i, j\}\big\}$, then the Banzhaf values satisfy the following property:
    
    \begin{equation}
        \psi_{\{i, j\}}(u') = \psi_i(u) + \psi_j(u).
    \end{equation}
\end{itemize}

The 2-efficiency property roughly states that credit allocation is immune to the merging of players. The Banzhaf value has multiple interpretations, but the most useful for our purpose is that it represents the difference between the mean value of coalitions that do and do not include the $i$th player when coalitions are chosen uniformly at random. With this perspective, we observe that the RISE summarization technique is closely related to the Banzhaf value: RISE calculates the mean value of coalitions that include $i$, but, unlike the Banzhaf value, it disregards the value of coalitions that do not include $i$. While the RISE technique \citep{petsiuk2018rise} was not motivated by the Banzhaf value, it is unsurprising that such a natural idea has been explored in cooperative game theory.

Both Shapley and Banzhaf values are mathematically appealing because they can be understood as the weighted average of a player's marginal contributions (Eq.~\ref{eq:shapley_expression}-\ref{eq:banzhaf}). This is a stronger version of the marginalism property introduced by Young's axiomatization of the Shapley value \citep{young1985monotonic}, and solution concepts of this form are known as \textit{probabilistic values} \citep{weber1988probabilistic}. Probabilistic values have their own axiomatic characterization: they have been shown to be the unique values that satisfy a specific subset of the Shapley value's properties \citep{monderer2002variations}.

The notion of probabilistic values reveals links with two other solution concepts. The techniques of removing individual players (e.g., Occlusion, PredDiff, CXPlain, permutation tests and feature ablation) and including individual players (e.g., univariate predictors) can also be understood as probabilistic values, although they are simple averages that put all their weight on a single marginal contribution (Eqs.~\ref{eq:remove_individual}-\ref{eq:include_individual}). Unlike the Shapley and Banzhaf values, these methods neglect
complex interactions when quantifying each player's contribution.

The methods discussed thus far (Shapley value, Banzhaf value, removing individual players, including individual players) all satisfy the symmetry, dummy, additivity, and marginalism axioms. What makes the Shapley value unique among these approaches is the efficiency axiom, which lets us view it as a distribution of the grand coalition's value among the players \citep{dubey1979mathematical}. However, whether the
efficiency property or the Banzhaf value's 2-efficiency property is preferable may depend on the use-case.


\subsection{Modeling cooperative games} \label{sec:linear_models}

Unlike the methods discussed so far, LIME provides a flexible approach for summarizing each player's influence. As its summarization technique, LIME fits an additive
model to the cooperative game (Eq.~\ref{eq:lime}), leaving the user the option of specifying a weighting kernel $\pi$ and regularization term $\Omega$ for the weighted least squares objective.

Although fitting a model to a cooperative game seems distinct from the allocation strategies discussed so far, these ideas are in fact intimately connected. Fitting models to cooperative games, including both linear and nonlinear models, has been studied by numerous works in cooperative game theory \citep{charnes1988extremal, hammer1992approximations, grabisch2000equivalent, ding2008formulas, ding2010transforms, marichal2011weighted}, and specific choices for the weighting kernel can yield recognizable attribution values.

If we fit an additive model to the underlying game with no regularization ($\Omega = 0$), then we can identify several weighting kernels that correspond to other summarization techniques:

\begin{itemize}
    \item When we use the weighting kernel $\pi_{\mathrm{Rem}}(S) = \mathbbm{1}(|S| \geq d - 1)$, where $\mathbbm{1}(\cdot)$ is an indicator function, the attribution values are the marginal contributions from removing individual players from the grand coalition, or the values $a_i = u(D) - u(D \setminus \{i\})$. This is the summarization technique used by Occlusion, PredDiff, CXPlain, permutation tests, and feature ablation.
    
    \item When we use the weighting kernel $\pi_{\mathrm{Inc}}(S) = \mathbbm{1}(|S| \leq 1)$, the attribution values are the marginal contributions from adding individual players to the empty coalition, or the values $a_i = u(\{i\}) - u(\{\})$. This is the summarization technique used by the univariate predictors approach.
    
    \item Results from \cite{hammer1992approximations} show that when we use the weighting kernel $\pi_{\mathrm{B}}(S) = 1$, the attribution values are the Banzhaf values $a_i = \psi_i(u)$.
    
    \item Results from \cite{charnes1988extremal} and \cite{lundberg2017unified} show that the attribution values are equal to the Shapley values $a_i = \phi_i(u)$ when we use the weighting kernel $\pi_{\mathrm{Sh}}$, defined as follows:

    \begin{align}
        \pi_{\mathrm{Sh}}(S) = \frac{d - 1}{\binom{d}{|S|}|S|(d - |S|)}.
    \end{align}
\end{itemize}

Although the Shapley value connection has been noted in the model explanation context, the other results we present are new observations about LIME (proofs in Appendix~\ref{app:lime}). These results show that the weighted least squares problem solved by LIME provides sufficient flexibility to yield both the Shapley and Banzhaf values, as well as simpler quantities such as the marginal contributions from removing or including individual players. The additive model approach captures every feature attribution technique discussed so far, with the caveat that RISE uses a modified version of the Banzhaf value. And, like the other allocation strategies, attributions arising from fitting an additive model can be shown to satisfy different sets of Shapley axioms (Appendix~\ref{app:alternative_axioms}).

We have thus demonstrated that the additive model approach for summarizing feature influence not only has precedent in cooperative game theory, but that it is intimately connected to the other allocation strategies. This suggests that the feature attributions generated by many removal-based explanations can be understood as additive decompositions of the underlying cooperative game.

\subsection{Identifying coalitions using excess}

To better understand removal-based explanations that perform feature selection, we look to a more basic concept in cooperative game theory: the notion of \textit{excess}. Excess is defined for a given allocation $z \in \R^d$ and
coalition $S \subseteq D$ as the difference between the coalition's value and its total allocation \citep{narahari2014game}. More formally, it is defined as follows.

\begin{definition}
    Given a cooperative game $u: \mathcal{P}(D) \mapsto \R$, a coalition $S \subseteq D$, and an allocation $z = (z_1, z_2, \ldots, z_d) \in \R^d$, the {\normalfont \textbf{excess}} of $S$ at $z$ is defined as
    
    \begin{equation}
        e(S, z) = u(S) - \sum_{i \in S} z_i.
    \end{equation}
\end{definition}

The excess $e(S, z)$ represents the coalition's degree of unhappiness under the allocation $z$, because an allocation is unfavorable to a coalition if its value $u(S)$ exceeds its cumulative allocation $\sum_{i \in S} z_i$. For cooperative game theory concepts including the core, the nucleolus and bargaining sets, a basic assumption is that coalitions with higher excess (greater unhappiness) are more likely to break off and refuse participation in the game \citep{narahari2014game}. We have no analogue for features refusing participation in ML, but the notion of excess is useful for understanding several removal-based explanation methods.

Below, we show that removal-based explanations that select sets of influential features are equivalent to proposing equal allocations to all players and then determining a high-valued coalition by examining each coalition's excess. Given equal allocations, players with high contributions will be less satisfied than players with low contributions, so solving this problem leads to
a set of high- and low-valued players. We show this by reformulating each method's optimization problem as follows:

\begin{itemize}
    \item (Minimize excess) MP isolates a low-value coalition by finding the coalition with the lowest excess, or the highest satisfaction, given equal allocations $z = \mathbf{1}\lambda$:
    
    \begin{equation}
        S^* = \argmin_S \; e(\bar S, \mathbf{1}\lambda). \label{eq:mp_excess}
    \end{equation}
    
    \noindent The result $S^*$ is a coalition whose complement $D \setminus S^*$ is most satisfied with the equal allocations.

    \item (Maximize excess) MIR isolates the smallest possible coalition that achieves a sufficient level of excess given allocations equal to zero. For a level of excess $t$, the optimization problem is given by:
    
    \begin{equation}
        S^* = \argmin_S \; |S| \quad \mathrm{s.t.} \;\; e(S, \mathbf{0}) \geq t. \label{eq:mir_excess}
    \end{equation}
    
    L2X and EP solve a similar problem but switch the objective and constraint. The optimization problem represents the coalition of size $k$ with the highest excess given allocations of zero to each player:
    
    \begin{equation}
        S^* = \argmax_S \; e(S, \mathbf{0}) \quad \mathrm{s.t.} \;\; |S| = k. \label{eq:l2x_excess}
    \end{equation}
    
    Finally, FIDO-CA, INVASE and REAL-X find the coalition with the highest excess given equal allocations $z = \mathbf{1}\lambda$:
    
    \begin{equation}
        S^* = \argmax_S \; e(S, \mathbf{1}\lambda). \label{eq:invase_excess}
    \end{equation}
    
    \item (Maximize difference in excess) MM combines the previous approaches. Rather than focusing on a coalition with low or high excess, MM partitions players into two coalitions while maximizing the difference in excess given equal allocations $z = \mathbf{1} \frac{\lambda}{1 + \gamma}$:
    
    \begin{equation}
        S^* = \argmax_S \; e\Big(S, \mathbf{1}\frac{\lambda}{1 + \gamma}\Big) - \gamma e\Big(\bar S, \mathbf{1}\frac{\lambda}{1 + \gamma}\Big). \label{eq:mm_excess}
    \end{equation}
    
    \noindent The result $S^*$ is a coalition of dissatisfied players whose complement $D \setminus S^*$ is comparably more satisfied.
    
\end{itemize}

All of the approaches listed above are different formulations of the same multi-objective optimization problem: the intuition is that there is a small set of high-valued players $S$ and a comparably larger set of low-valued players $D \setminus S$. Most methods focus on just one of these coalitions (MP focuses on $D \setminus S$, and MIR, L2X, EP, INVASE, REAL-X and FIDO-CA focus on $S$), while the optimization problem solved by MM considers both coalitions.

MM's summarization technique can therefore be understood as a generalization of the other methods. The MP and INVASE/REAL-X/FIDO-CA problems (Eqs.~\ref{eq:mp_excess},~\ref{eq:invase_excess}), for example, are special cases of the MM problem. The other problems (Eq.~\ref{eq:mir_excess},~\ref{eq:l2x_excess}) cannot necessarily be cast as special cases of the MM problem (Eq.~\ref{eq:mm_excess}), but the MM problem resembles the Lagrangians of these constrained problems; more precisely, a special case of the MM problem shares the same optimal coalition with the dual to these constrained problems \citep{boyd2004convex}.

By reformulating the optimization problems solved by each method, we can see that each feature selection summarization technique can be described as minimizing or maximizing excess given equal allocations for all players. In cooperative game theory, examining each coalition's level of excess (or dissatisfaction) helps determine whether allocations will incentivize participation, and the same tool is used by removal-based explanations to find the most influential features for a model.

These feature selection approaches can be viewed as mappings of the form $E: \mathcal{U} \mapsto \mathcal{P}(D)$ because they identify coalitions $S^* \subseteq D$. Although the Shapley axioms apply only to mappings of the form $E: \mathcal{U} \mapsto \R^d$ (i.e., attribution methods), we find that these feature selection approaches satisfy certain analogous properties (Appendix~\ref{app:alternative_axioms}); however, the properties we identify are insufficient to derive an axiomatically unique method.

\subsection{Summary}

This discussion has shown that every removal-based explanation can be understood using ideas from cooperative game theory. Table~\ref{tab:cooperative} displays our findings regarding each method's game-theoretic interpretation and lists the relevant aspects of the literature for each summarization technique. Under our framework, removal-based explanations are implicitly based on an underlying cooperative game (Section~\ref{sec:behaviors}), and these connections show that the model explanation field has in many cases either reinvented or borrowed ideas that were previously well-understood
in cooperative game theory. We speculate that ongoing model explanation research may benefit from borrowing more ideas from cooperative game theory.

These connections are also important because they help use reason about the advantages of different summarization techniques via the properties that each method satisfies. Among the various techniques, we argue that the Shapley value provides the most complete explanation because it satisfies many desirable properties and gives a granular view of each player's contributions (Section~\ref{sec:solution_concepts}). Unlike the other methods, it divides the grand coalition's value (due to the efficiency property) while capturing the nuances of each player's contributions.

In contrast, the other methods have potential shortcomings, although such issues depend on the use-case. Measuring a single marginal contribution, either by removing or including individual players, ignores player interactions; for example, removing individual players may lead to near-zero attributions for groups of correlated features, even if they are collectively important. The Banzhaf value provides a more nuanced view of each player's contributions, but it cannot be viewed as a division of the grand coalition's value because it violates the efficiency axiom \citep{dubey1979mathematical}. Finally, feature selection explanations provide only a coarse summary of each player's value contribution, and their results depend on user-specified hyperparameters; moreover, these methods are liable to select only one member out of a group of correlated features, which may be misleading to users.

\begin{table}
\caption{Each method's summarization technique can be understood in terms of concepts from cooperative game theory.}
\label{tab:cooperative}
\vskip 0.2in
\begin{center}
\begin{small}
\begin{tabular}{ccc}
\toprule
\textsc{Summarization} & \textsc{Methods} & \textsc{Related To} \\
\midrule
Shapley value & \makecell{Shapley Net Effects, IME, \\QII, SHAP, TreeSHAP, \\KernelSHAP, LossSHAP, \\Shapley Effects, SAGE, SPVIM} & \makecell{Shapley value, \\probabilistic values,\\modeling cooperative games} \\
\midrule
Mean value when included & RISE & \makecell{Banzhaf value, \\probabilistic values, \\modeling cooperative games} \\
\midrule
\makecell{Remove/include individual \\players} & \makecell{Occlusion, PredDiff, CXPlain, \\permutation tests,\\ univariate predictors, \\feature ablation (LOCO)} & \makecell{Probabilistic values, \\modeling cooperative games} \\
\midrule
Fit additive model & LIME & \makecell{Shapley value, Banzhaf value, \\modeling cooperative games} \\
\midrule
High/low value coalitions & \makecell{MP, EP, MIR, MM, \\L2X, INVASE, \\REAL-X, FIDO-CA} & \makecell{Maximum/minimum excess} \\
\bottomrule
\end{tabular}
\end{small}
\end{center}
\vskip -0.2in
\end{table}

\section{Information-Theoretic Explanations} \label{sec:information}

We now examine how removal-based explanations are connected to information theory. We begin by describing how features can be removed using knowledge of the underlying data distribution, and we then show that many feature removal approaches approximate marginalizing out features using their conditional distribution. Finally, we prove that this approach gives every removal-based explanation an information-theoretic interpretation.

\subsection{Removing features consistently} \label{sec:consistency}

There are many ways to remove features from a model (Section~\ref{sec:missingness_strategies}), so we consider how to do so while accounting
for the underlying data distribution. Recall that removal-based explanations evaluate models while withholding groups of features using a subset function $F \in \mathfrak{F}$, which is typically a subset extension of an existing model $f \in \mathcal{F}$ (Definition~\ref{def:extension}). We begin by introducing a specific perspective on how to interpret a subset function's predictions, which are represented by $F(x_S)$.

Supervised ML models typically approximate the response variable's conditional distribution given the input. This is clear for classification models that estimate the conditional probability $f(x) \approx p(Y \mid X = x)$, but it is also true for regression models that estimate the conditional expectation $f(x) \approx \E[Y \mid X = x]$. (These approximations are implicit in conventional loss functions such as cross entropy and MSE.) We propose that a subset function $F \in \mathfrak{F}$ can be viewed equivalently as a conditional probability/expectation estimate given \textit{subsets} of features.

To illustrate this idea in the classification case, we denote the model's estimate as $q(Y \mid X = x) \equiv f(x)$, where the model's output is a discrete probability distribution. Although it is not necessarily equal to the true conditional distribution $p(Y \mid X = x)$, the estimate $q(Y \mid X = x)$ represents a valid probability distribution for all $x \in \mathcal{X}$. Similarly, we denote the subset function's estimate as $q(Y \mid X_S = x_S) \equiv F(x_S)$. A subset function $F$ represents a set of conditional distributions $\big\{q(Y \mid X_S) : S \subseteq D\big\}$, and this interpretation is important because we must consider whether these distributions are probabilistically valid. In particular, we must verify that standard probability laws (non-negativity, unitarity, countable additivity, Bayes rule) are not violated.

As we show below, this cannot be guaranteed. The laws of probability impose a relationship between $q(Y \mid X = x)$ and $q(Y \mid X_S = x_S)$ for any $x \in \mathcal{X}$ and $S \subset D$: they are linked by the underlying distribution on $\mathcal{X}$, or, more specifically, by the conditional distribution $X_{\bar S} \mid X_S = x_S$. In fact, any distribution $q(X)$ implies a unique definition for $q(Y \mid X_S)$ based on $q(Y \mid X)$ due to Bayes rule and the countable additivity property (described below). The flexibility of subset functions regarding how to remove features is therefore problematic, because certain removal approaches do not yield a valid set of conditional distributions.

Constraining the feature removal approach to be probabilistically valid can ensure that the model's subset extension $F$ is faithful both to the original model $f$ and an underlying data distribution. Building on this perspective, we define the notion of \textit{consistency} between a subset function and a data distribution as follows.

\begin{definition} \label{def:consistency}
    A subset function $F \in \mathfrak{F}$ that estimates a random variable $Y$'s conditional distribution is {\normalfont \textbf{consistent}} with a data distribution $q(X)$ if its estimates satisfy the following properties:
    
    \begin{enumerate}
        \item
        (Countable additivity) The probability of the union of a countable number of disjoint events is the sum of their probabilities. Given events $A_1, A_2, \ldots$ such that $A_i \cap A_j = \varnothing$ for $i \neq j$, we have
        
        \begin{equation*}
            P\Big(\bigcup_{i = 1}^\infty A_i\Big) = \sum_{i = 1}^\infty P(A_i).
        \end{equation*}
        
        \item (Bayes rule) The conditional probability $P(A \mid B)$ for events $A$ and $B$ is defined as
        
        \begin{equation*}
            P(A \mid B) = \frac{P(A, B)}{P(B)}.
        \end{equation*}
    \end{enumerate}
\end{definition}

This definition of consistency describes a class of subset functions that obey fundamental probability axioms \citep{laplace1781memoire, andrei1950kolmogorov}.  Restricting a subset function to be consistent does not make its predictions correct, but it makes them compatible with a particular data distribution $q(X)$. Allowing for a distribution $q(X)$ that differs from the true distribution $p(X)$ reveals that certain approaches implicitly assume modified data distributions.

Based on Definition~\ref{def:consistency}, the next two results show that there is a unique subset extension $F \in \mathfrak{F}$ for a model $f \in \mathcal{F}$ that is consistent with a given data distribution $q(X)$ (proofs in Appendix~\ref{app:proofs}). The first result relates to subset extensions of classification models that estimate conditional probabilities.

\begin{proposition} \label{theorem:unique_probability}
    For a classification model $f \in \mathcal{F}$ that estimates a discrete $Y$'s conditional probability, there is a unique subset extension $F \in \mathfrak{F}$ that is consistent with $q(X)$,
    
    \begin{equation*}
        F(x_S) = \E_{q(X_{\bar S} \mid X_S = x_S)}\big[f(x_S, X_{\bar S})\big],
    \end{equation*}
    
    \noindent where $q(X_{\bar S} \mid X_S = x_S)$ is the conditional distribution induced by $q(X)$, i.e., the distribution
    
    \begin{equation*}
        q(x_{\bar S} \mid X_S = x_S) = \frac{q(x_{\bar S}, x_S)}{\int_{X_{\bar S}} q(X_{\bar S}, x_S)}.
    \end{equation*}
\end{proposition}

The next result arrives at a similar conclusion, but it is specific to subset extensions of regression models that estimate the response variable's conditional expectation.

\begin{proposition} \label{theorem:unique_expectation}
    For a regression model $f \in \mathcal{F}$ that estimates a real-valued $Y$'s conditional expectation, there is a unique subset extension $F \in \mathfrak{F}$ that is consistent with $q(X)$,
    
    \begin{equation*}
        F(x_S) = \E_{q(X_{\bar S} \mid X_S = x_S)}\big[f(x_S, X_{\bar S})\big],
    \end{equation*}
    
    \noindent where $q(X_{\bar S} \mid X_S = x_S)$ is the conditional distribution induced by $q(X)$.
\end{proposition}

These results differ in their focus on classification and regression models, but the conclusion in both cases is the same: the only subset extension of a model $f$ that is consistent with $q(X)$ is one that averages the full model output $f(X)$ over the distribution of values $X_{\bar S}$ given by $q(X_{\bar S} \mid X_S)$.

When defining a model's subset extension,
the natural choice is to make it consistent with the true data distribution $p(X)$. This yields precisely the approach presented by SHAP (the conditional version), SAGE, and several other methods, which is to marginalize out the removed features using their conditional distribution $p(X_{\bar S} \mid X_S = x_S)$ \citep{strobl2008conditional, zintgraf2017visualizing, lundberg2017unified, aas2019explaining, covert2020understanding, frye2020shapley}.

Besides this approach, only a few other approaches are consistent with any distribution. The QII approach (Eq.~\ref{eq:qii}) is consistent with a distribution that is the product of marginals, $q(X) = \prod_{i = 1}^d p(X_i)$. The more recent IME approach (Eq.~\ref{eq:ime_uniform}) is consistent with a distribution that is the product of uniform distributions, $q(X) = \prod_{i = 1}^d u_i(X_i)$. And finally, any approach that sets features to default values $r \in \mathcal{X}$ (Eqs.~\ref{eq:zeros}-\ref{eq:default}) is consistent with a distribution that puts all of its mass on those values---which we refer to as a \textit{constant distribution}. These approaches all achieve consistency because they are based on a simplifying assumption of feature independence.

\subsection{Conditional distribution approximations} \label{sec:conditional_approximations}

While few removal-based explanations explicitly suggest marginalizing out features using their conditional distribution, several methods can be understood as approximations of this approach. These represent practical alternatives to the exact conditional distribution, which is unavailable in practice and often difficult to estimate.

The core challenge in using the conditional distribution is modeling it accurately. Methods that use the marginal distribution sample rows from the dataset \citep{breiman2001random, lundberg2017unified}, and it is possible to filter for rows that agree with the features to be conditioned on \citep{sundararajan2019many}; however, this technique does not work well for high-dimensional or continuous-valued data. A relaxed alternative to this approach is using cohorts of rows with similar values \citep{mase2019explaining}.

While properly representing the conditional distribution for every subset of features is challenging, there are several approaches that provide either rough or high-quality approximations. These approaches include:

\begin{itemize}
    \item \textbf{Assume feature independence.} If we assume that the features $X_1, \ldots, X_d$ are independent, then the conditional distribution $p(X_{\bar S} \mid X_S)$ is equivalent to the joint marginal distribution $p(X_{\bar S})$, and it is even equivalent to the product of marginals $\prod_{i \in \bar S} p(X_i)$. The removal approaches used by KernelSHAP and QII can therefore be understood as rough approximations to the conditional distribution that assume feature independence \citep{datta2016algorithmic, lundberg2017unified}.
    
    \item \textbf{Assume model linearity (and feature independence).} As \cite{lundberg2017unified} pointed out, replacing features with their mean can be interpreted as an additional assumption of model linearity:
    
    \begin{align}
        \E\big[f(X) \mid X_S = x_S\big]
        &= \E_{p(X_{\bar S} \mid X_S = x_S)}\big[f(x_S, X_{\bar S})\big] \tag{{\small Conditional distribution}} \\
        &\approx \E_{p(X_{\bar S})}\big[f(x_S, X_{\bar S})\big] \tag{{\small Assume feature independence}} \\
        &\approx f\big(x_S, \E[X_{\bar S}]\big). \tag{{\small Assume model linearity}}
    \end{align}
    
    While this pair of assumptions rarely holds in practice, particularly with the complex models for which explanation methods are designed, it provides some justification for methods that replace features with default values (e.g., LIME, Occlusion, MM, CXPlain, RISE).
    
    \item \textbf{Parametric assumptions.} Recent work on Shapley values has proposed parametric approximations of the conditional distribution, e.g., multivariate Gaussian and copula-based models \citep{aas2019explaining, aas2021explaining}. While the parametric assumptions may not hold exactly, these approaches can provide better conditional distribution approximations than feature independence assumptions.
    
    \item \textbf{Generative model.} FIDO-CA proposes removing features by drawing samples from a conditional generative model \citep{chang2018explaining}. If the generative model $p_G$ (e.g., a conditional GAN) is trained to optimality, then it produces samples from the true conditional distribution. We can then write
    
    \begin{equation}
        p_G(X_{\bar S} \mid X_S) \stackrel{d}{=} p(X_{\bar S} \mid X_S),
    \end{equation}
    
    \noindent where $\stackrel{d}{=}$ denotes equality in distribution. Given a sample $\tilde x_{\bar S} \sim p_G(X_{\bar S} \mid X_S = x_S)$, the prediction $f(x_S, \tilde x_{\bar S})$ can be understood as a single-sample Monte Carlo approximation of the expectation $\E[f(X) \mid X_S = x_S]$. \cite{agarwal2019explaining} substituted the generative model approach into several existing methods (Occlusion, MP, LIME) and observed improvements in their performance across numerous metrics. \cite{frye2020shapley} demonstrated a similar approach using a variational autoencoder-like model \citep{ivanov2018variational}, and future work could leverage other conditional generative models \citep{douglas2017universal, belghazi2019learning}.
    
    \item \textbf{Surrogate model.} Several methods require a surrogate model that is trained to match the original model's predictions given subsets of features, where missing features are represented by zeros (or other values that do not appear in the dataset). This circumvents the task of modeling an exponential number of conditional distributions and is equivalent to parameterizing a subset function $F \in \mathfrak{F}$ and training it with the following objective:
    
    \begin{equation}
        \min_F \; \E_X \E_S \Big[\ell\big(F(X_S), f(X) \big) \Big]. \label{eq:surrogate_loss}
    \end{equation}
    
    In Appendix~\ref{app:surrogate}, we prove that for certain loss functions $\ell$, the subset function $F$ that optimizes this objective is equivalent to marginalizing out features using their conditional distribution. \cite{frye2020shapley} show a similar result for MSE loss, and we also show that a cross entropy loss can be used for classification models. Finally, we find that L2X \citep{chen2018learning} may not provide a faithful approximation because the subsets $S$ are not distributed independently from the inputs $X$---an issue recently addressed by REAL-X \citep{jethani2021have}.
    
    \item \textbf{Missingness during training.} Rather than training a surrogate with missing features, we can instead learn the original model with missingness introduced during training. This is equivalent to parameterizing a subset function $F \in \mathfrak{F}$ and optimizing the following objective, which resembles the standard training loss:
    
    \begin{equation}
        \min_F \; \E_{XY} \E_S \Big[ \ell\big(F(X_S), Y \big) \Big]. \label{eq:missingness_loss}
    \end{equation}
    
    In Appendix~\ref{app:missingness}, we prove that if the model optimizes this objective, then it is equivalent to marginalizing out features from $f(x) \equiv F(x)$ using the conditional distribution. An important aspect of this objective is that the subsets $S$ must be independently distributed from the data $(X, Y)$. INVASE \citep{yoon2018invase} does not satisfy this property, which may prevent it from providing an accurate approximation.
    
    \item \textbf{Separate models.} Finally, Shapley Net Effects \citep{lipovetsky2001analysis}, SPVIM \citep{williamson2020efficient} and the original IME \citep{vstrumbelj2009explaining} propose training separate models for each feature subset, or $\{f_S : S \subseteq D\}$. If every model is optimal, then this approach is equivalent to marginalizing out features with their conditional distribution (Appendix~\ref{app:separate}). This is due to a relationship that arises between models that optimize the population risk for different sets of features; for example, with cross entropy loss, the optimal model (the Bayes classifier) for $X_S$ is given by $f_S(x_S) = p(Y \mid X_S = x_S)$, which is equivalent to $\E[f_D(X) \mid X_S = x_S]$ because the optimal model given all features is $f_D(x) = p(Y \mid X = x)$.

\end{itemize}

This discussion shows that although few methods explicitly suggest removing features with the conditional distribution, numerous methods approximate this approach. Training separate models should provide the best approximation because each model is given a relatively simple prediction task, but this approach is unable to scale to high-dimensional datasets. The generative model approach amortizes knowledge of an exponential number of conditional distributions into a single model, which is more scalable and effective for image data \citep{yu2018generative}; the supervised surrogate and missingness during training approaches also require learning up to one additional model, and these are trained with far simpler optimization objectives (Eq.~\ref{eq:surrogate_loss},~\ref{eq:missingness_loss}) than conditional generative models.

We conclude that, under certain assumptions of feature independence or model optimality, several feature removal strategies are consistent with the data distribution $p(X)$. Our definition of consistency provides a new lens for comparing different feature removal strategies, and Table~\ref{tab:consistency} summarizes our findings.

\begin{table}[t]
\caption{Consistency properties of different feature removal strategies.}
\label{tab:consistency}
\vskip 0.2in
\begin{center}
\begin{small}
\begin{tabular}{ccc}
\toprule
\textsc{Removal} & \textsc{Methods} & \textsc{Consistency} \\
\midrule
Marginalize (conditional) & \makecell{Cond. permutation tests, \\ PredDiff, SHAP, LossSHAP, \\SAGE, Shapley Effects} & Consistent with $p(X)$ \\
\midrule
Generative model & FIDO-CA & \multirow{9}{*}{\makecell{Consistent with $p(X)$\\ (assuming model optimality)}}\\
\cmidrule{1-2}
Supervised surrogate & \makecell{L2X, REAL-X, \\\cite{frye2020shapley}} & \\
\cmidrule{1-2}
Missingness during training & INVASE & \\
\cmidrule{1-2}
Separate models & \makecell{Feature ablation (LOCO), \\univariate predictors, \\Shapley Net Effects, \\SPVIM, IME (2009)} & \\
\midrule
Marginalize (marginal) & \makecell{Permutation tests, \\KernelSHAP} & \multirow{2.5}{*}{\makecell{Consistent with $p(X)$ \\(assuming independence)}} \\
\cmidrule{1-2}
Marginalize (marginals product) & QII & \\
\midrule
Marginalize (marginals product) & QII & \multirow{2.5}{*}{\makecell{Consistent with $q(X)$ \\with feature independence}} \\
\cmidrule{1-2}
Marginalize (uniform) & IME (2010) & \\
\midrule
Zeros & \makecell{Occlusion, PredDiff \\ RISE, CXPlain} & \multirow{2.5}{*}{\makecell{Consistent with constant \\distributions $q(X)$}} \\
\cmidrule{1-2}
Default values & LIME (images), MM & \\
\midrule
Tree distribution & TreeSHAP & \multirow{2.5}{*}{\makecell{Not consistent with \\any $q(X)$}}\\
\cmidrule{1-2}
Extend pixel values & MIR & \\
\midrule
Blurring & MP, EP & \multirow{2.5}{*}{Not valid $F \in \mathfrak{F}$}\\
\cmidrule{1-2}
Marginalize (replacement dist.) & LIME (tabular) & \\
\bottomrule
\end{tabular}
\end{small}
\end{center}
\end{table}

\subsection{Connections with information theory} \label{sec:information_theory}

Conventional wisdom suggests that explanation methods quantify the information contained in each feature. However, we find that precise information-theoretic connections can be identified only when held-out features are marginalized out with their conditional distribution. Our analysis expands on prior work by showing that every removal-based explanation has a probabilistic or information-theoretic interpretation when features are removed properly \citep{owen2014sobol, chen2018learning, covert2020understanding}.

To aide our presentation, we assume that the model $f$ is optimal, i.e., it is the Bayes classifier $f(x) = p(Y \mid X = x)$ for classification tasks or the conditional expectation $f(x) = \E[Y \mid X = x]$ for regression tasks. This assumption is optimistic, but because models are typically trained to approximate one of these functions, the resulting explanations are approximately based on the information-theoretic quantities derived here.

By assuming model optimality and marginalizing out removed features with their conditional distribution, we can guarantee that the prediction given any subset of features is optimal for those features. Specifically, we have the Bayes classifier $F(x_S) = p(Y \mid X_S = x_S)$ in the classification case and the conditional expectation $F(x_S) = \E[Y \mid X_S = x_S]$ in the regression case. Using these subset functions, we can derive probabilistic and information-theoretic interpretations for each explanation method.

These connections focus on the set functions analyzed by each removal-based explanation (Section~\ref{sec:behaviors}). We present results for classification models that use cross entropy loss here, and we show analogous results for regression models in Appendix~\ref{app:games}. Under the assumptions described above, the set functions analyzed by each method can be interpreted as follows:

\begin{itemize}
    \item The set function $u_x(S) = F(x_S)$ represents the response variable's conditional probability for the chosen class $y$:
    
    \begin{equation}
        u_x(S) = p(y \mid X_S = x_S).
    \end{equation}
    
    This lets us examine each feature's true association with the response variable.
    
    \item The set function $v_{xy}(S) = - \ell\big(F(x_S), y\big)$ represents the log probability of the correct class $y$, which is equivalent to the \textit{pointwise mutual information} $\mathrm{I}(y; x_S)$ (up to a constant value):
    
    \begin{equation}
        v_{xy}(S) = \mathrm{I}(y; x_S) + c.
    \end{equation}
    
    This quantifies how much information $x_S$ contains about the outcome $y$, or how much less surprising $y$ is given knowledge of $x_S$ \citep{fano1961transmission}. 
    
    \item The set function $v_x(S) = - \E_{p(Y \mid X = x)}\big[\ell(F(x_S), Y)\big]$ represents the negative Kullback-Leibler (KL) divergence between the label's conditional distribution and its partial conditional distribution (up to a constant value):
    
    \begin{equation}
        v_x(S) = c - \KL\big(p(Y \mid X = x) \; \big|\big| \; p(Y \mid X_S = x_S)\big).
    \end{equation}
    
    As mentioned by \cite{yoon2018invase}, this provides an information-theoretic measure of the deviation between the response variable's true distribution and its distribution when conditioned on a subset of features \citep{cover2012elements}.
    
    \item The set function $v(S) = - \E_{XY}\big[\ell\big(F(X_S), Y\big)\big]$ represents the \textit{mutual information} with the response variable (up to a constant value):
    
    \begin{equation}
        v(S) = \mathrm{I}(Y ; X_S) + c.
    \end{equation}
    
    As discussed by \cite{covert2020understanding}, this quantifies the amount of information, or the predictive power, that the features $X_S$ communicate about the response variable $Y$ \citep{cover2012elements}.
    
    \item The set function $w_x(S) = - \ell\big(F(x_S), f(x)\big)$ represents the KL divergence between the full model output and its output given a subset of features. Specifically, if we define $Z$ to be a categorical random variable $Z \sim \mathrm{Cat}\big(f(X)\big)$, then we have:
    
    
    \begin{equation}
        w_x(S) = c - \KL\big( p(Z \mid X = x) \; \big|\big| \; p(Z \mid X_S = x_S) \big).
    \end{equation}
    
    This result does not require model optimality, but under the assumption that $f$ is the Bayes classifier, this quantity is equivalent to the KL divergence between the label's conditional and partial conditional distribution (up to a constant value):
    
    
    \begin{equation}
        w_x(S) = c - \KL\big( p(Y \mid X = x) \; \big|\big| \; p(Y \mid X_S = x_S) \big).
    \end{equation}
    
    We can therefore see that under these assumptions, L2X, INVASE and REAL-X are based on the same set function and have a similar information-theoretic interpretation \citep{chen2018learning, yoon2018invase, jethani2021have}.
    
    \item The set function $w(S) = - \E_{XY}\big[\ell\big(F(X_S), f(X)\big)\big]$ represents the information that $X_S$ communicates about the model output $f(X)$. Specifically, if we let $Z \sim \mathrm{Cat}\big(f(X)\big)$, then we have:
    
    \begin{equation}
        w(S) = \mathrm{I}\big(Z ; X_S\big) + c.
    \end{equation}
    
    This is the classification version of the conditional variance decomposition provided by Shapley Effects \citep{owen2014sobol}. This result does not require model optimality, but if $f$ is the Bayes classifier, then this is equivalent to the mutual information with the response variable (up to a constant value):
    
    \begin{align}
        w(S) = \mathrm{I}(Y ; X_S) + c.
    \end{align}
\end{itemize}


\begin{table}[t]
\caption{Each method's underlying set function has an information-theoretic interpretation when features are removed appropriately.}
\label{tab:information}
\begin{center}
\begin{small}
\begin{tabular}{cccc}
\toprule
\textsc{Model Behavior} & \textsc{Set Function} & \textsc{Methods} & \textsc{Related To} \\
\midrule
Prediction & $u_x$ & \makecell{Occlusion, MIR, MM, \\IME, QII, LIME, MP, EP, \\ FIDO-CA, RISE, SHAP, \\KernelSHAP, TreeSHAP} & \makecell{Conditional probability, \\conditional expectation} \\
\midrule
Prediction loss & $v_{xy}$ & LossSHAP, CXPlain & \makecell{Pointwise mutual \\information} \\
\midrule
Prediction mean loss & $v_x$ & INVASE & \makecell{KL divergence with \\conditional distribution} \\
\midrule
Dataset loss & $v$ & \makecell{Permutation tests, \\univariate predictors, \\feature ablation (LOCO), \\Shapley Net Effects, \\SAGE, SPVIM} & \makecell{Mutual information \\(with label)} \\
\midrule
Prediction loss (output) & $w_x$ & L2X, REAL-X & \makecell{KL divergence with \\ full model output}\\
\midrule
Dataset loss (output) & $w$ & Shapley Effects & \makecell{Mutual information \\(with output)} \\
\bottomrule
\end{tabular}
\end{small}
\end{center}
\end{table}

As noted, two assumptions are required to derive these results. The first is that features are marginalized out using the conditional distribution; although many methods use different removal approaches, they can be modified to use this approach or an approximation (Section~\ref{sec:conditional_approximations}). The second is that models are optimal; this assumption rarely holds in practice, but since conventional loss functions train models to approximate either the Bayes classifier or the conditional expectation, we can view these information-theoretic quantities as the values that each set function approximates.

We conclude that when features are removed appropriately, explanation methods quantify the information communicated by each feature (see summary in Table~\ref{tab:information}). No single set function provides the ``right'' approach to model explanation; rather, these information-theoretic quantities span a range of perspectives that could be useful for understanding a complex ML model.

Removal-based explanations that are consistent with the observed data distribution can provide well-grounded insight into intrinsic statistical relationships in the data, and this is useful for finding hidden model influences (e.g., detecting bias from sensitive attributes) or when using ML as a tool to discover real-world relationships. However, this approach has the potentially undesirable property that features may appear important even if they are not used by the model in a functional sense \citep{merrick2019explanation, chen2020true}. When users are more interested in the model's mechanism for calculating predictions, other removal approaches may be preferable, such as interventional approaches motivated by a causal analysis of the model \citep{janzing2019feature, heskes2020causal}.

\section{A Cognitive Perspective on Removal-Based Explanations} \label{sec:psychology}

While the previous sections provide a mathematical perspective on removal-based explanations, we now consider this class of methods through a different lens: that of the social sciences. Analyzing this broad class of methods provides an opportunity to discuss how they all relate to cognitive psychology due to their shared reliance on feature removal.

Model explanation tools are not typically designed based on research from the social sciences \citep{miller2017explainable}, but, as we show, removal-based explanations have clear parallels with cognitive theories about how people understand causality. We first discuss our framework's foundation in counterfactual reasoning and then describe a trade-off between simple explanations and those that convey richer information about models.

\subsection{Subtractive counterfactual reasoning}

Explaining a model's predictions is fundamentally a causality question: \textit{what makes the model behave this way?} Each input feature is a potential cause, multiple features may be causal, and explanations should quantify each feature's degree of influence on the model. We emphasize the distinction between this model-focused causality and causality between the input and response (e.g., whether a feature causes the outcome) because real-world causality is difficult to discern from observational data \citep{pearl2009causality}.

In philosophy and psychology, \textit{counterfactual reasoning} is a standard tool for understanding causality. A counterfactual example changes certain facts of a situation (e.g., the route a person drove to get home) to potentially achieve a different outcome (e.g., getting home safely), and counterfactuals shed light on whether each aspect of a situation caused the actual outcome (e.g., a fatal car crash). In an influential philosophical account of causality, John Stuart Mill presented five methods of induction that use counterfactual reasoning to explain cause-effect relationships \citep{mill1884system, mackie1974cement}. In the psychology literature, counterfactual thinking is the basis of multiple theories about how people explain the causes of events \citep{tversky1982simulation, hilton1990conversational}.

Removal-based explanations perform a specific type of counterfactual reasoning. In psychology, the process of removing an event to understand its influence on an outcome is called a \textit{subtractive counterfactual} \citep{epstude2008functional}, and this is precisely how removal-based explanations work. In philosophy, the same principle is called the \textit{method of difference}, and it is one of Mill's five methods for inferring cause-effect relationships \citep{mill1884system}. This type of logic is also found in cognitive theories about how people understand and discuss causality \citep{tversky1982simulation, jaspars1983attribution, hilton1990conversational}.

The principle of removing something to examine its influence is pervasive in social sciences, not only as a philosophical approach but as part of descriptive psychological theories; this explains the remarkable prevalence of the feature removal principle in model explanation, even among computational researchers who were not explicitly inspired by psychology research. Perhaps surprisingly, the reliance on subtractive counterfactual reasoning (or equivalently, the method of difference) has been overlooked thus far, even by work that examined the psychological basis for SHAP \citep{merrick2019explanation, kumar2020problems}.

Some prior work applies a different form of counterfactual reasoning to model explanation \citep{verma2020counterfactual}. For example, one influential approach suggests showing users counterfactuals that adjust a small number of features to change the model output \citep{wachter2017counterfactual}. This approach undoubtedly provides information about how a model works, but many such counterfactuals are required to fully illustrate each feature's influence. Removal-based explanations can provide more insight by concisely summarizing the results of many subtractive counterfactuals, e.g., via Shapley values.

\subsection{Norm theory and the downhill rule}

Subtractive counterfactuals are an intuitive way to understand each feature's influence, but their implementation is not straightforward. Removal-based explanations aim to remove the information that a feature communicates, or subtract the fact that it was observed, but it is not obvious how to do this: given an input $x \in \mathcal{X}$ and subset $S \subseteq D$, it is unclear how to retain $x_S$ while removing $x_{\bar S}$. We consult two psychological theories to contextualize the approaches that have been considered by different methods.

Many removal-based explanations remove features by averaging the model output over a distribution of possible values for those features; this has a clear correspondence with the cognitive model described by \textit{Norm theory} \citep{kahneman1986norm}. According to this theory, people assess normality by gathering summary statistics from a set of recalled and simulated representations of a phenomenon (e.g., loan application outcomes for individuals with a set of characteristics). In these representations, certain features are fixed (or \textit{immutable}) while others are allowed to vary (\textit{mutable}); in our case these correspond to the retained and removed features, respectively.

Taking inspiration from Norm theory, we may equate a model's behavior when certain features are blocked from exerting influence (i.e., the removed features) with a ``normal'' outcome for the remaining features. With this perspective, we can see that Norm theory provides a cognitive justification for averaging the model output over a distribution of values for the removed features. \cite{merrick2019explanation} make a similar observation to justify how SHAP removes features.

The choice of distribution for averaging outcomes is important, but Norm theory does not prescribe a specific approach. Rather, Norm theory is a descriptive cognitive model that recognizes that individuals may have different perspectives of normality based on their experiences \citep{kahneman1986norm}. Future model explanation research may consider how to tailor explanations to each user, as suggested by \cite{miller2019explanation}, but we also require systematic methods that do not solicit user input. We therefore consider whether any approach used by existing methods is justifiable from a cognitive perspective.

For guidance on the choice of distribution, we look to research on human tendencies when assigning blame. In their study of \textit{mental undoing}, \cite{tversky1982simulation} examined people's biases when proposing counterfactuals that change an undesirable event's outcome. Their clearest finding was the \textit{downhill rule}, which states that people are more likely to propose changes that remove a surprising aspect of a story or otherwise increase the story's internal coherence. In other words, people are more likely to assign blame to an aspect of a situation if it has a more likely alternative that would change the outcome.

One feature removal strategy is reminiscent of the downhill rule because it considers alternative values in proportion to their plausibility: when marginalizing out removed features using their conditional distribution, alternative values and their corresponding outcomes are averaged in proportion to the coherence of the full feature set, which is represented by the data distribution $p(X_{\bar S} \mid X_S) \propto p(X)$. This is consistent with the downhill rule because if certain high-likelihood values change the outcome, their influence on the model will be apparent when integrating $f(X)$ over the distribution $p(X \mid X_S = x_S)$.

In summary, Norm theory provides a cognitive analogue for removing features by averaging over a distribution of alternative values, and the downhill rule suggests that the plausibility or likelihood of alternative values should be taken into account. These theories provide cognitive justification for certain approaches to removing features, but our review of relevant psychology research is far from comprehensive. However, interestingly, our findings lend support to the same approach that yields connections with information theory (Section~\ref{sec:information}), which is marginalizing out features using their conditional distribution.

\subsection{Simplicity versus completeness}

Building on our discussion of the human psychology aspect of removal-based explanations, we now discuss a trade-off between the amount of information conveyed by an explanation and the user's likelihood of drawing the correct conclusions. We describe this trade-off and then show that this class of methods provides the flexibility to balance these competing objectives.

Consider two explanation strategies with different levels of complexity. A counterfactual example that perturbs several features to change the model's prediction is easy to understand, but it does not convey detailed information about a model's dependence on each feature \citep{wachter2017counterfactual}. By contrast, SHAP's feature attributions provide a complex summary of each feature's influence by quantifying the impact of removing different groups of features. Perhaps due to their greater complexity, a recent user study showed that users were less likely to understand SHAP visualizations, experiencing a higher cognitive load than users who viewed simpler explanations \citep{kaur2020interpreting}. Similarly, a different study found that longer explanations required more time and effort to understand, in some cases impeding a user's ability to draw appropriate conclusions \citep{lage2019human}.

\begin{figure*}[t]
\vskip 0.2in
\begin{center}
\includegraphics[width=\columnwidth]{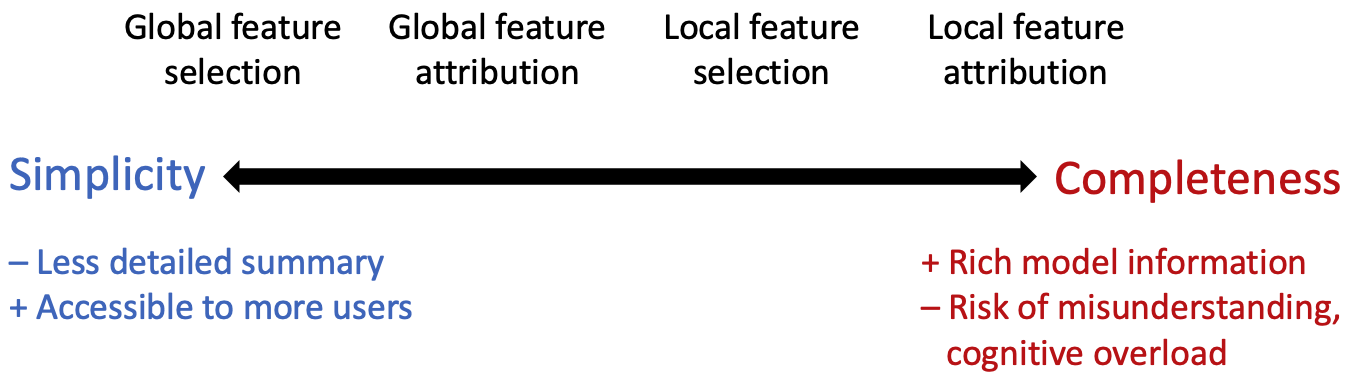}
\caption{Each model explanation strategy represents a trade-off between an explanation's simplicity and its completeness.}
\label{fig:tradeoff}
\end{center}
\vskip -0.2in
\end{figure*}

We view this as a trade-off between simplicity and completeness, because explanations are typically more complex when they provide more information about a model. Providing a more complete characterization may be preferable for helping users build a mental picture of how a model works, but complicated explanations also risk overloading users, potentially leading to a false or limited sense of model comprehension.

Recognizing this trade-off and its implications, we consider simplicity as a design goal and
find that removal-based explanations have the flexibility to adjust the balance between these goals. To reduce cognitive burden, one might focus on global rather than local explanations, e.g., by providing visualizations that display many local explanations \citep{lundberg2020local}, or by using global methods that summarize a model's behavior across the entire dataset \citep{owen2014sobol, covert2020understanding}. Alternatively, one may generate explanations that operate on feature groups rather than individual features, e.g., sets of correlated features or nearby pixels \citep{zeiler2014visualizing, ribeiro2016should}.

Certain explanation formats convey richer information than others (Figure~\ref{fig:tradeoff}). For example, feature attributions provide a granular view of a model by considering every feature as a cause and quantifying each feature's influence, and local explanations are more granular than global ones.
Richer information may not always be desirable, because psychology research shows that people report higher satisfaction with explanations that cite fewer causes \citep{thagard1989explanatory, read1993explanatory, lombrozo2007simplicity, miller2019explanation}. Users may therefore derive more insight from explanations that highlight fewer causes, such as the sparse feature attributions provided by LIME \citep{ribeiro2016should}.

Feature selection explanations offer the potential to go even further in the direction of simplicity. These explanations directly penalize the number of selected features (Section~\ref{sec:summarization_strategies}), guaranteeing that fewer features are labeled as important; and furthermore, they omit information about the granularity of each feature's influence. For a non-technical user, it may be simpler to understand that a model's prediction was dominated by a small number of highly informative features, whereas it may require a more sophisticated user to interpret real-valued attributions for individual features, e.g., as coefficients in an additive decomposition of a model's behavior (Section~\ref{sec:linear_models}).

Put simply, an explanation that paints an incomplete picture of a model may prove more useful if the end-users are able to understand it properly. Designers of explanation methods should be wary of overestimating people's abilities to store complex mental models \citep{norman1983some}, and the ML community can be mindful of this by tailoring explanations to users' degrees of sophistication.

\section{Experiments} \label{sec:experiments}

We have thus far analyzed removal-based explanations from a primarily theoretical standpoint, so we now conduct experiments to provide a complementary empirical perspective. Our experiments aim to accomplish three goals:

\begin{enumerate}
    \item Implement and compare many new methods by filling out the space of removal-based explanations (Figure~\ref{fig:methods_grid}).

    \item Demonstrate the advantages of removing features by marginalizing them out using their conditional distribution---an approach that we showed yields information-theoretic explanations (Section~\ref{sec:information}).

    \item Verify the existence of relationships between various explanation methods. Specifically, explanations may be similar if they use (i)~summary techniques that are probabilistic values of the same cooperative game (Section~\ref{sec:cooperative_game_theory}), or (ii)~feature removal strategies that are approximately equivalent (Section~\ref{sec:conditional_approximations}).
\end{enumerate}

To cover a wide range of methods, we consider many combinations of removal strategies (Section~\ref{sec:removal}), as well as model behaviors (Section~\ref{sec:behaviors}) and summary techniques (Section~\ref{sec:explanations}). Our implementation is available online,\footnote{\url{https://github.com/iancovert/removal-explanations}} and we tested 80 total methods (68 of which are new) that span our framework as follows:

\begin{itemize}
    \item For \textit{feature removal}, we considered replacing features with default values, and marginalizing out features using either (i) uniform distributions, (ii) the product of marginals, or (iii) the joint marginal distribution. Next, to approximate marginalizing out features using their conditional distribution, we trained surrogate models to match the original model's predictions with held-out features (Appendix~\ref{app:surrogate}). Finally, for one dataset we also trained separate models with all feature subsets.
    
    \item Our experiments analyze three \textit{model behaviors} using three different datasets. We explained individual classification probabilities for the census income dataset \citep{lichman2013uci}, the model's loss on individual predictions for MNIST \citep{lecun2010mnist}, and the dataset loss for a breast cancer classification task \citep{berger2018comprehensive}.
    
    \item For \textit{summary techniques}, we considered removing or including individual features, the mean when included strategy, and Banzhaf and Shapley values. For techniques that involve an exponential number of feature subsets, we used sampling-based approximations similar to those previously used for Shapley values \citep{vstrumbelj2010efficient}, and we detected convergence based on the width of confidence intervals.\footnote{We follow the approach from \cite{covert2020understanding}, which identifies convergence via the ratio of confidence interval width to the range in values.}
\end{itemize}

Implementing and comparing all combinations of these choices helps us better understand the relationships between methods and identify the most promising approaches. For more details about the models, datasets and hyperparameters, see Appendix~\ref{app:experiments}.


\subsection{Census income}

The census income dataset provides basic demographic information about individuals, and the task is to predict whether a person's annual income exceeds \$50k. We trained a LightGBM model \citep{ke2017lightgbm} and then generated explanations using all the combinations of removal and summary strategies described above, including training separate models for all feature subsets (as there are only 12 features). When following the default values removal strategy, we used the mean for continuous features and the mode for discrete ones.

\begin{figure}
\vskip 0.2in
\begin{center}
\includegraphics[width=\columnwidth]{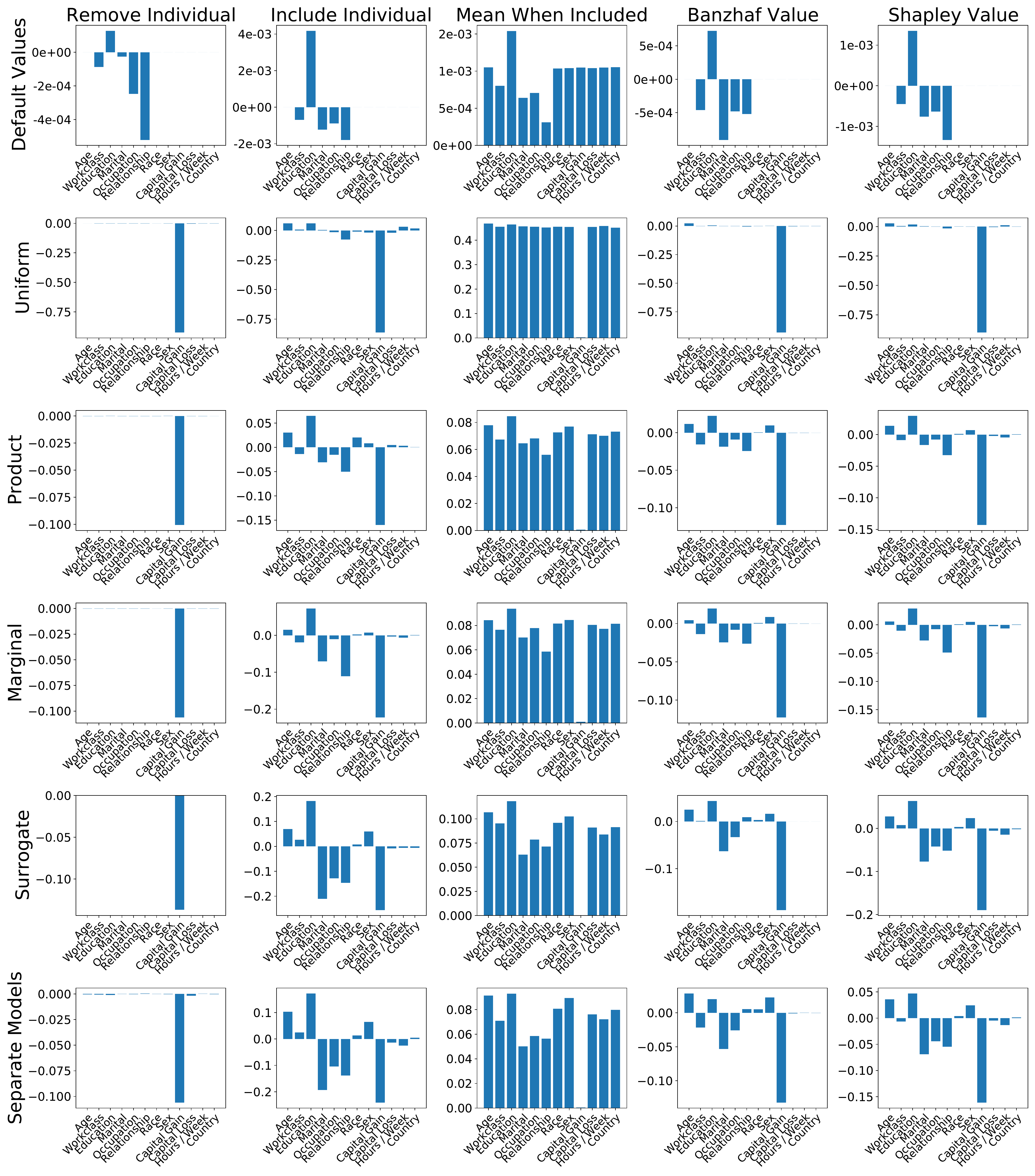}
\caption{Explanations for a single example in the census income dataset. Each bar chart represents attribution values for a different explanation method. The vertical axis represents feature removal strategies and the horizontal axis represents summary techniques.}
\label{fig:census_grid}
\end{center}
\vskip -0.2in
\end{figure}

These combinations of choices resulted in 30 distinct explanation methods, several of which are nearly equivalent to existing approaches (SHAP, QII, IME, Occlusion, RISE, PredDiff),\footnote{There are some implementation differences, e.g., our conditional distribution approximation for PredDiff.} but most of which are new. Figure~\ref{fig:census_grid} shows a grid of explanations generated for a single person whose income did not exceed \$50k. We offer two qualitative remarks about the relationships between these explanations:

\begin{enumerate}
    \item The explanations are sometimes similar despite using different feature removal strategies (see the bottom four rows of Figure~\ref{fig:census_grid}). This is most likely because these removal approaches all approximate the conditional distribution (product, marginal, surrogate, separate models). The first two rows (default values, uniform) deviate from the others because they offer low-quality conditional distribution approximations.
    
    \item The explanations are sometimes similar despite using different summary techniques (see the columns for include individual, Banzhaf and Shapley values). The similarity between probabilistic values suggests that each feature's marginal contributions are largely similar, at least outside of a saturating regime; the remove individual technique deviates from the others, possibly due to saturation effects when most features are included. In contrast, the mean when included technique differs significantly from the others because it is not a probabilistic value (in the game-theoretic sense).
\end{enumerate}

\begin{figure}[t] 
\vskip -0.2in
\begin{center}
\includegraphics[trim=0.8cm 0cm 0cm 0.5cm, clip=true, width=\columnwidth]{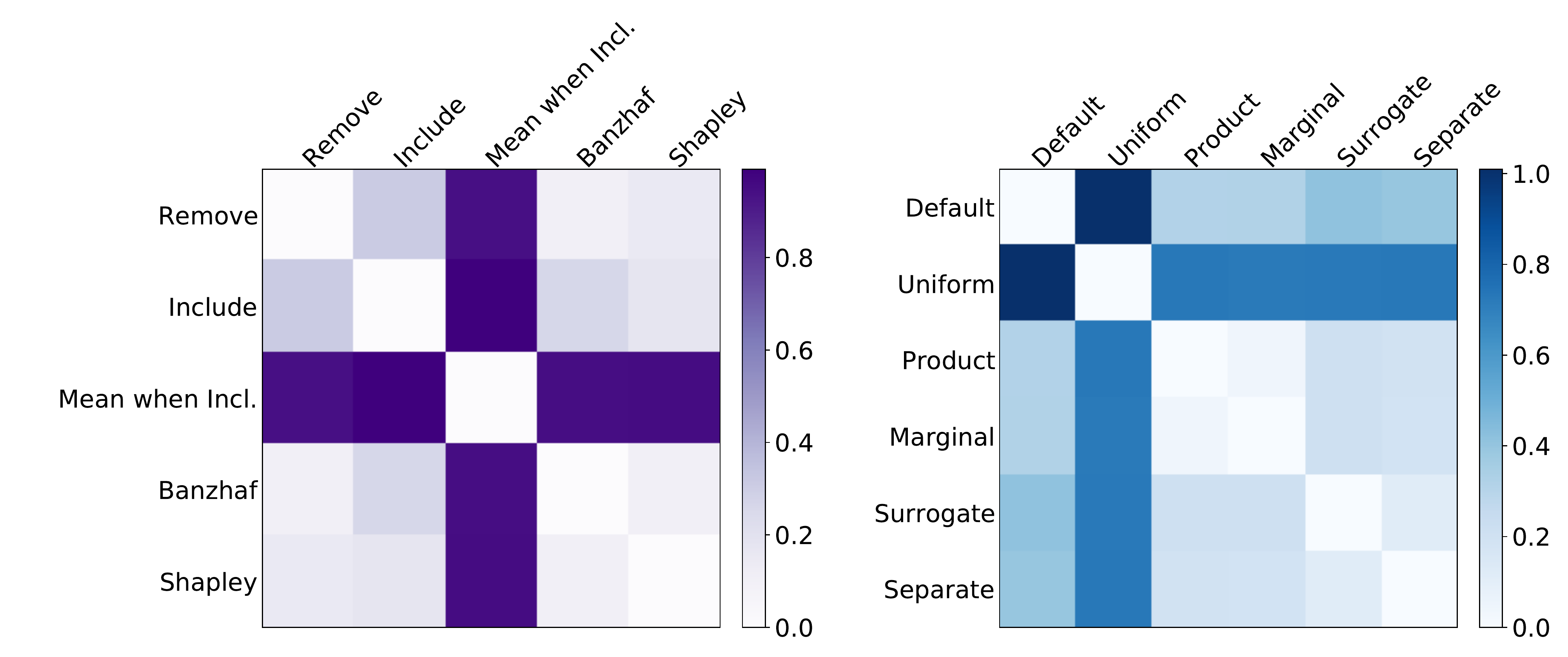}
\caption{Census income explanation differences. The colored entries in each plot represent the mean Euclidean distance between explanations that differ in only their summary technique (left) or feature removal strategy (right), so lighter entries indicate greater similarity.}
\label{fig:census_mse}
\end{center}
\vskip -0.2in
\end{figure}



\begin{figure}[t]
\begin{center}
\includegraphics[trim=1.5cm 0.5cm 1.5cm 1.0cm, clip=true, width=\textwidth]{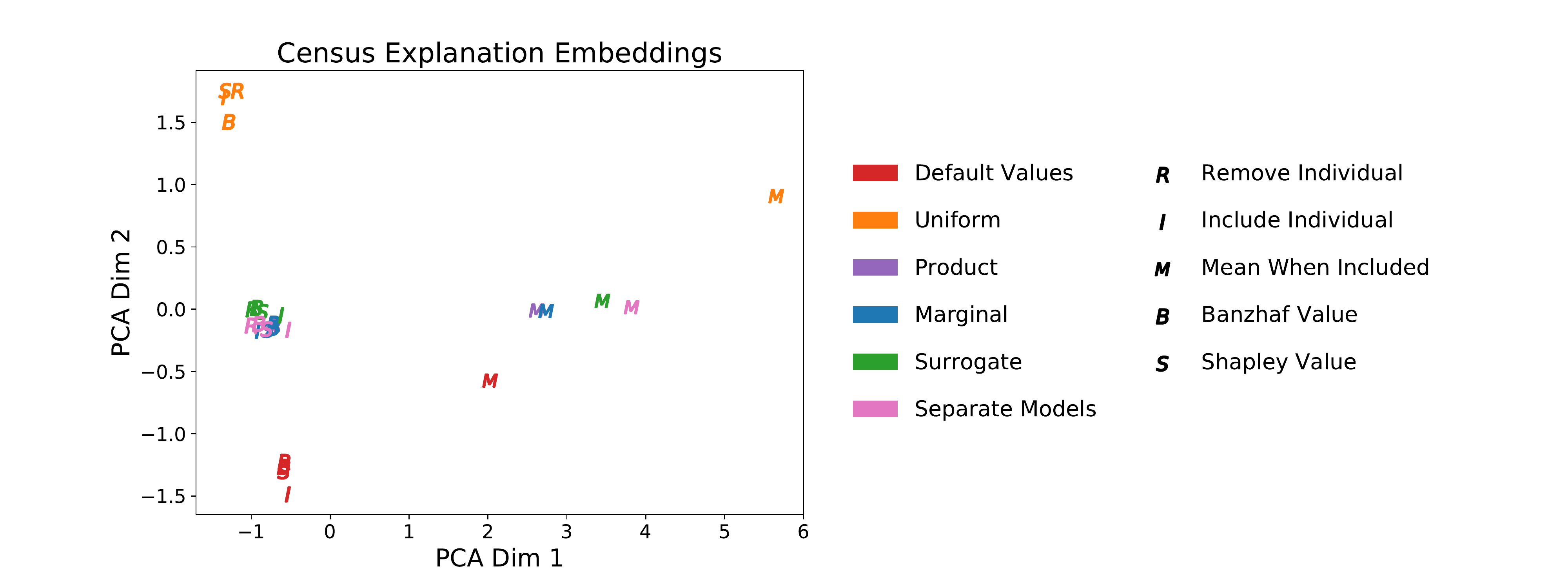}
\caption{Census income explanation embeddings. PCA was used to generate two-dimensional embeddings for each method, allowing us to observe which methods tend to produce similar results.}
\label{fig:census_embedding}
\end{center}
\vskip -0.2in
\end{figure}

To verify whether these relationships hold in general, we scaled up our comparison by generating explanations for 256 instances and quantifying the similarity between methods. We calculated the mean Euclidean distance between explanations generated by each method (averaged across the instances), and, rather than displaying the pair-wise distances for all 30 methods, we only considered comparisons between methods that differ just in their summary techniques or just in their removal strategies.

Figure~\ref{fig:census_mse} shows the results, with the distances between explanations grouped by either summary technique (left) or by feature removal strategy (right). Regarding the summary techniques, the clearest finding is that the mean when included approach produces very different explanations than all the other techniques. Among the remaining ones, Shapley values are relatively close to Banzhaf values, and they are similarly close to explanations that remove or include individual features; this matches the Shapley value's formulation, because like the Banzhaf value, it is a weighted average of all marginal contributions.

Regarding the different feature removal strategies (Figure~\ref{fig:census_mse} right), we observe that the default values and uniform marginalization strategies produce very different explanations than the other approaches. The remaining approaches are relatively similar; for example, the product of marginals and joint marginal produce nearly identical explanations, which suggests that the feature dependencies either are not strong in this dataset or are not captured by our model. We also observe that the surrogate approach is closest to training separate models, as we would expect, but that they are not identical.



\begin{figure}[t]
\begin{center}
\centerline{\includegraphics[width=\columnwidth]{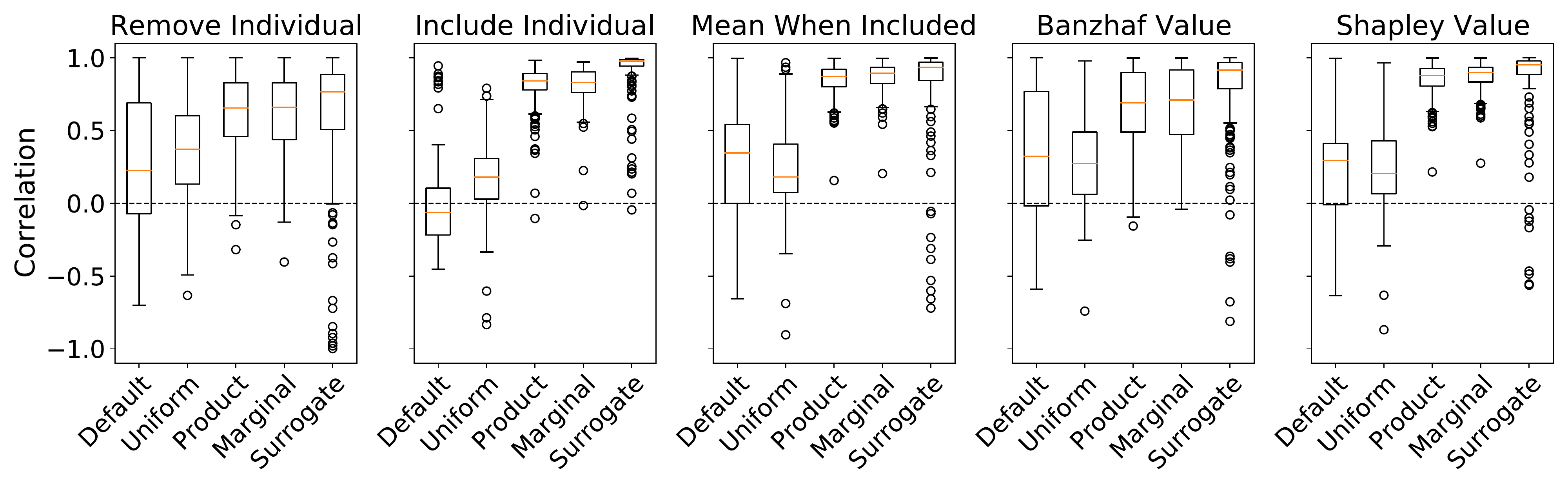}}
\centerline{\includegraphics[width=\columnwidth]{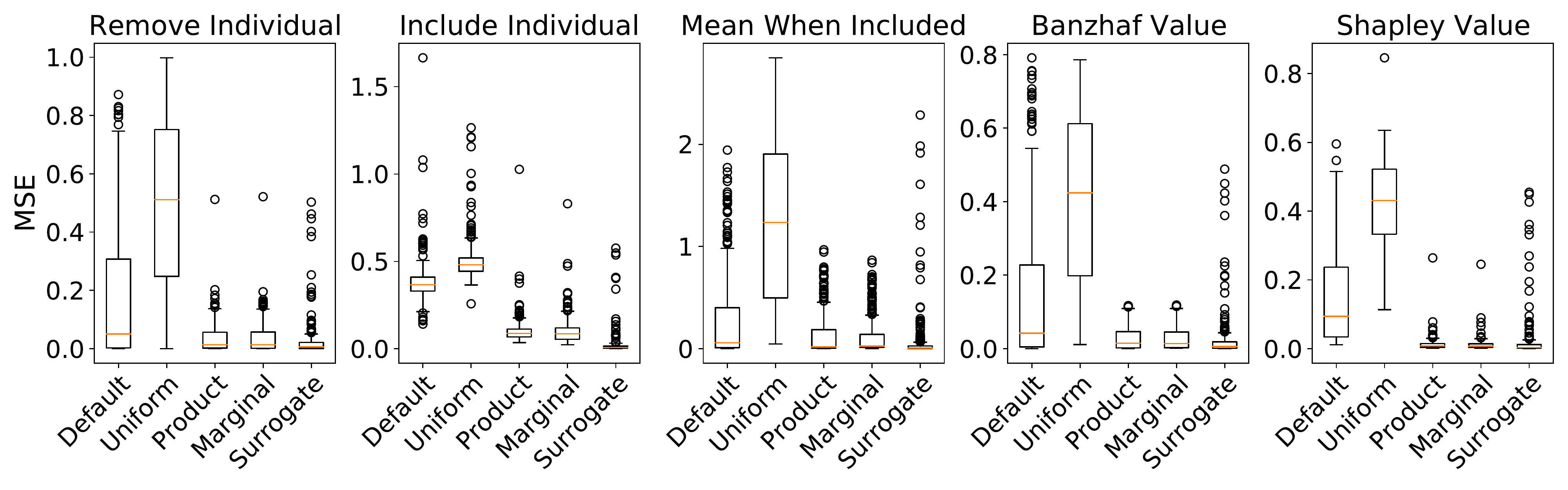}}
\caption{Boxplots quantifying the similarity of census income explanations to comparable explanations generated using separate models. The similarity metrics are correlation (top, higher is better) and MSE (bottom, lower is better).}
\label{fig:census_correctness}
\end{center}
\vskip -0.2in
\end{figure}

Next, we visualized the different explanations using low-dimensional embeddings (Figure~\ref{fig:census_embedding}). We generated embeddings by applying principal components analysis (PCA, \citealp{jolliffe1986principal}) to vectors containing the concatenated explanations for all 256 explanations (i.e., 30 vectors of size 3,072). The results further support our observation that the mean when included technique differs most from the others. The explanations whose feature removal strategy approximates the conditional distribution are strongly clustered, with the product of marginals and joint marginal approaches overlapping almost entirely.

Finally, since many feature removal strategies can be understood as approximating the conditional distribution approach (Section~\ref{sec:conditional_approximations}), we attempted to quantify the approximation quality. We cannot measure this exactly because we lack access to the conditional distributions, so we considered the explanations generated with separate models to be our ground truth, because this is our most reliable proxy. To measure each method's faithfulness to the underlying data distribution, we grouped explanations by their summary technique (e.g., Banzhaf value, Shapley value) and quantified their similarity to explanations generated with separate models (Figure~\ref{fig:census_correctness}).

We calculated two similarity metrics, MSE and correlation, and both show that the surrogate approach is closest to training separate models. This reflects the fact that both approaches provide flexible approximations to marginalizing out features using their conditional distribution (Section~\ref{sec:conditional_approximations}). The default and uniform approaches tend to produce very different explanations. The product and marginal approaches are sometimes competitive with the surrogate approach, but they are noticeably less similar in most cases. We remark, however, that the surrogate approach has more outliers; this suggests that although it is closest to using separate models on average, the surrogate approach is prone to occasionally making large errors.

In sum, these results show that the surrogate approach provides a reliable proxy for the conditional distribution, producing explanations that are faithful to the underlying data distribution. Unlike the separate models approach, the surrogate approach scales to high-dimensional datasets because it requires training just one additional model. And in comparison to other approaches that marginalize out features (uniform, product, marginal), the surrogate approach produces relatively fast explanations because it does not require a sampling-based approximation (i.e., considering multiple values for held-out features) when evaluating the prediction for each feature subset.

\begin{figure}
\vskip -0.2in
\begin{center}
\includegraphics[width=\columnwidth]{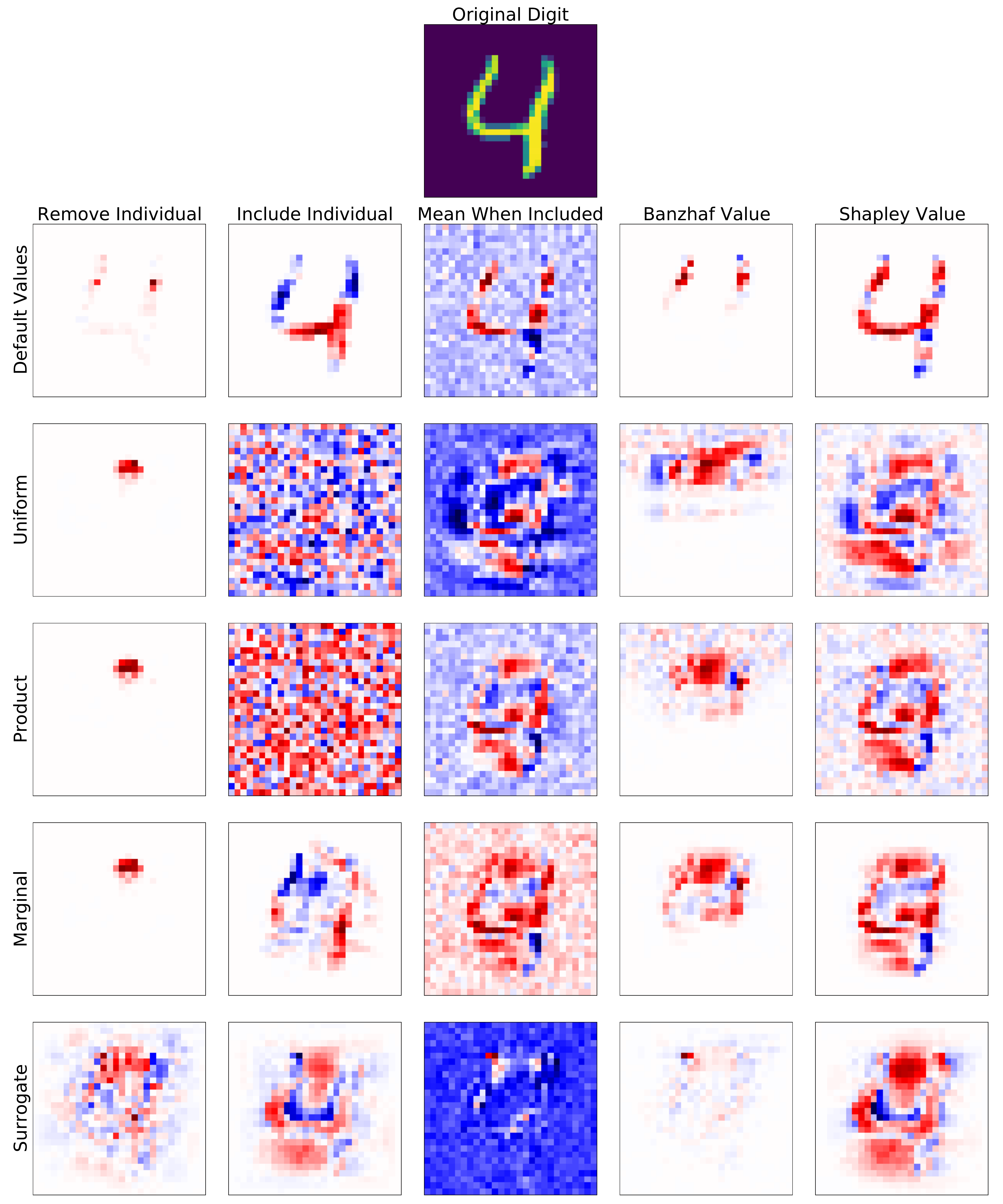}
\caption{Loss explanations for a single MNIST digit. Each heatmap represents attribution values for a different explanation method, with red (blue) pixels improving (hurting) the loss, except for explanations that use the mean when included technique. The vertical axis represents feature removal strategies and the horizontal axis represents summary techniques.}
\label{fig:mnist_grid}
\end{center}
\vskip -0.2in
\end{figure}

\subsection{MNIST}

For the MNIST digit recognition dataset, we trained a 14-layer CNN and generated explanations for the model's loss rather than its classification probabilities. We used zeros as default values, and, as in the previous experiment, we trained a surrogate model to approximate the conditional distribution.

Combining different removal and summary strategies resulted in 25 explanation methods, only two of which corresponded to existing approaches (LossSHAP, and CXPlain without the amortized explainer model). Using these methods, we first generated a grid of explanations for a single example in the dataset (Figure~\ref{fig:mnist_grid}). More so than in the previous experiment, we now observe significant differences between explanations generated by each method. We make the following qualitative observations about these explanations:

\begin{enumerate}
    \item The empty region at the top of the digit should be relevant because it clearly distinguishes fours from nines. This is successfully highlighted by most, but not all methods.

    \item Two removal strategies (uniform, product) frequently produce low-quality, noisy explanations. In contrast, the default value explanations are noiseless, but zero pixels always receive zero attribution because removing them does not impact the model; this is not ideal, because zero pixels can be highly informative.
    
    \item The mean when included technique is difficult to visualize because its attributions are not marginal contributions in the game-theoretic sense, where positive (negative) values improve (hurt) the model's loss.
    
    \item When using the surrogate approach, the Shapley value explanation (Figure~\ref{fig:mnist_grid} bottom right) is most similar to the one that includes individual features. In contrast, removing individual features has a negligible impact on the model, leading to significant noise artifacts (Figure~\ref{fig:mnist_grid} bottom left). This suggests that removing individual features, which is far more common than including individual features (see Table~\ref{tab:methods}), can be incompatible with close approximations of the conditional distribution, likely due to strong feature correlations.
\end{enumerate}

Overall, the most visually appealing explanation is the one produced by the surrogate and Shapley value combination (Figure~\ref{fig:mnist_grid} bottom right); this method roughly corresponds to LossSHAP \citep{lundberg2020local}, although it uses the surrogate approach as a conditional distribution approximation. For the digit displayed in Figure~\ref{fig:mnist_grid}, the explanation highlights two regions at the top and bottom that distinguish the four from an eight or a nine; it has minimal noise artifacts; and it indicates that the unusual curvature on the left side of the four may hurt the model's loss, which is highlighted by only a few other explanations. Appendix~\ref{app:experiments} shows more LossSHAP explanations on MNIST digits.

To avoid the pitfalls of a purely qualitative analysis, we also evaluate these explanations using quantitative metrics. Many works have proposed techniques for evaluating explanation methods \citep{cao2015look, ancona2017towards, petsiuk2018rise, zhang2018top, adebayo2018sanity, hooker2018evaluating, lundberg2020local, jethani2021have}, including sanity checks and measures of correctness. Among these, several consider human-assigned definitions of importance (e.g., the pointing game), while others focus on the model's prediction mechanism. Most of the model-focused metrics involve removing features and measuring their impact on the model's predictions \citep{ancona2017towards, petsiuk2018rise, hooker2018evaluating, lundberg2020local, jethani2021have}, an approach that we explore here.

\begin{figure}[t]
\begin{center}
\includegraphics[trim=1.2cm 0.4cm 0.5cm 0.5cm, clip=true, width=\textwidth]{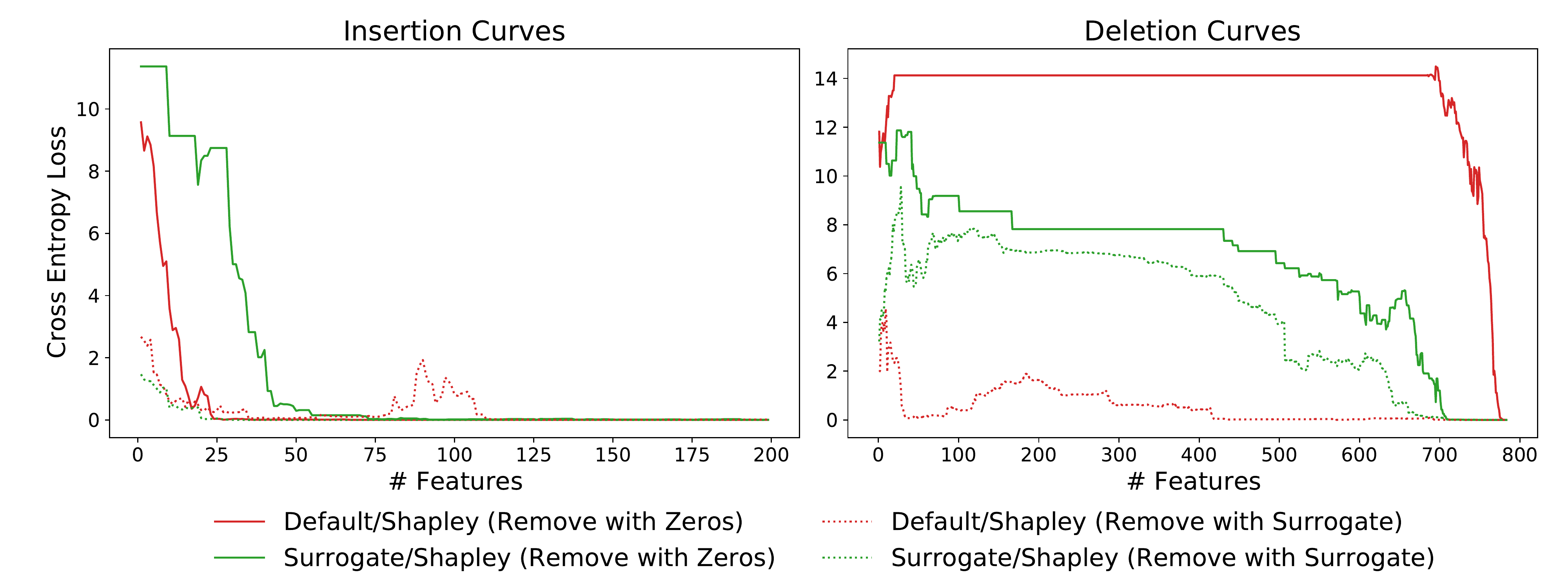}

\caption{Insertion and deletion curves for a single MNIST digit. Curves are displayed for two explanations (Default/Shapley and Surrogate/Shapley) and two feature removal approaches (Zeros, Surrogate). The insertion curves are only shown up to 200 features due to near-zero loss.}
\label{fig:mnist_curves}
\end{center}
\vskip -0.2in
\end{figure}

Evaluation metrics that rely on removing, masking or deleting features have a clear connection with our framework: they mirror the process that removal-based methods follow when generating explanations. With this perspective, we can see that seemingly neutral quantitative metrics may implicitly favor explanations that share their choices of how to remove features from the model.
For example, the sensitivity-$n$ metric \citep{ancona2017towards} evaluates the model's predictions when features are replaced with zeros, which favors methods that use this approach when generating explanations (e.g., Occlusion, RISE, CXPlain). Similarly, ROAR \citep{hooker2018evaluating} measures the decrease in model accuracy after training a new model, which favors methods that account for model retraining.

Rather than relying on a single evaluation metric, we demonstrate how we can create metrics that implicitly favor certain explanations. Following the \textit{insertion} and \textit{deletion} metrics from \cite{petsiuk2018rise}, we used explanations from each method to produce feature rankings, and we iteratively removed either the most important (deletion) or least important (insertion) features to observe the impact on the model. We evaluated the model's loss for each feature subset, tracing a curve whose area we then measured; we focus on the loss rather than the predictions because our MNIST explanations are based on the model's loss. Critically, we handled removed features using three different strategies: 1)~replacing them with zeros, 2)~using a surrogate model (Section~\ref{sec:conditional_approximations}), and 3)~using a new model trained with missing features.
(Note that this choice is made separately from the explanation method.)

\begin{table}[t]
\caption{Insertion and deletion scores for MNIST explanations when masking removed features with zeros. Results for the favored removal strategy are italicized, and the best scores are bolded (accounting for 95\% confidence intervals).}
\label{tab:mnist_zeros}
\vskip -0.2in
\begin{center}
\begin{small}
\begin{tabular}{lcccccccccc}
\toprule
& \multicolumn{5}{c}{Insertion (lower is better)} & \multicolumn{5}{c}{Deletion (higher is better)} \\
\cmidrule(lr){2-6} \cmidrule(lr){7-11}
& RI & II & MWI & BV & SV & RI & II & MWI & BV & SV \\
\midrule
\textit{Default} & \textit{0.683} & \textit{0.478} & \textit{\textbf{0.131}} & \textit{\textbf{0.176}} & \textit{\textbf{0.134}} & \textit{4.797} & \textit{\textbf{8.356}} & \textit{\textbf{9.508}} & \textit{5.636} & \textit{\textbf{10.286}} \\
Uniform & 1.523 & 1.188 & 0.427 & 0.791 & 0.424 & 4.121 & 2.325 & 4.708 & 4.105 & 5.003 \\
Product & 1.204 & 1.017 & 0.204 & 0.588 & 0.224 & 4.316 & 2.308 & 5.428 & 4.415 & 5.611 \\
Marginal & 1.208 & 1.038 & \textbf{0.126} & 0.335 & \textbf{0.125} & 4.289 & 3.726 & 6.213 & 4.895 & 6.797 \\
Surrogate & 1.745 & 0.479 & 0.516 & 0.803 & 0.347 & 1.486 & 3.239 & 2.913 & 1.943 & 4.443 \\
\bottomrule
\end{tabular}
\end{small}
\end{center}
\vskip -0.2in
\end{table}

\begin{table}[t]
\caption{Insertion and deletion scores for MNIST explanations when masking removed features with a surrogate model. Results for the favored removal strategy are italicized, and the best scores are bolded (accounting for 95\% confidence intervals).}
\label{tab:mnist_surrogate}
\begin{center}
\begin{small}
\begin{tabular}{lcccccccccc}
\toprule
& \multicolumn{5}{c}{Insertion (lower is better)} & \multicolumn{5}{c}{Deletion (higher is better)} \\
\cmidrule(lr){2-6} \cmidrule(lr){7-11}
& RI & II & MWI & BV & SV & RI & II & MWI & BV & SV \\
\midrule
Default & 0.085 & 0.152 & 0.053 & 0.096 & 0.100 & 0.403 & 0.638 & 0.325 & 0.296 & 0.466 \\
Uniform & 0.086 & 0.080 & 0.036 & 0.057 & 0.038 & 0.807 & 0.200 & 0.836 & 0.682 & 0.720 \\
Product & 0.068 & 0.067 & 0.027 & 0.046 & 0.027 & 0.873 & 0.179 & 0.934 & 0.777 & 0.888 \\
Marginal & 0.067 & 0.071 & 0.025 & 0.033 & 0.025 & 0.875 & 0.189 & 0.966 & 1.029 & 1.451 \\
\textit{Surrogate} & \textit{0.044} & \textit{\textbf{0.016}} & \textit{0.032} & \textit{0.033} & \textit{\textbf{0.014}} & \textit{0.426} & \textit{1.059} & \textit{1.716} & \textit{0.864} & \textit{\textbf{3.151}} \\
\bottomrule
\end{tabular}
\end{small}
\end{center}
\vskip -0.2in
\end{table}

\begin{table}[t]
\caption{Insertion and deletion scores for MNIST explanations when masking removed features using a model trained with missingness. The best scores for each metric are bolded (accounting for 95\% confidence intervals).}
\label{tab:mnist_missingness}
\vskip -0.2in
\begin{center}
\begin{small}
\begin{tabular}{lcccccccccc}
\toprule
& \multicolumn{5}{c}{Insertion (lower is better)} & \multicolumn{5}{c}{Deletion (higher is better)} \\
\cmidrule(lr){2-6} \cmidrule(lr){7-11}
& RI & II & MWI & BV & SV & RI & II & MWI & BV & SV \\
\midrule
Default & 0.060 & 0.105 & 0.043 & 0.054 & 0.054 & 0.449 & 0.757 & 0.477 & 0.353 & 0.690 \\
Uniform & 0.112 & 0.074 & 0.034 & 0.052 & 0.034 & 0.971 & 0.220 & 0.893 & 0.871 & 0.833 \\
Product & 0.079 & 0.068 & 0.024 & 0.044 & 0.024 & 1.080 & 0.191 & 1.189 & 0.933 & 1.060 \\
Marginal & 0.079 & 0.063 & 0.023 & 0.035 & 0.023 & 1.054 & 0.181 & 1.262 & 1.327 & 1.781 \\
Surrogate & 0.051 & \textbf{0.011} & 0.035 & 0.037 & \textbf{0.015} & 0.300 & 1.189 & 1.418 & 0.651 & \textbf{3.728} \\
\bottomrule
\end{tabular}
\end{small}
\end{center}
\end{table}

Based on this approach, Figure~\ref{fig:mnist_curves} shows examples of insertion and deletion curves for a single MNIST digit. We show curves for only two explanation methods---default values and the surrogate approach, both with Shapley values---and we use two masking strategies to illustrate how each one is advantageous for a particular method. When our evaluation replaces missing features with zeros, the explanation that uses default values produces a better (lower) insertion curve and a better (higher) deletion curve. When we switch to removing features using the surrogate model, the results are reversed, with the explanation that uses the surrogate model performing significantly better. This experiment demonstrates how evaluation metrics can be aligned with the explanation method.

Next, we performed a similar analysis with 100 MNIST digits: we generated insertion and deletion curves for each digit and measured the mean area under the curve (normalized by the number of features). We considered all combinations of removal strategies and summary techniques, and we generated results separately for the three versions of our insertion/deletion metrics (Tables~\ref{tab:mnist_zeros}-\ref{tab:mnist_missingness}). The results confirm our previous analysis: replacing missing features with zeros favors the default values approach (Table~\ref{tab:mnist_zeros}), with the corresponding explanations achieving the best insertion and deletion scores; similarly, handling them using a surrogate model favors the surrogate model approach (Table~\ref{tab:mnist_surrogate}). Interestingly, explanations that use the Shapley value produce the best results in both cases; we understand this as a consequence of the metrics checking subsets of all cardinalities, which matches the Shapley value's formulation (Section~\ref{sec:solution_concepts}).

Evaluating held out features using a separate model trained with missingness does not explicitly favor any explanation method, because
it does not mirror the process by which any explanations are generated. Rather, it can be understood as measuring the information content of the features identified by each method, independently of the original model's prediction mechanism.
However, we find that the results in Table~\ref{tab:mnist_missingness} are very close to those in Table~\ref{tab:mnist_surrogate}, with the surrogate approach outperforming the other removal strategies by a significant margin. We understand this as a consequence of the relationship between the surrogate and the model trained with missingness: although the explanation and evaluation metric are based on different models, both models approximate removing features from the Bayes classifier using the conditional distribution (Section~\ref{sec:conditional_approximations}). According to this metric, the method that performs best is also the surrogate model and Shapley value combination, or LossSHAP \citep{lundberg2020local}---the same method that provided the most visually appealing explanations above.

Measuring an explanation's faithfulness to a model is challenging because it requires making a choice for how to accommodate missing features. As our experiments show, the metrics proposed by recent work implicitly favor methods that align with how the metric is calculated. While using a separate model trained with missingness is less obviously aligned with any explanation method, it still favors methods that approximate the conditional distribution; and furthermore, it measures information content rather than faithfulness to the original model. Users must be wary of the hidden biases in available metrics, and they should rely on evaluations that are most relevant to their intended use-case.


\subsection{Breast cancer subtype classification}

In our final experiment, we analyzed gene microarray data from The Cancer Genome Atlas (TCGA)\footnote{\url{https://www.cancer.gov/tcga}} for breast cancer (BRCA) patients whose tumors were categorized into different molecular subtypes \citep{berger2018comprehensive}. Due to the small dataset size (only 510 patients), we prevented overfitting by analyzing a random subset of 100 genes (details in Appendix~\ref{app:experiments}) and training a regularized logistic regression model. Rather than explaining individual predictions, we explained the dataset loss to determine which genes contain the most information about BRCA subtypes.

\begin{figure}
\begin{center}
\includegraphics[width=\columnwidth]{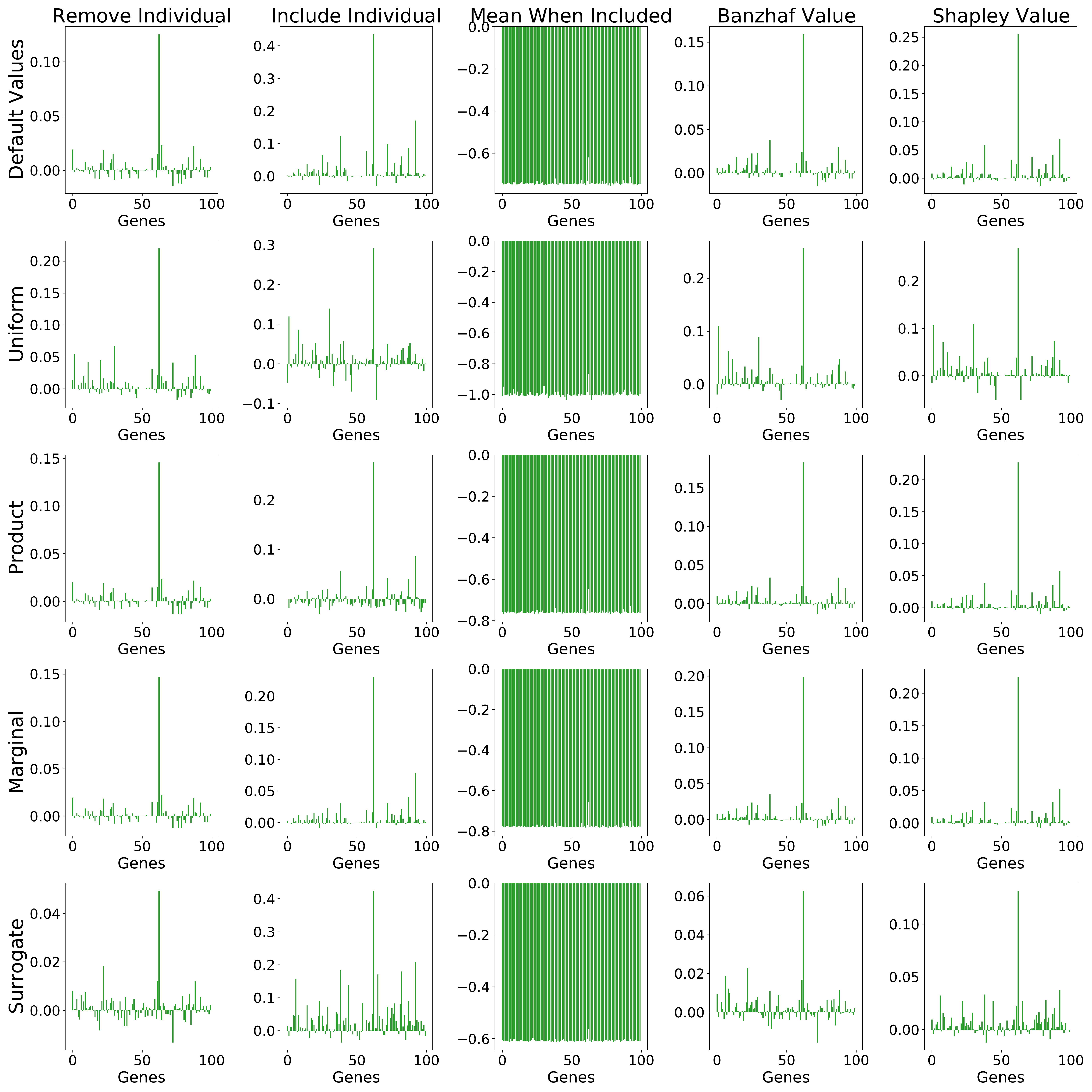}
\caption{Dataset loss explanations for BRCA subtype classification. Each bar chart represents gene attribution values for a different explanation method, with positive (negative) values improving (hurting) the dataset loss, except for the mean when included technique. The vertical axis represents feature removal strategies and the horizontal axis represents summary techniques.
}
\label{fig:brca_grid}
\end{center}
\vskip -0.2in
\end{figure}

We used the same removal and summary strategies as in the previous experiments, including using the mean expression as default values and training a surrogate model to approximate the conditional distribution. Figure~\ref{fig:brca_grid} shows a grid of explanations generated by each method. Three of these explanations correspond to existing approaches (permutation tests, conditional permutation tests, SAGE), but many are new methods. We observe that most explanations identify the same gene as being most important (ESR1), but, besides this, the explanations are difficult to compare qualitatively. In Appendix~\ref{app:experiments}, we measure the similarity between each of these explanations, finding that there are often similarities across removal strategies
and sometimes across summary techniques.

\begin{figure}
\vskip -0.1in
\begin{center}
\includegraphics[width=\columnwidth]{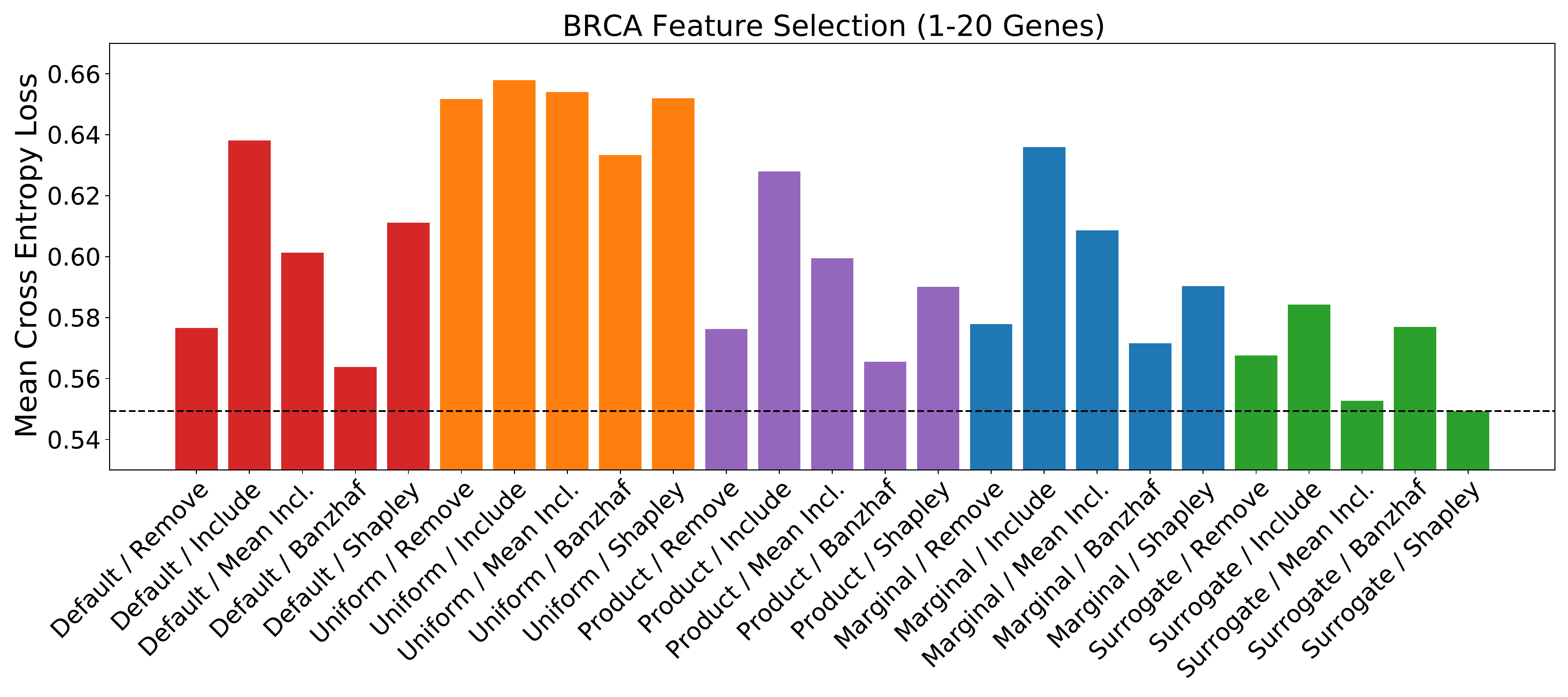}
\caption{Feature selection results for BRCA subtype classification when using top genes identified by each global explanation. Each bar represents the average loss for models trained using 1-20 top genes (lower is better).}
\label{fig:brca_selection}
\end{center}
\vskip -0.2in
\end{figure}

For a quantitative evaluation, we assessed each method by training new models with the top-ranked genes. The best performing method for the top $n$ genes may differ for different values of $n$, so we trained separate models with the top $n = 1, 2, \ldots, 20$ genes and averaged their performance (Figure~\ref{fig:brca_selection}). This is similar to the insertion metric \citep{petsiuk2018rise} used above, except that we measure the dataset loss and focusing on subsets with relatively small cardinalities. The results represent each method's usefulness for performing global feature selection.

According to Figure~\ref{fig:brca_selection}, the surrogate and Shapley value combination performs best overall; this method roughly corresponds to SAGE \citep{covert2020understanding}.\footnote{The SAGE authors suggested marginalizing out features with their conditional distribution, but the surrogate model approximation had not been developed at the time of publication \citep{covert2020understanding}.}
The mean when included summary performs comparably well, and the results are generally better when using the surrogate rather than any other feature removal approach. Including individual features performs poorly, possibly because it neglects feature interactions; and removing features with uniform distributions consistently yields poor results.

We performed a literature review for the most important genes identified by the surrogate and Shapley value combination (SAGE), and we found that many had documented BRCA associations, including ESR1 \citep{robinson2013activating}, CCNB2 \citep{shubbar2013elevated} and TXNL4B \citep{nordgard2008genome}. Other highly ranked genes had known associations with other cancers, including DDC \citep{koutalellis2012dopa} and GSS \citep{kim2015involvement}. We did not evaluate explanation methods based their ability to identify true associations, however, due to the ambiguity in verifying an association from the literature. Beyond confirming known relationships, we remark that these global explanations can also be used to generate new scientific hypotheses to be tested in a lab setting.


\section{Discussion} \label{sec:discussion}

In this work, we developed a unified framework that encompasses a significant portion of the model explanation field, including 26 existing methods. These methods vary in numerous ways and represent disparate parts of the explainability literature, but our framework systematizes them
by showing that each method is specified by three precise mathematical choices:

\begin{enumerate}
    \item \textbf{How the method removes features.} Each method specifies a subset function $F \in \mathfrak{F}$ to make predictions with subsets of features, often based on an existing model $f \in \mathcal{F}$ trained using all the features.
    
    \item \textbf{What model behavior the method analyzes.} Each method relies on a set function $u: \mathcal{P}(D) \mapsto \R$ to represent the model's dependence on different groups of features. The set function describes the model's behavior either for an individual prediction (local explanations) or across the entire dataset (global explanations).

    \item \textbf{How the method summarizes each feature's influence.} Each method generates explanations that provide a concise summary of the features' contributions to the set function $u \in \mathcal{U}$. Mappings of the form $E: \mathcal{U} \mapsto \R^d$ generate feature attribution explanations, and mappings of the form $E: \mathcal{U} \mapsto \mathcal{P}(D)$ generate feature selection explanations.
\end{enumerate}

Our framework reveals that the field is highly interconnected: we find that many state-of-the-art methods are built on the same foundation, are based on different combinations of interchangeable choices, and in many cases differ along only one or two dimensions (recall Table~\ref{tab:common_combinations}). This perspective provides a systematic characterization of the literature and makes it easier to reason about the conceptual and computational advantages of different approaches.

To shed light on the relationships and trade-offs between different methods, we explored perspectives from three related fields that have been overlooked by most explainability research: cooperative game theory (Section~\ref{sec:cooperative_game_theory}), information theory (Section~\ref{sec:information}) and cognitive psychology (Section~\ref{sec:psychology}). We found that removal-based explanations are a simple application of \textit{subtractive counterfactual reasoning}, or, equivalently, of Mill's \textit{method of difference}; and we showed that most explanation methods can be interpreted using existing ideas in cooperative game theory and information theory, in many cases leading to a richer understanding of how each method should be interpreted.

Consulting the game theory literature helped us draw connections between several approaches that can be viewed as \textit{probabilistic values} (in the game-theoretic sense), and which are equivalent to fitting a weighted additive model to a cooperative game \citep{ribeiro2016should}. We also found that many feature selection methods can be understood in terms of coalitional excess, and that these approaches are generalized by the optimization problem solved in the Masking Model approach \citep{dabkowski2017real}. These game-theoretic connections allow us to compare the various summarization techniques through the properties they satisfy (e.g., the Shapley axioms), as well as through their ease of interpretation for end-users (Section~\ref{sec:psychology}).

Because of its axiomatic derivation and its many desirable properties, the Shapley value provides, in our view, the most complete summary of how a model works. However, while several approaches provide fast Shapley value approximations
\citep{vstrumbelj2010efficient, lundberg2017unified, lundberg2020local, covert2021improving}, these techniques are considerably slower than the fastest removal-based explanations (Section~\ref{sec:complexity}), particularly in the model-agnostic setting. Furthermore, Shapley value-based explanations may not always be the best choice: they can be difficult for users to interpret due to their complexity, and variations of the Shapley value may be required to reflect causal relationships in the data \citep{frye2019asymmetric, heskes2020causal, wang2021shapley}.

Building on work that discusses probabilistic and information-theoretic explanations \citep{owen2014sobol, chen2018learning, covert2020understanding}, we found that multiple model behaviors can be understood as information-theoretic quantities. These connections require that removed features are marginalized out using their conditional distribution, and although this approach is challenging to implement, we showed that several removal strategies provide high-quality conditional distribution approximations (e.g., generative modeling approaches and models trained with missing features). Our work is among a growing number of papers lending support to this approach \citep{strobl2008conditional, zintgraf2017visualizing, agarwal2019explaining, aas2019explaining, slack2020fooling, covert2020understanding, frye2020shapley}, but we present a novel perspective on why this choice is justified: besides providing information-theoretic explanations, we find that marginalizing out features with their conditional distribution is the only approach that yields predictions
that satisfy standard probability axioms (Section~\ref{sec:consistency}).

We recognize, however, that users do not always require information-theoretic explanations. These explanations have the potentially undesirable property that features can be deemed important even if they are not used by the model in a functional sense; this property is useful in certain applications (e.g., bias detection), but in other cases it may be confusing to users. Another concern with this approach is spreading credit among correlated features \citep{merrick2019explanation, kumar2020problems}, because marginalizing out features with their conditional distribution guarantees that perfectly correlated features receive equal attributions \citep{covert2020understanding}. This property can be avoided by using a different feature removal approach \citep{janzing2019feature}, but recent work also argues for sparsifying attributions using a variant of the Shapley value \citep{frye2019asymmetric}; interestingly, this is a summary technique (third dimension of our framework) that helps resolve an issue that arises from the feature removal strategy (first dimension of our framework).

The growing interest in black-box ML models has spurred a remarkable amount of model explanation research, and the past decade has yielded numerous publications proposing innovative new methods. However, as the field has matured we have also seen a growing number of unification theories that reveal underlying similarities and implicit relationships among state-of-the-art methods \citep{lundberg2017unified, ancona2017towards, covert2020understanding}. Our framework for removal-based explanations is the broadest unifying theory yet, and it bridges the gap between parts of the explainability literature that may have otherwise been viewed as disjoint. We believe that this work represents an important step towards making the field more organized, rigorous, and easier to navigate.

An improved understanding of the field presents new opportunities for both explainability practitioners and researchers. For practitioners, our framework enables more explicit reasoning about the trade-offs between available explanation tools. The unique advantages of different methods are difficult to understand when each approach is viewed as a monolithic algorithm, but disentangling their choices makes it simpler to reason about their strengths and weaknesses, and potentially develop hybrid methods (as in Section~\ref{sec:experiments}).

For researchers, our framework offers a new theoretical perspective that can guide and strengthen ongoing research. Using the tools we developed, future work will be better equipped to (i)~specify the dimensions along which new methods differ from existing ones, (ii)~distinguish the implicit objectives of new approaches from their approximations, (iii)~resolve shortcomings in existing explanation methods using solutions along different dimensions of the framework, and (iv)~rigorously justify new approaches in light of their connections with cooperative game theory, information theory and psychology. As the number of removal-based explanations continues to grow, we hope that our framework will serve as a strong foundation upon which future research can build.

\newpage

\appendix
\section{Method Details} \label{app:methods}

Here, we provide additional details about the methods discussed in the main text. In several cases, we presented generalized versions of methods that deviated from their descriptions in the original papers, so we clarify those reformulations here.

\subsection{Meaningful Perturbations (MP)}

Meaningful Perturbations \citep{fong2017interpretable} considers multiple ways of deleting information from an input image. Given a mask $m \in [0, 1]^d$, MP uses a function $\Phi(x, m)$ to denote the modified input and suggests that the mask may be used to 1) set pixels to a constant value, 2) replace them with Gaussian noise, or 3) blur the image. In the blurring approach, which the authors recommend, each pixel $x_i$ is blurred separately using a Gaussian kernel with standard deviation given by $\sigma \cdot m_i$ (for a user-specified $\sigma > 0$).

To prevent adversarial solutions, MP incorporates a total variation norm on the mask, upsamples the mask from a low-resolution version, and uses a random jitter on the image during optimization. Additionally, MP uses a continuous mask $m \in [0, 1]^d$ in place of a binary mask $\{0, 1\}^d$ and a continuous $\ell_1$ penalty on the mask in place of the $\ell_0$ penalty. Although MP's optimization tricks are key to providing visually compelling explanations, our presentation focuses on the most important part of the optimization objective, which is reducing the classification probability while blurring only a small part of the image (Eq.~\ref{eq:mp_opt}).

\subsection{Extremal Perturbations (EP)}

Extremal Perturbations \citep{fong2019understanding} is an extension of MP with several modifications. The first is switching the objective from a ``removal game'' to a ``preservation game,'' which means learning a mask that preserves rather than removes the most salient information. The second is replacing the penalty on the subset size (or the mask norm) with a constraint. In practice, the constraint is enforced using a penalty, but the authors argue that it should be viewed as a constraint due to the use of a large regularization parameter.

EP uses the same blurring operation as MP and introduces new tricks to ensure a smooth mask, but our presentation focuses on the most important part of the optimization problem, which is maximizing the classification probability while blurring a fixed portion of the image (Eq.~\ref{eq:l2x_opt}).

\subsection{FIDO-CA}

FIDO-CA \citep{chang2018explaining} is similar to EP, but it replaces the blurring operation with features drawn from a generative model. The generative model $p_G$ can condition on arbitrary subsets of features, and although its samples are non-deterministic, FIDO-CA achieves strong results using a single sample. The authors consider multiple generative models but recommend a generative adversarial network (GAN) that uses contextual attention \citep{yu2018generative}. The optimization objective is based on the same ``preservation game'' as EP, and the authors use the Concrete reparameterization trick \citep{maddison2016concrete} for optimization.

\subsection{Minimal Image Representation (MIR)}

In the Minimal Image Representation approach, \cite{zhou2014object} remove information from an image to determine which regions are salient for the desired class. MIR creates a segmentation of edges and regions and iteratively removes segments from the image (selecting those that least decrease the classification probability) until the remaining image is incorrectly classified. We view this as a greedy approach for solving the constrained optimization problem

\begin{equation*}
    \min_S \; |S| \quad \mathrm{s.t.} \;\; u(S) \geq t,
\end{equation*}

\noindent where $u(S)$ represents the prediction with the specified subset of features, and $t$ represents the minimum allowable classification probability. Our presentation of MIR in the main text focuses on this view of the optimization objective rather than the specific choice of greedy algorithm (Eq.~\ref{eq:mir_opt}).

\subsection{Masking Model (MM)}

The Masking Model approach \citep{dabkowski2017real} observes that removing salient information (while preserving irrelevant information) and removing irrelevant information (while preserving salient information) are both reasonable approaches for understanding image classifiers. The authors refer to these tasks as discovering the smallest destroying region (SDR) and smallest sufficient region (SSR).

The authors adopt notation similar to \cite{fong2017interpretable}, using $\Phi(x, m)$ to denote the transformation to the input given a mask $m \in [0, 1]^d$. For an input $x \in \mathcal{X}$, MM aims to solve the following optimization problem, which requires four hyperparameters:

\begin{equation*}
    \min_m \; \lambda_1 \mathrm{TV}(m) + \lambda_2 ||m||_1 - \log f\big(\Phi(x, m)\big) + \lambda_3 f\big(\Phi(x, 1 - m)\big)^{\lambda_4}.
\end{equation*}

The $\mathrm{TV}$ (total variation) and $\ell_1$ penalty terms are both similar to MP and respectively encourage smoothness and sparsity in the mask. Unlike MP, MM learns a global explainer model that outputs approximate solutions to this problem in a single forward pass. In the main text, we present a simplified version of the problem that does not include the logarithm in the third term or the exponent in the fourth term (Eq.~\ref{eq:mm_opt}). We view these as monotonic link functions that provide a more complex trade-off between the objectives, but that are not necessary for finding informative solutions.

\subsection{Learning to Explain (L2X)}

The L2X method performs instance-wise feature selection by learning an auxiliary model $g_\alpha$ and a selector model $V_\theta$ (see Eq.~6 of \citealp{chen2018learning}). These models are learned jointly and are optimized via the similarity between predictions from $g_\alpha$ and from the original model, denoted as $\mathbbm{P}_m$ by \cite{chen2018learning}. With slightly modified notation that highlights the selector model's dependence on $X$, the L2X objective can be written as:

\begin{equation}
    \max_{\alpha, \theta} \; \E_{X, \zeta} \E_{Y \sim \mathbbm{P}_m(X)} \Big[ \log g_\alpha\big(V_\theta(X, \zeta) \odot X, Y\big) \Big]. \label{eq:l2x_objective}
\end{equation}

In Eq.~\ref{eq:l2x_objective}, the random variables $X$ and $\zeta$ are sampled independently, $Y$ is sampled from the model's distribution $\mathbbm{P}_m(X)$, $V_\theta(X, \zeta) \odot X$ represents an element-wise multiplication with (approximately) binary indicator variables $V_\theta(X, \zeta)$ sampled from the Concrete distribution \citep{maddison2016concrete}, and $\log g_\alpha(\cdot, Y)$ represents the model's estimate of $Y$'s log-likelihood.

We can gain more insight into this objective function by reformulating it. If we let $V_\theta(X, \zeta)$ be a deterministic function $\epsilon(X)$, interpret the log-likelihood as a loss function $\ell$ for the prediction from $g_\alpha$ (e.g., cross entropy loss) and represent $g_\alpha$ as a subset function $F$, then we can rewrite the L2X objective as follows:

\begin{equation*}
    \max_{F, \epsilon} \; \E_{X} \E_{Y \sim \mathbbm{P}_m(X)} \Big[ \ell\big( F\big(X, \epsilon(X)\big), Y\big) \Big].
\end{equation*}

Next, rather than considering the expected loss for labels $Y$ distributed according to $\mathbbm{P}_m(X)$, we can rewrite this as a loss between the subset function's prediction $F\big(X, \epsilon(X)\big)$ and the full model prediction $f(X) \equiv \mathbbm{P}_m(X)$:

\begin{equation*}
    \max_{F, \epsilon} \; \E_{X} \Big[ \ell\big( F\big(X, \epsilon(X)\big), f(X)\big) \Big].
\end{equation*}

Finally, we can see that L2X implicitly trains a surrogate model $F$ to match the original model's predictions, and that the optimization objective for each input $x \in \mathcal{X}$ is given by

\begin{equation*}
    S^* = \argmax_{|S| = k} \; \ell\big(F(x_S), f(x)\big).
\end{equation*}

This matches the description of L2X provided in the main text (Eqs.~\ref{eq:surrogate}, \ref{eq:plo_game}, \ref{eq:l2x_opt}). It is only when we have $f(x) = \mathbbm{P}_m(x) = p(Y \mid X = x)$ and $F(x_S) = p(Y \mid X_S = x_S)$ that L2X's information-theoretic interpretation holds, at least in the classification case. Or, in the regression case, where we can replace the log-likelihood with a simpler MSE loss, L2X can be interpreted in terms of conditional variance minimization (rather than mutual information maximization) when we have $f(x) = \E[Y \mid X = x]$ and $F(x_S) = \E[Y \mid X_S = x_S]$.

\subsection{Instance-wise Variable Selection (INVASE)} \label{app:invase}

INVASE \citep{yoon2018invase} is similar to L2X in that it performs instance-wise feature selection using a learned selector model. However, INVASE has several differences in its implementation and objective function. INVASE relies on three separate models: a prediction model, a baseline model and a selector model. The baseline model is trained to predict the true label $Y$ given the full feature vector $X$, and it can be trained independently of the remaining models; the predictor model makes predictions given subsets of features $X_S$ (with $S$ sampled according to the selector model), and it is trained to predict the true labels $Y$; and finally, the selector model takes a feature vector $X$ and outputs a probability distribution for a subset $S$.

The selector model, which ultimately outputs explanations, relies on the baseline model primarily for variance reduction purposes \citep{yoon2018invase}. Because the sampled subsets are used only for the predictor model, which is trained to predict the true label $Y$ (rather than the baseline model's predictions), we view the prediction model as the model being explained, and we understand it as removing features via a strategy of introducing missingness during training (Eq.~\ref{eq:missingness}).

For the optimization objective, \cite{yoon2018invase} explain that their aim is to minimize the following KL divergence for each input $x \in \mathcal{X}$:

\begin{equation*}
    S^* = \argmin_S \; \KL\big(p(Y \mid X = x) \; \big|\big| \; p(Y \mid X_S = x_S) \big) + \lambda |S|.
\end{equation*}

\noindent This is consistent with their learning algorithm if we assume that the predictor model outputs the Bayes optimal prediction $p(Y \mid X_S = x_S)$.
If we denote their predictor model as a subset function $F$ and interpret the KL divergence as a loss function with the true label $Y$ (i.e., cross entropy loss), then we can rewrite this objective as follows:

\begin{equation*}
    S^* = \argmin_S \; \E_{Y \mid X = x} \Big[ \ell\big(F(x_S), Y\big) \Big] + \lambda |S|.
\end{equation*}

\noindent This is the description of INVASE provided in the main text.



\subsection{REAL-X}

REAL-X \citep{jethani2021have} is similar to L2X and INVASE in that it uses a learned selector model to perform instance-wise feature selection. REAL-X is designed to resolve a flaw in L2X and INVASE, which is that both methods learn the selector model jointly with their subset functions, enabling label information to be leaked via the selected subset $S$. 

To avoid this issue, REAL-X learns a subset function $F$ independently from the selector model using the following objective function (with modified notation):

\begin{equation*}
    \min_F \; \E_{X} \E_{Y \sim f(X)} \E_{S} \Big[ \ell\big(F(X_S), Y\big) \Big].
\end{equation*}

\noindent The authors point out that $Y$ may be sampled from its true conditional distribution $p(Y \mid X)$ or from a model's distribution $Y \sim f(X)$; we remark that the former is analogous to INVASE (missingness introduced during training) and that the latter is analogous to L2X (training a surrogate model).
Notably, unlike L2X or INVASE, REAL-X optimizes its subset function with the subsets $S$ sampled independently from the input $X$, enabling it to approximate the Bayes optimal predictions $F(x_S) \approx p(Y \mid X_S = x_S)$ (Appendix~\ref{app:conditional_approximations}).

We focus on the case with the label sampled according to $Y \sim f(X)$, which can be understood as fitting a surrogate model $F$ to the original model $f$. With the learned subset function $F$ fixed, REAL-X then learns a selector model that optimizes the following objective for each input $x \in \mathcal{X}$:

\begin{equation*}
    S^* = \argmin_S \; \E_{Y \sim f(x)} \Big[ \ell\big( F(x_S), Y \big) \Big] + \lambda |S|.
\end{equation*}

\noindent Rather than viewing this as the mean loss for labels sampled according to $f(x)$, we interpret this as a loss function between $F(x_S)$ and $f(x)$, as we did with L2X:

\begin{equation*}
    S^* = \argmin_S \; \ell\big( F(x_S), f(x) \big) + \lambda |S|.
\end{equation*}

\noindent This is our description of REAL-X provided in the main text.

\subsection{Prediction Difference Analysis (PredDiff)}

Prediction Difference Analysis \citep{zintgraf2017visualizing}, which is based on a precursor method to IME \citep{robnik2008explaining, vstrumbelj2009explaining}, removes individual features (or groups of features) and analyzes the difference in the model's prediction. Removed pixels are imputed by conditioning on their bordering pixels, which approximates sampling from the full conditional distribution. The authors then calculate attribution scores based on the difference in log-odds ratio:

\begin{equation*}
    a_i = \log \frac{F(x)}{1 - F(x)} - \log \frac{F(x_{D \setminus \{i\}})}{1 - F(x_{D \setminus \{i\}})}.
\end{equation*}

\noindent We view this as another way of analyzing the difference in the model output for an individual prediction, simply substituting the log-odds ratio for the classification probabilities.

\subsection{Causal Explanations (CXPlain)}

CXPlain removes single features (or groups of features) for individual inputs and measures the change in the loss function \citep{schwab2019cxplain}. The authors propose calculating the attribution values

\begin{equation*}
    a_i(x) = \ell\big(F(x_{D \setminus \{i\}}), y\big) - \ell\big(F(x, y)\big)
\end{equation*}

\noindent and then computing the normalized values

\begin{equation*}
    w_i(x) = \frac{a_i(x)}{\sum_{j=1}^d a_j(x)}.
\end{equation*}

\noindent The normalization step enables the use of a KL divergence-based learning objective for the explainer model (although it is not obvious how to handle negative values $a_i(x) < 0$), which is ultimately used to calculate attribution values in a single forward pass. The authors explain that this approach is based on a ``causal objective,'' but CXPlain is only causal in the same sense as every other method described in our work (Section~\ref{sec:psychology}), i.e., it measures how each feature causes the model output to change. This is not to be confused with methods that integrate ideas from causal inference \citep{heskes2020causal, wang2021shapley}.

\subsection{Randomized Input Sampling for Explanation (RISE)}

RISE \citep{petsiuk2018rise} begins by generating a large number of randomly sampled binary masks. In practice, the masks are sampled by dropping features from a low-resolution mask independently with probability $p$, upsampling to get an image-sized mask, and then applying a random jitter. Due to the upsampling, the masks have values $m \in [0, 1]^d$ rather than $m \in \{0, 1\}^d$.

The mask generation process induces a distribution over the masks, which we denote as $p(m)$. The method then uses the randomly generated masks to obtain a Monte Carlo estimate of the following attribution values:

\begin{equation*}
    a_i = \frac{1}{\E[M_i]} \E_{p(M)}\big[f(x \odot M) \cdot M_i\big].
\end{equation*}

If we ignore the upsampling step that creates continuous mask values, we see that these attribution values are the mean prediction when a given pixel is included:

\begin{align*}
    a_i &= \frac{1}{\E[M_i]} \E_{p(M)}\big[f(x \odot M) \cdot M_i\big] \\
    &= \sum_{m \in \{0, 1\}^d} f(x \odot m) \cdot m_i \cdot \frac{p(m)}{\E[M_i]} \\
    &= \E_{p(M \mid M_i = 1)}\big[f(x \odot M)\big].
\end{align*}

\noindent This is our description of RISE provided in the main text.

\subsection{Interactions Methods for Explanations (IME)}

IME was presented in two separate papers \citep{vstrumbelj2009explaining, vstrumbelj2010efficient}. In the original version, the authors recommended training a separate model for each subset of features; in the second version, the authors proposed the more efficient approach of marginalizing out the removed features from a single model $f$.

The latter paper is somewhat ambiguous about the specific distribution used when marginalizing out held out features. \cite{lundberg2017unified} view that features are marginalized out using their distribution from the training dataset (i.e., the marginal distribution), whereas \cite{merrick2019explanation} view IME as marginalizing out features using a uniform distribution. We opt for the uniform interpretation, but IME's specific choice of distribution does not impact any of our conclusions.

\subsection{SHAP}

SHAP \citep{lundberg2017unified} explains individual predictions by decomposing them with the game-theoretic Shapley value \citep{shapley1953value}, similar to IME \citep{vstrumbelj2009explaining, vstrumbelj2010efficient} and QII \citep{datta2016algorithmic}. The original work proposed marginalizing out removed features with their conditional distribution but remarked that the joint marginal provided a practical approximation (see Section~\ref{sec:conditional_approximations} for a similar argument). Marginalizing using the joint marginal distribution is now the default behavior in SHAP's online implementation. KernelSHAP is an approximation approach based on solving a weighted least squares problem \citep{lundberg2017unified, covert2021improving}.

\subsection{TreeSHAP}

TreeSHAP uses a unique approach to handle held out features in tree-based models \citep{lundberg2020local}: it accounts for missing features using the distribution induced by the underlying trees, which, since it exhibits no dependence on the held out features, is a valid subset extension of the original model. However, it cannot be viewed as marginalizing out features using a simple distribution (i.e., one whose density function we can write down).

Given a subset of features, TreeSHAP makes a prediction separately for each tree and then combines each tree's prediction in the standard fashion. But when a split for an unknown feature is encountered, TreeSHAP averages predictions over the multiple paths in proportion to how often the dataset follows each path. This is similar but not identical to the conditional distribution because each time this averaging step is performed, TreeSHAP conditions only on coarse information about the features that preceded the split.

\subsection{LossSHAP}

LossSHAP is a version of SHAP that decomposes the model's loss for an individual prediction rather than the prediction itself. The approach was first considered in the context of TreeSHAP \citep{lundberg2020local}, and it has been discussed in more detail as a local analogue to SAGE \citep{covert2020understanding}.

\subsection{Shapley Net Effects}

Shapley Net Effects \citep{lipovetsky2001analysis} was proposed as a variable importance measure for linear models. The method becomes impractical with large numbers of features or non-linear models, but \cite{williamson2020efficient} generalize the approach by using an efficient linear regression-based Shapley value estimator: SPVIM can be run with larger datasets and non-linear models because it requires checking a smaller number of feature subsets. Both Shapley Net Effects and SPVIM can also be used with other model performance measures, such as area under the ROC curve or the $R^2$ value.

\subsection{Shapley Effects}

Shapley Effects analyzes a variance-based measure of a function's sensitivity to its inputs, with the goal of discovering which features are responsible for the greatest variance reduction in the model output \citep{owen2014sobol}. The cooperative game described in the original paper is the following:

\begin{equation*}
    u(S) = \Var\Big(\E\big[f(X) \mid X_S\big]\Big).
\end{equation*}

We present a generalized version to cast this method in our framework. In the appendix of \cite{covert2020understanding}, it was shown that this game can be reformulated as follows:

\begin{align*}
    u(S) &= \Var\Big(\E\big[f(X) \mid X_S\big]\Big) \\
    &= \Var\big(f(X)\big) - \E\Big[\Var\big(f(X) \mid X_S\big)\Big] \\
    &= c - \E\Big[\ell\big(\E\big[f(X) \mid X_S\big], f(X)\big)\Big] \\
    &= c - \underbrace{\E\Big[\ell\big(F(X_S), f(X)\big)\Big]}_{\mathclap{\text{Dataset loss w.r.t. output}}}.
\end{align*}

This derivation assumes that the loss function $\ell$ is MSE and that the subset function $F$ is $F(x_S) = \E[f(X) \mid X_S = x_S]$. Rather than the original formulation, we present a cooperative game that is equivalent up to a constant value and that provides flexibility in the choice of loss function:

\begin{equation*}
    w(S) =  - \E\Big[\ell\big(F(X_S), f(X)\big)\Big].
\end{equation*}

\subsection{LIME}

For an overview of LIME \citep{ribeiro2016should}, we direct readers to Appendix~\ref{app:lime}.

\section{Additive Model Proofs} \label{app:lime}

LIME calculates feature attribution values by fitting a weighted regularized linear model to an \textit{interpretable representation} of the input \citep{ribeiro2016should}. If we consider that the interpretable representation is binary (the default behavior in LIME's implementation), then the model is represented by a set function $u: \mathcal{P}(D) \mapsto \R$ when we take an expectation over the distribution of possible feature imputations. LIME is therefore equivalent to fitting an additive model to a set function, which means solving the optimization problem

\begin{align*}
    \min_{b_0, \ldots, b_d} \; L(b_0, \ldots, b_d) + \Omega(b_1, \ldots, b_d), \label{eq:lime_objective}
\end{align*}

\noindent where we define $L$ (the weighted least squares component) as

\begin{align}
    L(b_0, \ldots, b_d) = \sum_{S \subseteq D} \pi(S) \Big(b_0 + \sum_{i \in S} b_i - u(S) \Big)^2.
\end{align}

For convenience, we refer to this as the weighted least squares (WLS) approach to summarizing set functions, and we show that several familiar attribution values can be derived by choosing different weighting kernels $\pi$ and omitting the regularization term (i.e., setting $\Omega = 0$).

\subsection{Include individual}

\noindent Consider the weighting kernel $\pi_{\mathrm{Inc}}(S) = \mathbbm{1}(|S| \leq 1)$, which puts weight only on coalitions that have no more than one player. With this kernel, the WLS problem reduces to:

\begin{align*}
    a_1, \ldots a_d = \argmin_{b_0, \ldots, b_d} \; 
    \Big(b_0 - u(\{\})\Big)^2 + \sum_{i = 1}^d \Big(b_0 + b_i - u(\{i\})\Big)^2.
\end{align*}

\noindent It is clear that the unique global minimizer of this problem is given by the following solution:

\begin{align*}
    a_0 &= u(\{\}) \\
    a_i &= u(\{i\}) - u(\{\}).
\end{align*}

The WLS approach will therefore calculate the attribution values $a_i = u(\{i\}) - u(\{\})$, which is equivalent to how Occlusion, PredDiff and CXPlain calculate local feature importance and how permutation tests and feature ablation (LOCO) summarize global feature importance \citep{breiman2001random, strobl2008conditional, zeiler2014visualizing, zintgraf2017visualizing, lei2018distribution, schwab2019cxplain}.

\subsection{Remove individual}

\noindent Consider the weighting kernel $\pi_{\mathrm{Ex}}(S) = \mathbbm{1}(|S| \geq d - 1)$, which puts weight only on coalitions that are missing no more than one player. With this kernel, the WLS problem reduces to:

\begin{align*}
    a_1, \ldots, a_d = &\argmin_{b_0, \ldots, b_d} \; 
    \Big(b_0 + \sum_{j \in D} b_j - u(D) \Big)^2 + \sum_{i = 1}^d \Big(b_0 + \sum_{j \in D \setminus \{i\}} b_j - u(D \setminus \{i\})\Big)^2.
\end{align*}

\noindent It is clear that the unique global minimizer of this problem is given by the following solution:

\begin{align*}
    a_0 &= u(D) - \sum_{i \in D} a_i \\
    a_i &= u(D) - u(D \setminus \{i\}).
\end{align*}

The WLS approach will therefore calculate the attribution values $a_i = u(D) - u(D \setminus \{i\})$, which is equivalent to how the univariate predictors approach summarizes global feature importance \citep{guyon2003introduction}.

\subsection{Banzhaf value}

Consider the weighting kernel $\pi_{\mathrm{B}}(S) = 1$, which yields an unweighted least squares problem. This version of the WLS
problem has been analyzed in prior work, which showed that the optimal coefficients are the Banzhaf values \citep{hammer1992approximations}.

As an alternative proof, we demonstrate that a solution that uses the Banzhaf values is optimal by proving that its partial derivatives are zero. To begin, consider the following candidate solution, which uses the Banzhaf values $a_i = \psi_i(u)$ for $i = 1, \ldots, d$ and a carefully chosen intercept term $a_0$:

\begin{align*}
    a_0 &= \frac{1}{2^d}\sum_{S \subseteq D} u(S) - \frac{1}{2}\sum_{j = 1}^d a_j \\
    a_i &= \frac{1}{2^{d-1}} \sum_{S \subseteq D \setminus \{i\}} \Big( u(S \cup \{i\}) - u(S) \Big) = \psi_i(u).
\end{align*}

We can verify whether this is a solution to the unweighted least squares problem by checking if the partial derivatives are zero. We begin with the derivative for the intercept:

\begin{align*}
    \frac{\partial}{\partial b_0} L(a_0, \ldots, a_d)
    &= 2 \sum_{S \subseteq D} \Big( a_0 + \sum_{j \in S} a_j - u(S) \Big) \\
    &= 2 \Big(2^d a_0 + 2^{d-1}\sum_{j = 1}^d a_j - \sum_{S \subseteq D} u(S) \Big) \\
    &= 0.
\end{align*}

\noindent Next, we verify the derivatives for the other parameters $a_i$ for $i = 1, \ldots, d$:

\begin{align*}
    \frac{\partial}{\partial b_i} L(a_0, \ldots, a_d)
    &= 2 \sum_{T \supseteq \{i\}} \Big(a_0 + \sum_{j \in T} a_j - u(T) \Big) \\
    &= 2 \Big( 2^{d-1} a_0 + 2^{d-2}\sum_{j = 1}^d a_j + 2^{d-2} a_i - \sum_{T \supseteq \{i\}} u(T) \Big) \\
    &= 2^{d-1} a_i + \sum_{S \subseteq D \setminus \{i\}} \Big( u(S) - u(S \cup \{i\}) \Big) \\
    &= 0.
\end{align*}

Because the gradient is zero and the problem is jointly convex in $(b_0, \ldots, b_d)$, we conclude that the solution given above is optimal and unique. The optimal coefficients $(a_1, \ldots, a_d)$ are precisely the Banzhaf values $\psi_1(u), \ldots, \psi_d(u)$ of the cooperative game $u$.

\subsection{Shapley value}

The weighted least squares problem is optimized by the Shapley value when we use the following weighting kernel:

\begin{align*}
    \pi_{\mathrm{Sh}}(S) = \frac{d - 1}{\binom{d}{|S|}|S|(d - |S|)}.
\end{align*}

\noindent Since this connection has been noted in other model explanation works, we direct readers to existing proofs \citep{charnes1988extremal, lundberg2017unified}.

\section{Axioms for Other Approaches} \label{app:alternative_axioms}

Here, we describe which Shapley axioms or Shapley-like axioms apply to other summarization approaches.

\subsection{Additive model axioms} 

By fitting a regularized weighted least squares model to a cooperative game, as in LIME \citep{ribeiro2016should}, we effectively create an explanation mapping of the form $E: \mathcal{U} \mapsto \R^d$. We can show that this mapping satisfies a subset of the Shapley value axioms (Section~\ref{sec:solution_concepts}). To do so, we make the following assumptions about the weighting kernel $\pi$ and regularization function $\Omega$ (introduced in Eq.~\ref{eq:lime}):

\begin{enumerate}
    \item The weighting kernel $\pi$ is non-negative and finite for all $S \subseteq D$ except for possibly the sets $\{\}$ and $D$.
    \item The weighting kernel $\pi$ satisfies the inequality
    
    \begin{align*}
        \begin{pmatrix}\mathbf{1} & X\end{pmatrix}^T W \begin{pmatrix}\mathbf{1} & X\end{pmatrix} \succeq 0,
    \end{align*}
    
    where $X \in \R^{2^d \times d}$ contains an enumeration of binary representations for all subsets $S \subseteq D$, and $W \in \R^{2^d \times 2^d}$ is a diagonal matrix containing an aligned enumeration of $\pi(S)$ for $S \subseteq D$. This ensures that the weighted least squares component of the objective
    is strictly convex.

    \item The regularizer $\Omega$ is convex (e.g., the Lasso or ridge penalty).
\end{enumerate}

\noindent We now address each property in turn for the weighted least squares (WLS) approach:

\begin{itemize}
    \item (Efficiency) The WLS approach satisfies the efficiency property only when the weighting kernel is chosen such that $\pi(\{\}) = \pi(D) = \infty$. These weights are equivalent to the constraints $b_0 = u(\{\})$ and $\sum_{i \in D} b_i = u(D) - u(\{\})$. In cooperative game theory, additive models with these constraints are referred to as \textit{faithful linear approximations} \citep{hammer1992approximations}.
    
    \item (Symmetry) The WLS approach satisfies the symmetry axiom as long as the weighting kernel $\pi$ and the regularizer $\Omega$ are permutation-invariant (i.e., $\pi$ is a function of the subset size, and $\Omega$ is invariant to the ordering of parameters). To see this, consider an optimal solution with parameters $(b_0^*, \ldots b_d^*)$. Swapping the coefficients $b_i^*$ and $b_j^*$ for features with identical marginal contributions gives the same objective value, so this is still optimal. The strict convexity of the objective function implies that there is a unique global optimum, so we conclude that $b_i^* = b_j^*$.
    
    \item (Dummy) The dummy property holds for the Shapley and Banzhaf weighting kernels $\pi_{\mathrm{B}}$ and $\pi_{\mathrm{Sh}}$, but it does not hold for arbitrary $\pi, \Omega$.
    
    \item (Additivity) The additivity property holds when the regularizer is set to $\Omega$ = 0. This can be seen by viewing the solution to the WLS problem as a linear function of the response variables \citep{kutner2005applied}.
    
    \item (Marginalism) Given two games $u, u'$ where players have identical marginal contributions, we can see that $u' = u + c$ for some $c \in \R$. The WLS approach satisfies the marginalism property because it learns identical coefficients $b_1^*, \ldots, b_d^*$ but different intercepts connected by the equation $b_0^*(u') = b_0^*(u) + c$.
\end{itemize}

\subsection{Feature selection axioms} 

The feature selection summarization techniques (MP, MIR, L2X, INVASE, REAL-X, EP, MM, FIDO-CA) satisfy properties that are similar to the Shapley value axioms. Each method outputs an optimal coalition $S^* \subseteq D$ rather than an allocation $a \in \R^d$, so the Shapley value axioms do not apply directly. However, we identify the following analogous properties:

\begin{itemize}
    \item (Symmetry) If there are two players $i, j$ with identical marginal contributions
    and there exists an optimal coalition $S^*$ that satisfies $i \in S$ and $j \notin S$, then the coalition $\big(S^* \cup \{j\}\big) \setminus \{i\}$ is also optimal.
    
    \item (Dummy) For a player $i$ that makes zero marginal contribution, there must be an optimal solution $S^*$ such that $i \notin S^*$.
    
    \item (Marginalism) For two games $u, u'$ where all players have identical marginal contributions,
    the coalition $S^*$ is optimal for $u$ if and only if it is optimal for $u'$.
\end{itemize}

The feature selection explanations do not seem to satisfy properties that are analogous to the efficiency or additivity axioms. And unlike the attribution case, these properties are insufficient to derive a unique, axiomatic approach.

\section{Consistency Proofs} \label{app:proofs}

Here, we restate and prove the results from Section~\ref{sec:consistency}, which relate to subset extensions of a model $f \in \mathcal{F}$ that are consistent (Definition~\ref{def:consistency}) with a probability distribution $q(X)$. Both results can be shown using simple applications of basic probability laws.

\bigskip

\noindent \textbf{Proposition \ref{theorem:unique_probability}}
{\it For a classification model $f \in \mathcal{F}$ that estimates a discrete $Y$'s conditional probability, there is a unique subset extension $F \in \mathfrak{F}$ that is consistent with $q(X)$,

\begin{equation*}
    F(x_S) = \E_{q(X_{\bar S} \mid X_S = x_S)}[f(x_S, X_{\bar S})],
\end{equation*}

\noindent where $q(X_{\bar S} \mid X_S = x_S)$ is the conditional distribution induced by $q(X)$.}

\bigskip

\begin{proof}
    We begin by assuming the existence of a subset function $F \in \mathfrak{F}$ that satisfies $q(Y \mid X = x) = F(x) = f(x)$ for all $x \in \mathcal{X}$ and consider how the probability axioms can be used to compute the conditional probability $q(Y \mid X_S = x_S)$ given only $q(X)$ and $q(Y \mid X = x)$.
    
    For this proof we consider the case of discrete $X$, but a similar argument can be used to prove the same result with continuous $X$. Below, we provide a step-by-step derivation of $q(Y \mid X_S = x_S)$ that indicates which axioms or definitions are used in each step. (Axiom 1 refers to the countable additivity property and Axiom 2 refers to Bayes rule.)

    \begin{align*}
        q(y \mid x_S)
        &= \sum_{x_{\bar S} \in \mathcal{X}_{\bar S}} q(y, x_{\bar S} \mid x_S) \tag*{\small (Axiom 1)} \\
        &= \sum_{x_{\bar S} \in \mathcal{X}_{\bar S}} \frac{q(y, x_S, x_{\bar S})}{q(x_S)} \tag*{\small (Axiom 2)} \\
        &= \sum_{x_{\bar S} \in \mathcal{X}_{\bar S}} q(y \mid x_S, x_{\bar S})\frac{q(x_S, x_{\bar S})}{q(x_S)} \tag*{\small (Axiom 2)} \\
        &= \sum_{x_{\bar S} \in \mathcal{X}_{\bar S}} f(x_S, x_{\bar S}) \cdot q(x_{\bar S} \mid x_S) \tag*{\small (Definition of $f$, Axiom 2)} \\
        &= \E_{q(X_{\bar S} \mid X_S = x_S)}\big[f(x_S, X_{\bar S})\big] \tag*{\small(Definition of expectation)}
    \end{align*}
    
    This derivation shows that in order to be consistent with $q(X)$ according to the probability laws, $F$ must be defined as follows:
    
    \begin{equation}
        F(x_S) \equiv \E_{q(X_{\bar S} \mid X_S = x_S)}\big[f(x_S, X_{\bar S})\big]. \label{eq:consistent}
    \end{equation}
    
    To complete the proof, we consider whether there are other ways of deriving $F$'s behavior that may demonstrate inconsistency. The only other case to consider is whether we can derive a unique definition for $F(x_{S \setminus T})$ when beginning from $F(x)$ and when beginning from $F(x_S)$ for $T \subset S \subset D$. The first result is given by Eq.~\ref{eq:consistent}, and we derive the second result as follows:
    
    \begin{align*}
        F(x_{S \setminus T}) &= q(y \mid x_{S \setminus T}) \\
        &= \sum_{x_T \in \mathcal{X}_T} q(y, x_T \mid x_{S \setminus T}) \tag*{\small (Axiom 1)} \\
        &= \sum_{x_T \in \mathcal{X}_T} \frac{q(y, x_T, x_{S \setminus T})}{q(x_{S \setminus T})} \tag*{\small (Axiom 2)} \\
        &= \sum_{x_T \in \mathcal{X}_T} q(y \mid x_T, x_{S \setminus T}) \frac{q(x_T, x_{S \setminus T})}{q(x_{S \setminus T})} \tag*{\small (Axiom 2)} \\
        &= \sum_{x_T \in \mathcal{X}_T} q(y \mid x_T, x_{S \setminus T}) \cdot q(x_T \mid x_{S \setminus T}) \tag*{\small (Axiom 2)} \\
        &= \sum_{x_T \in \mathcal{X}_T} F(x_T, x_{S \setminus T}) \cdot q(x_T \mid x_{S \setminus T}) \tag*{\small (Definition of $F$)} \\
        &= \sum_{x_T \in \mathcal{X}_T} \E_{q(X_{\bar S} \mid X_S = x_S)}\big[f(x_T, x_{S \setminus T}, X_{\bar S})\big] \cdot q(x_T \mid x_{S \setminus T}) \tag*{\small (Definition of $F$)} \\
        &= \sum_{x_T \in \mathcal{X}_T} \sum_{x_{\bar S} \in \mathcal{X}_{\bar S}} f(x_T, x_{S \setminus T}, x_{\bar S}) \cdot q(x_{\bar S} \mid x_T, x_{S \setminus T}) \cdot q(x_T \mid x_{S \setminus T}) \tag*{\small (Expectation)} \\
        &= \sum_{x_T \in \mathcal{X}_T} \sum_{x_{\bar S} \in \mathcal{X}_{\bar S}} f(x_T, x_{S \setminus T}, x_{\bar S}) \cdot q(x_T, x_{\bar S} \mid x_{S \setminus T}) \tag*{\small (Axiom 2)} \\
        &= \E_{q(X_T, X_{\bar S} \mid X_{S \setminus T} = x_{S \setminus T})}\big[f(X_T, x_{S \setminus T}, X_{\bar S})\big] \tag*{\small (Definition of expectation)}
    \end{align*}
    
    This result shows that deriving $F(x_{S \setminus T})$ from $F(x)$ or from $F(x_S)$ yields a consistent result. We conclude that our definition of $F$, which marginalizes out missing features with the conditional distribution induced by $q(X)$, provides the unique subset extension of $f$ that is consistent with $q(X)$.
\end{proof}

\bigskip

\noindent \textbf{Proposition \ref{theorem:unique_expectation}}
{\it For a regression model $f \in \mathcal{F}$ that estimates $Y$'s conditional expectation, there is a unique subset extension $F \in \mathfrak{F}$ that is consistent with $q(X)$,
    
\begin{equation*}
    F(x_S) = \E_{q(X_{\bar S} \mid X_S = x_S)}[f(x_S, X_{\bar S})],
\end{equation*}

\noindent where $q(X_{\bar S} \mid X_S = x_S)$ is the conditional distribution induced by $q(X)$.}

\bigskip

\begin{proof}
    Unlike the previous proof, $F$ does not directly define some $q(Y \mid X_S = x_S)$.
    However, $F$ represents an estimate of the conditional expectation $\E[Y \mid X_S = x_S]$ for each $S \subseteq D$, so we can assume the existence of conditional distributions $q(Y \mid X_S = x_S)$
    that satisfy
    
    \begin{equation*}
        F(x_S) = \E_{q(Y \mid X_S = x_S)}[Y].
    \end{equation*}
    
    \noindent We show that our probability laws are sufficient to constrain the conditional expectation represented by $F$ to have a unique definition for each $S \subset D$.
    
    Consider how the probability laws can be used to compute $q(Y \mid X_S = x_S)$ given only $q(X)$ and the assumed $q(Y \mid X = x)$. For this proof, we consider the case of discrete $X$ and $Y$, but a similar argument can be used to prove the same result for continuous $X$ and $Y$. Below, we provide a step-by-step derivation of $q(Y \mid X_S = x_S)$ that indicates which axioms or definitions are used in each step. (Axiom 1 refers to the countable additivity property and Axiom 2 refers to Bayes rule.)

    \begin{align*}
        q(y \mid x_S)
        &= \sum_{x_{\bar S} \in \mathcal{X}_{\bar S}} q(y, x_{\bar S} \mid x_S) \tag*{\small (Axiom 1)} \\
        &= \sum_{x_{\bar S} \in \mathcal{X}_{\bar S}} \frac{q(y, x_S, x_{\bar S})}{q(x_S)} \tag*{\small (Axiom 2)} \\
        &= \sum_{x_{\bar S} \in \mathcal{X}_{\bar S}} q(y \mid x_S, x_{\bar S})\frac{q(x_S, x_{\bar S})}{q(x_S)} \tag*{\small (Axiom 2)} \\
        &= \sum_{x_{\bar S} \in \mathcal{X}_{\bar S}} q(y \mid x_S, x_{\bar S}) \cdot q(x_{\bar S} \mid x_S) \tag*{\small (Axiom 2)} \\
        &= \E_{q(X_{\bar S} \mid X_S = x_S)}\big[q(y \mid x_S, X_{\bar S})\big] \tag*{\small (Definition of expectation)}
    \end{align*}
    
    This derivation shows that in order to be consistent with $q(X)$,
    the conditional distribution $q(Y \mid X_S = x_S)$ must be defined as follows:
    
    \begin{equation*}
        q(Y \mid X_S = x_S) = \E_{q(X_{\bar S} \mid X_S = x_S)}[q(x_S, X_{\bar S})].
    \end{equation*}
    
    \noindent Since $F$ represents the expectation of these distributions, it can be derived as follows:
    
    \begin{align*}
        F(x_S) &= \E_{q(Y \mid X_S = x_S)}[Y] \tag*{\small (Definition of $F$)} \\
        &= \sum_{y \in \mathcal{Y}} y \cdot q(y \mid x_S) \tag*{\small (Definition of expectation)} \\
        &= \sum_{y \in \mathcal{Y}} \sum_{x_{\bar S} \in \mathcal{X}_{\bar S}} y \cdot q(y \mid x_S, x_{\bar S}) \cdot q(x_{\bar S} \mid x_S) \tag*{\small (Previous derivation)} \\
        &= \sum_{x_{\bar S} \in \mathcal{X}_{\bar S}} \E[Y \mid x_S, x_{\bar S}] \cdot q(x_{\bar S} \mid x_S) \tag*{\small (Interchanging order of sums)} \\
        &= \sum_{x_{\bar S}} f(x_S, x_{\bar S}) \cdot q(x_{\bar S} \mid x_S) \tag*{\small (Definition of $f$)} \\
        &= \E_{q(X_{\bar S} \mid X_S = x_S)}\big[f(x_S, X_{\bar S})\big] \tag*{\small (Definition of expectation)}
    \end{align*}
    
    \noindent According to this result, the probability laws imply that
    $F$ must be defined as follows:
    
    \begin{align}
        F(x_S) \equiv \E_{q(X_{\bar S} \mid X_S = x_S)}\big[f(x_S, X_{\bar S})\big]. \label{eq:consistent2}
    \end{align}
    
    To complete the proof, we consider whether there are other ways of deriving $F$'s behavior that may demonstrate inconsistency. The only other case to consider is whether we can derive a unique definition for $F(x_{S \setminus T})$ when beginning from $F(x)$ and when beginning from $F(x_S)$ for $T \subset S \subset D$. The first result is given by Eq.~\ref{eq:consistent2}, and we now derive the second result. To begin, we derive $q(Y \mid X_{S \setminus T})$ from $q(Y \mid X_S)$:
    
    \begin{align*}
        q(y \mid x_{S \setminus T}) &= \sum_{x_T \in \mathcal{X}_T} q(y, x_T \mid x_{S \setminus T}) \tag*{\small (Axiom 1)} \\
        &= \sum_{x_T \in \mathcal{X}_T} \frac{q(y, x_T, x_{S \setminus T})}{q(x_{S \setminus T})} \tag*{\small (Axiom 2)} \\
        &= \sum_{x_T \in \mathcal{X}_T} q(y \mid x_T, x_{S \setminus T}) \frac{q(x_T, x_{S \setminus T})}{q(x_{S \setminus T})} \tag*{\small (Axiom 2)} \\
        &= \sum_{x_T \in \mathcal{X}_T} q(y \mid x_T, x_{S \setminus T}) \cdot q(x_T \mid x_{S \setminus T}) \tag*{\small (Axiom 2)} \\
        &= \E_{q(X_T \mid X_{S \setminus T} = x_{S \setminus T})}\big[q(y \mid x_T, x_{S \setminus T})\big] \tag*{\small (Definition of expectation)}
    \end{align*}
    
    \noindent We can now derive $F(x_{S \setminus T})$ by taking the expectation of this distribution:
    
    \begin{align*}
        F(x_{S \setminus T}) &= \E_{q(Y \mid X_{S \setminus T} = x_{S \setminus T})}[Y] \tag*{\small (Definition of $F$)} \\
        &= \sum_{y \in \mathcal{Y}} y \cdot q(y \mid x_{S \setminus T}) \tag*{\small (Definition of expectation)} \\
        &= \sum_{y \in \mathcal{Y}} \sum_{x_T \in \mathcal{X}_T} y \cdot q(y \mid x_T, x_{S \setminus T}) \cdot q(x_T \mid x_{S \setminus T}) \tag*{\small (Previous derivation)} \\
        &= \sum_{x_T \in \mathcal{X}_T} \E[Y \mid x_T, x_{S \setminus T}] \cdot q(x_T \mid x_{S \setminus T}) \tag*{\small (Interchanging order of sums)} \\
        &= \sum_{x_T \in \mathcal{X}_T} F(x_T, x_{S \setminus T}) \cdot q(x_T \mid x_{S \setminus T}) \tag*{\small (Definition of $F$)} \\
        &= \sum_{x_T \in \mathcal{X}_T} \sum_{x_{\bar S} \in \mathcal{X}_{\bar S}} f(x_T, x_{S \setminus T}, x_{\bar S}) \cdot q(x_{\bar S} \mid x_T, x_{S \setminus T}) \cdot q(x_T \mid x_{S \setminus T}) \tag*{\small (Definition of $F$)} \\
        &= \E_{q(X_T, X_{\bar S} \mid X_{S \setminus T} = x_{S \setminus T})}\big[f(X_T, x_{S \setminus T}, X_{\bar S})\big]] \tag*{\small (Definition of expectation)}
    \end{align*}
    
    This result shows that deriving $F(x_{S \setminus T})$ from $F(x)$ or from $F(x_S)$ yields a consistent result. We conclude that our definition of $F$, which marginalizes out missing features with the conditional distribution induced by $q(X)$, provides the unique subset extension of $f$ that is consistent with $q(X)$.
\end{proof}

\section{Conditional Distribution Approximations} \label{app:conditional_approximations}

We now describe how several approaches to removing features can be understood as approximations of marginalizing out missing features using their conditional distribution.

\subsection{Separate models} \label{app:separate}

Shapley Net Effects \citep{lipovetsky2001analysis}, SPVIM \citep{williamson2020efficient} and the original IME \citep{vstrumbelj2009explaining} require training separate models for each subset of features. As in the main text, we denote these models as $\{f_S : S \subseteq D\}$. Similarly, the univariate predictors approach requires training models with individual features, and feature ablation requires training models with individual features held out. These models are used to make predictions in the presence of missing features, and they can be represented by the subset function $F(x_S) = f_S(x_S)$.


Note that this $F$ satisfies the necessary properties to be a subset extension of $f_D$ (invariance to missing features and agreement with $f_D$ in the presence of all features) despite the fact that its predictions with held out features do not explicitly reference $f_D$. However, under the assumption that each $f_S$ optimizes the population risk, we show that each $f_S$ can be understood in relation to $f_D$. 

For a regression task where each model $f_S$ is trained with MSE loss, the model that optimizes the population risk is the conditional expectation $f_S(x_S) = \E[Y \mid X_S = x_S]$. Similarly, if the prediction task is classification and the loss function is cross entropy (or another strictly proper scoring function, see \citealp{gneiting2007strictly}) then the model that optimizes the population risk is the conditional probability function or Bayes classifier $f_S(x_S) = p(Y \mid X_S = x_S)$. In both cases, if each $f_S$ for $S \subseteq D$ optimizes the population risk, then we observe the following relationship between $F$ and $f_D$:

\begin{align*}
    F(x_S) = f_S(x_S) = \E\big[f_D(X) \mid X_S = x_S\big].
\end{align*}

\noindent This is precisely the approach of marginalizing out missing features from $f_D$ using their conditional distribution.

\subsection{Missingness during training} \label{app:missingness}

INVASE \citep{yoon2018invase} is based on a strategy of introducing missing features during model training (Appendix~\ref{app:invase}). It trains a model where zeros (or potentially other values) take the place of removed features, so that the model can recognize these as missing values and make the best possible prediction given the available information.

We show here how this approach can be understood as an approximation of marginalizing out features with their conditional distribution. First, we note that replacing features with a default value is problematic if that value is observed in the dataset, because the model then faces ambiguity about whether the value is real or represents missingness. This issue can be resolved either by ensuring that the replacement value does not occur in the dataset, or by providing a mask vector $m \in \{0, 1\}^d$ indicating missingness as an additional input to the model.

We assume for simplicity that this binary vector is provided as an additional input, and we let $x \odot m$ (the element-wise product) represent feature masking and $f(x \odot m, m)$ denote the model's prediction. This can be viewed as a technique for parameterizing a subset function $F \in \mathfrak{F}$ because it ensures invariance to the features that are not selected. Specifically, we can write

\begin{equation*}
    F(x_S) = f\big(x \odot m(S), m(S)\big),
\end{equation*}

\noindent where $m(S)$ is a binary vector with ones corresponding to the members in $S$. If we let $M$ denote a random binary mask variable representing feature subsets, then the loss for training this model is:

\begin{equation*}
    \min_f \; \E_{MXY} \Big[ \ell \big(f(X \odot M, M), Y \big) \Big].
\end{equation*}

\noindent Then, if $M$ is independent from $(X, Y)$, we can decompose the loss as follows:

\begin{align*}
    \E_{MXY} \Big[ \ell \big(f(X \odot M, M), Y \big) \Big]
    &= \E_M \E_{XY} \Big[ \ell \big(f(X \odot M, M), Y \big) \Big] \\
    &= \sum_{m} p(m) \cdot \E_{XY} \Big[ \ell \big(f(X \odot m, m), Y \big) \Big].
\end{align*}

For each value of $m$, we can regard $f(x \odot m, m)$ as a separate function on the specified subset of features $\{x_i : m_i = 1\}$. Then, for classification tasks using cross entropy loss, the objective is optimized by the function $f^*$ such that

\begin{equation*}
    f^*(x \odot m, m) = p(Y \mid X_S = x_S),
\end{equation*}

\noindent where $S = \{i : m_i = 1\}$. A similar result holds other strictly proper scoring functions \citep{gneiting2007strictly} and for regression tasks trained with MSE loss (with the conditional probability replaced by the conditional expectation). In these cases, the result is equivalent to marginalizing out missing features according to their conditional distribution.

One issue with INVASE, noted by \cite{jethani2021have}, is that the mask variable $M$ is not independent from the input vector $X$. Intuitively, this means that the selected features can communicate information about the held out features, which then inform the model's prediction about the label $Y$. The INVASE model therefore may not approximate $p(Y \mid X_S = x_S)$, potentially invalidating its information-theoretic interpretation in terms of KL divergence minimization (Section~\ref{sec:information_theory}).

Nonetheless, it is possible to learn a model with missingness introduced at training time that approximates marginalizing out features using their conditional distribution. It suffices to sample masks during training independently from the model input, e.g., by sampling masks uniformly at random \citep{jethani2021have} or according to the distribution described in Appendix~\ref{app:surrogate}.

\subsection{Surrogate models} \label{app:surrogate}

To remove features from an existing model $f$, we can train a surrogate model to match its predictions when features are held out. This technique, described by \cite{frye2020shapley} and implemented by L2X \citep{chen2018learning} and REAL-X \citep{jethani2021have}, can also approximate marginalizing out features using their conditional distribution.

To train such a model, we require a mechanism for removing features. We let $f$ denote the original model and $g$ the surrogate, and similar to the model trained with missingness (Appendix~\ref{app:missingness}), we can remove features using a mask variable $m \in \{0, 1\}^d$. \cite{frye2020shapley} suggest replacing held out features using a value that does not occur in the training set, but we can also provide the mask variable as an input to the surrogate. We therefore represent the surrogate's predictions as $g(x \odot m, m)$, where $g$ can be understood as a subset function $F \in \mathfrak{F}$ (with $m$ defined as in Appendix~\ref{app:missingness}):

\begin{equation*}
    F(x_S) = g\big(x \odot m(S), m(S)\big).
\end{equation*}

\noindent The surrogate is then trained using the following objective function:

\begin{equation}
    \min_g \; \E_{MX} \Big[\ell\big(g(X \odot M, M), f(X)\big)\Big]. \label{eq:surrogate_objective}
\end{equation}

\noindent If $M$ is sampled independently from $X$, then we can decompose the loss as follows:

\begin{align*}
    \E_{MX} \Big[\ell\big(g(X \odot M, M), f(X)\big)\Big]
    &= \E_{M} \E_X \Big[\ell\big(g(X \odot M, M), f(X)\big)\Big] \\
    &= \sum_{m} p(m) \E_X \Big[ \ell\big(g(X \odot m, m), f(X)\big) \Big].
\end{align*}

For each value of $m$, we can regard $g(x \odot m, m)$ as a separate function on the specified subset of features $\{x_i : m_i = 1\}$. Then, for specific loss functions, we can reason about the optimal surrogate that matches the original model $f$ most closely. For models that are compared using MSE loss, we point to a result from \cite{covert2020understanding}:

\begin{align*}
    \min_h \; \E_{X} \Big[ \big(f(X) - h(X_S)\big)^2 \Big] = \E_{X} \Big[ \big( f(X) - \E[f(X) \mid X_S] \big)^2 \Big].
\end{align*}

This shows that to match $f$'s predictions in the sense of MSE loss, the optimal surrogate function $g^*$ is given by

\begin{equation}
    g^*(x \odot m, m) = \E[f(X) \mid X_S = x_S], \label{eq:optimal_surrogate}
\end{equation}

\noindent where $S = \{i: m_i = 1\}$. This result justifies the approach used by \cite{frye2020shapley}; however, while \cite{frye2020shapley} focus on MSE loss, we also find that a cross entropy loss can be used for classification tasks. When the original model $f$ and $g$ both output discrete probability distributions, their predictions can be compared through a soft cross entropy loss, which we define as follows:

\begin{equation*}
    H(a, b) = - \sum_{j} a_j \log b_j.
\end{equation*}

Now, for classifications models that are compared via cross entropy loss, we point to the following result from \cite{covert2020understanding}:

\begin{equation*}
    \min_h \; \E_{X} \Big[ H\big(f(X), h(X_S)\big) \Big] = \E_X \Big[ H\big(f(X), \E[f(X) \mid X_S] \big) \Big].
\end{equation*}

This shows that the optimal surrogate model is once again given by marginalizing out the missing features using their conditional distribution, as in Eq.~\ref{eq:optimal_surrogate}.
Notably, these results require that $X$ and $M$ are sampled independently during training, which is a property that is satisfied by \cite{frye2020shapley} and \cite{jethani2021have} (REAL-X), but not \cite{chen2018learning} (L2X).

The only detail left to specify is the distribution for $M$ in Eq.~\ref{eq:surrogate_objective}. Any distribution that places mass on all $m \in \{0, 1\}^d$ should suffice, at least in principle. Masks could be sampled uniformly at random, but for models with large numbers of features, this places nearly all the probability mass on subsets with approximately half of the features included. We therefore opt to use a distribution that samples the \textit{subset size} uniformly at random.
In our experiments, we sample masks $m$ as follows:

\begin{enumerate}
    \item Sample $k \in \{0, 1, \ldots, d\}$ uniformly at random.
    \item Sample $k$ indices $(i_1, \ldots, i_k)$ from $\{1, 2, \ldots, d\}$ at random and without replacement.
    \item Set $m$ with $m_i = \mathbbm{1}\big(i \in \{i_1, \ldots, i_k\}\big)$.
\end{enumerate}

\section{Information-Theoretic Connections in Regression} \label{app:games}

Here, we describe probabilistic interpretations of each explanation method's underlying set function (Section~\ref{sec:behaviors}) in the context of regression models rather than classification models (Section~\ref{sec:information_theory}). We assume that the models are evaluated using MSE loss, and, as in the main text, we assume model optimality. The different set functions have the following interpretations.

\begin{itemize}
    \item The set function $u_x(S) = F(x_S)$ quantifies the response variable's conditional expectation:
    
    \begin{equation}
        u_x(S) = \E[Y \mid X_S = x_S].
    \end{equation}
    
    This set function lets us examine each feature's true relationship with the response variable.
    
    \item The set function $v_{xy}(S) = - \ell\big(F(x_S), y\big)$ quantifies the squared distance between the model output and the correct label:
    
    \begin{equation}
        v_{xy}(S) = - \Big( \E[Y \mid X_S = x_S] - y \Big)^2. \label{eq:squared_distance}
    \end{equation}
    
    Under the assumption that the response variable's conditional distribution is Gaussian, this represents the pointwise mutual information between $x_S$ and $y$ (up to factors that depend on $S$):
    
    \begin{align}
        \mathrm{I}(y; x_S)
        = \;& - \log p(y) - \frac{1}{2} \log 2\pi - \frac{1}{2} \log \Var(Y \mid X_S = x_S) \nonumber \\
        &- \frac{1}{2} \underbrace{\Big(\E[Y \mid X_S = x_S] - y \Big)^2}_{\textstyle v_{xy}(S)} / \; \Var(Y \mid X_S = x_S).
    \end{align}
    
    \item The set function $v_x(S) = - \E_{p(Y \mid X = x)}\big[\ell\big(F(x_S), Y\big)\big]$ quantifies the squared difference of the conditional expectation from the model output, up to a constant value:
    
    \begin{equation}
        v_x(S) = - \Big(\E[Y \mid X = x]  - \E[Y \mid X_S = x_S]\Big)^2 + c. \label{eq:distribution_squared_distance}
    \end{equation}
    
    Under the assumption that the response variable's distribution conditioned on $X = x$ and $X_S = x_S$ are both Gaussian, this quantity has a relationship with the negative KL divergence between $p(Y \mid X = x)$ and $p(Y \mid X_S = x_S)$:
    
    \begin{align}
        &\KL\big( p(Y \mid X = x) \; \big|\big| \; p(Y \mid X_S = x_S) \big) \nonumber \\
        = \;& \frac{1}{2}\log \frac{\Var(Y \mid X_S = x_S)}{\Var(Y \mid X = x)} - \frac{1}{2} \nonumber \\
        &+ \Big(\Var(Y \mid X = x) + \underbrace{\big( \E[Y \mid X = x] - \E[Y \mid X_S = x_S]\big)^2}_{\textstyle v_x(S)} \Big) / \Big(2 \cdot \Var(Y \mid X_S = x_S)\Big).
    \end{align}
    
    \item The set function $v(S) = - \E_{XY}\Big[\ell\big(F(X_S), Y\big)\Big]$ quantifies the explained variance in the response variable $Y$, up to a constant value:
    
    \begin{align}
        v(S) = \Var(Y) - \E[\Var(Y \mid X_S)] + c. \label{eq:response_conditional_variance}
    \end{align}
    
    Using the entropy maximizing property of the Gaussian distribution \citep{cover2012elements}, we see that the explained variance has the following relationship with the mutual information between $X_S$ and the response variable $Y$:
    
    \begin{align}
        \mathrm{I}(Y ; X_S)
        &= H(Y) - \E\big[H(Y \mid X_S)\big] \nonumber \\
        &\geq H(Y) - \frac{1}{2} \E\Big[\log \big( 2 \pi e \cdot  \Var(Y \mid X_S) \big) \Big] \nonumber \\
        &\geq H(Y) - \frac{1}{2} \log 2 \pi e - \frac{1}{2} \log \underbrace{\E\big[\Var(Y \mid X_S)\big]}_{\textstyle v(S)}.
    \end{align}
    
    Equality is achieved in the first bound if the distribution $p(Y \mid X_S = x_S)$ is Gaussian. The second bound is due to Jensen's inequality.
    
    
    
    \item The set function $w_x(S) = - \ell\big(F(x_S), f(x)\big)$ quantifies the squared difference between the model output and the expected model output when conditioned on a subset of features:
    
    \begin{equation}
        w_x(S) = - \Big( f(x) - \E[f(X) \mid X_S = x_S] \Big)^2. \label{eq:output_distance}
    \end{equation}
    
    If we assume that $p(f(X) \mid X_S)$ is a Gaussian distribution, then we can view this as a component in the KL divergence between $p(f(X) \mid X)$ (a deterministic distribution) and $p(f(X) \mid X_S = x_S)$:
    
    \begin{align}
        &\KL\big(p(f(X) \mid X = x) \; \big|\big| \; p(f(X) \mid X_S = x_S) \big) \nonumber \\
        = \;&\lim_{\sigma \to 0} \; \frac{1}{2} \log \frac{\Var(f(X) \mid X_S = x_S)}{\sigma^2} - \frac{1}{2} \nonumber \\
        &\quad\quad+ \Big(\sigma^2 + \underbrace{\big( f(x) - \E[f(X) \mid X_S = x_S] \big)^2}_{\textstyle w_X(S)}\Big) / \Big(2\Var(f(X) \mid X_S = x_S)\Big) \nonumber \\
        = \;&\infty.
    \end{align}
    
    This result is somewhat contrived, however, because the KL divergence is ultimately infinite.
    
    \item The set function $w(S) = - \E_{X}\Big[\ell\big(F(X_S), f(X)\big)\Big]$ quantifies the explained variance in the model output, up to a constant value:
    
    \begin{equation}
        w(S) = 
        \Var\Big(f(X)\Big) - \E\Big[\Var\big(f(X) \mid X_S\big)\Big] + c. \label{eq:output_conditional_variance}
    \end{equation}
    
    This is the variance decomposition result presented for Shapley Effects \citep{owen2014sobol}. Using the same entropy maximizing property of the Gaussian distribution, we can also see that the explained variance is related to the mutual information with the model output, which can be viewed as a random variable $f(X)$:
    
    \begin{align}
        \mathrm{I}\big(f(X) ; X_S\big)
        &= H\big(f(X)\big) - \E\big[H(f(X) \mid X_S)\big] \nonumber \\
        &\geq H\big(f(X)\big) - \frac{1}{2} \E\Big[ \log \big( 2 \pi e \cdot \Var(f(X) \mid X_S) \big) \Big] \nonumber \\
        &\geq H\big(f(X)\big) - \frac{1}{2} \log 2 \pi e - \frac{1}{2} \log \underbrace{\E\Big[\Var\big(f(X) \mid X_S\big)\Big]}_{\textstyle w(S)}
    \end{align}
    
    Equality is achieved in the first bound if $f(X)$ has a Gaussian distribution when conditioned on $X_S = x_S$. The second bound is due to Jensen's inequality.
\end{itemize}

These results are analogous to those from the classification case, and they show that each explanation method's set function has an information-theoretic interpretation even in the context of regression tasks. However, the regression case requires stronger assumptions about the data distribution to yield these information-theoretic links (i.e., Gaussian distributions).
To avoid strong distributional assumptions, it is more conservative to interpret these quantities in terms of Euclidean distances (Eqs.~\ref{eq:squared_distance}, \ref{eq:distribution_squared_distance}, \ref{eq:output_distance}) and conditional variances (Eqs.~\ref{eq:response_conditional_variance}, \ref{eq:output_conditional_variance}).

\section{Experiment Details} \label{app:experiments}

This section provides additional details about models, datasets and hyperparameters, as well as some additional results.

\subsection{Hyperparameters}

The original models used for each dataset are:

\begin{itemize}
    \item For the census income dataset, we trained a LightGBM model with a maximum of 10 leaves per tree and a learning rate of 0.05 \citep{ke2017lightgbm}.
    \item For MNIST, we trained a 14-layer CNN consisting of convolutional layers with kernel size 3, max pooling layers, and ELU activations \citep{clevert2015fast}. The output was produced by flattening the convolutional features and applying two fully connected layers, similar to the VGG architecture \citep{simonyan2014very}. We trained the model with Adam using a learning rate of $10^{-3}$ \citep{kingma2014adam}.
    \item For the BRCA dataset, we trained a $\ell_1$ regularized logistic regression model and selected the regularization parameter using a validation set.
\end{itemize}

For the BRCA dataset, to avoid overfitting, we also randomly selected a subset of 100 genes to analyze out of 17,814 total. To ensure that a sufficient number BRCA-associated genes were selected, we tried 10 random seeds for the gene selection step and selected the seed whose 100 genes achieved the best performance (displayed in Table~\ref{tab:genes}). A small portion of missing expression values were imputed with their mean, and the data was centered and normalized prior to fitting the regularized logistic regression model.

When generating explanations using feature removal approaches that required sampling multiple values for the missing features (marginalizing with uniform, product of marginals, or joint marginal), we used 512 samples for the census income and MNIST datasets and 372 for the BRCA dataset (the size of the train split).

As described in the main text (Section~\ref{sec:experiments}) and in Appendix~\ref{app:surrogate}, we trained surrogate models to represent marginalizing out features according to the conditional distribution. Our surrogate models were trained as follows:

\begin{itemize}
    \item For the census income data, the surrogate was a MLP with a masking layer and four hidden layers of size 128 followed by ELU activations. During training, the mask variable was sampled according to the procedure described in Appendix~\ref{app:surrogate}. Our masking layer replaced missing values with -1 (a value that did not occur in the dataset) and also appended the mask as an additional set of features, which improved its ability to match the original model's predictions.
    \item For MNIST, the surrogate was a CNN with an identical architecture to the original model (see above) except for a masking layer at the input. The masking layer replaced missing values with zeros and appended the mask along the channels dimension.
    \item For the BRCA data, the surrogate was an MLP with two hidden layers of size 64 followed by ELU activations. The masking layer replaced missing values with their mean and appended the mask as an additional set of features.
\end{itemize}

\begin{table}[t]
\caption{List of genes analyzed.}
\label{tab:genes}
\vskip 0.2in
\begin{center}
\begin{small}
\begin{tabular}{cccccc}
\toprule
Genes 1-17 & Genes 18-34 & Genes 35-51 & Genes 52-68 & Genes 69-85 & Genes 86-100 \\
\midrule
OSTbeta & NBR2 & TSHR & HPS4 & GRINA & C20orf111 \\
STATH & CCDC64 & C7 & ZFPM1 & YTHDF3 & OMA1 \\
MAPK10 & NUP210 & CRYBB2 & OAS2 & TMCC1 & NCAPH2 \\
PLEKHG5 & HEMGN & PPAPDC3 & TUBA1C & UBE1DC1 & GPX2 \\
ERO1L & SLC25A3 & TXNL4B & OR8K5 & C6orf15 & BPY2C \\
ZNF711 & LEF1 & CHST9 & THSD3 & PDE6A & ZNF324 \\
ZNF385 & MVD & HACE1 & ATP6V0C & PEO1 & CDC27 \\
OR52E8 & OTUD3 & AYTL1 & RAB22A & TMEM52 & CCNB2 \\
SLC5A11 & KIAA1949 & PRSS35 & AP1B1 & PARP1 & CNOT7 \\
P4HA3 & SLC44A3 & ZNF408 & CTAGE6 & GSS & BIRC3 \\
LHFPL4 & ZNF775 & DDC & C6orf26 & RDH11 & GAL3ST3 \\
MGC33657 & THY1 & CSTL1 & ESR1 & STXBP1 & PLEKHM1 \\
CAPZB & DYNC1I2 & OR2F1 & UPK3B & ACLY & SPOCD1 \\
RBM15B & CYP1A1 & C12orf50 & ROBO4 & TMSB10 & PENK \\
C1orf176 & SPTA1 & SH3YL1 & TMEFF1 & TUBB & TAS2R9 \\
KLF3 & CLEC4M & SNUPN & KIAA1279 & LIPK & \\
OLFM4 & RXFP3 & COL25A1 & ZFP36L1 & HRC & \\
\bottomrule
\end{tabular}
\end{small}
\end{center}
\vskip -0.2in
\end{table}

\subsection{Additional results}

\begin{figure}
\begin{center}
\includegraphics[width=\columnwidth]{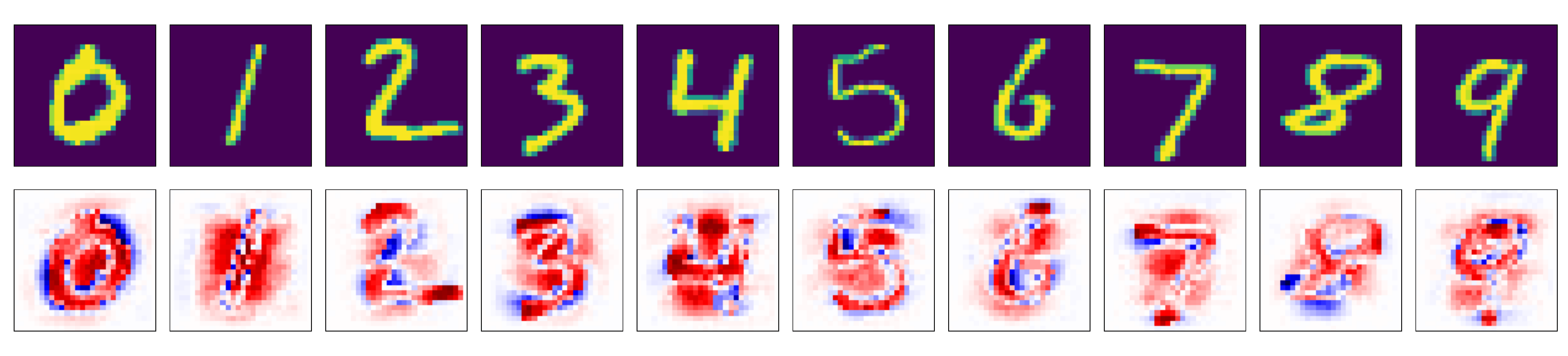}
\caption{MNIST prediction loss explanations using LossSHAP (feature removal with the conditional distribution and summary with the Shapley value).}
\label{fig:mnist_examples}
\end{center}
\vskip -0.2in
\end{figure}

We present two supplementary results for the experiments described in Section~\ref{sec:experiments}. Figure~\ref{fig:mnist_examples} shows more examples of MNIST explanations using the combination of the conditional distribution and Shapley value (i.e., LossSHAP). These explanations consistently highlight important pixels within the digit as well as empty regions that distinguish the digit from other possible classes (e.g., see 3, 4, 9).

\begin{figure} 
\vskip -0.4in
\begin{center}
\centerline{\includegraphics[width=0.78\columnwidth]{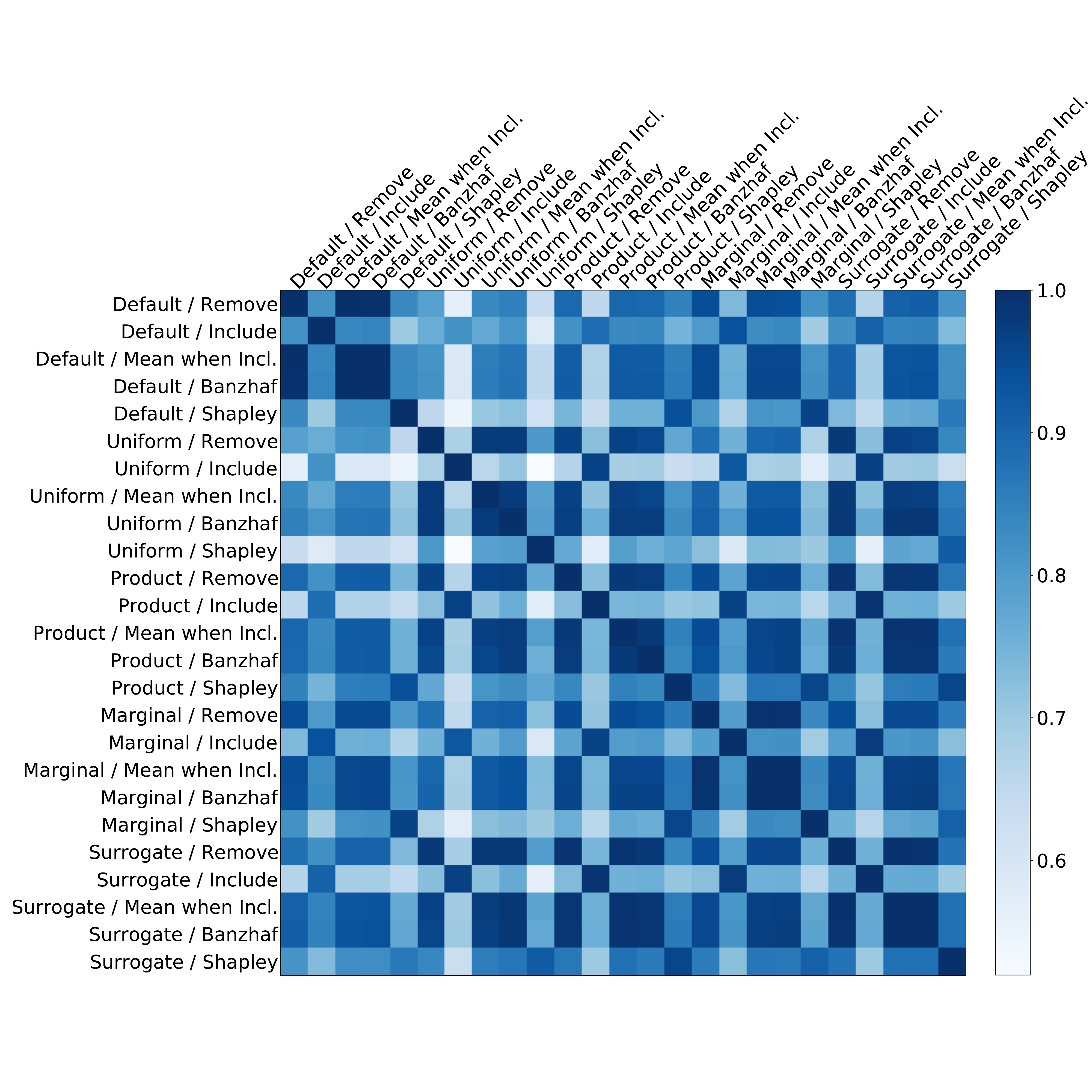}}

\vskip -0.8in

\centerline{\includegraphics[width=0.78\columnwidth]{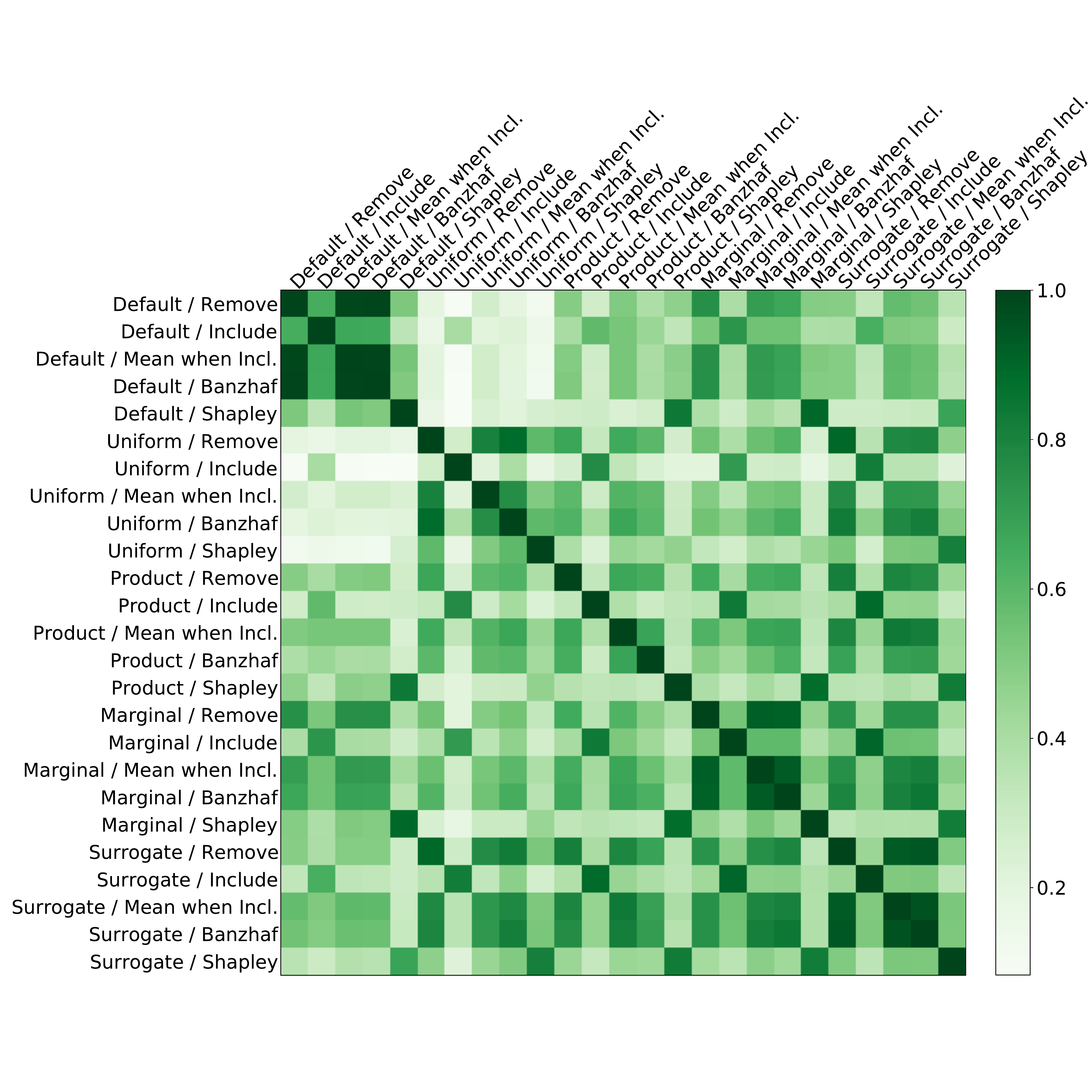}}

\vskip -0.45in

\caption{Mean correlation (top) and Spearman rank correlation (bottom) between different explanation methods on the BRCA dataset.}
\label{fig:brca_correlation}
\end{center}
\vskip -0.2in
\end{figure}

Figure~\ref{fig:brca_correlation} quantifies the similarity between dataset loss explanations for the BRCA dataset using their correlation (top) and Spearman rank correlation (bottom). We see patterns in these explanations that are similar to those seen with the census dataset: explanations generated using Shapley values are relatively similar to those that remove individual features or use Banzhaf values, while the include individual technique tends to differ most from the others. Because these plots visualize correlation rather than Euclidean distance, we can also see that Banzhaf value explanations are correlated with those that use the mean when included technique, which was predicted by our theory (Section~\ref{sec:solution_concepts}). 
Interestingly, the Shapley value explanations are more strongly correlated across different removal strategies than they are to explanations that use other summarization strategies but
that remove features in the same way (see bottom row of Figure~\ref{fig:brca_correlation}).

\clearpage
\bibliography{20-1316}

\end{document}